\documentclass[11pt]{./elsarticle}

\newif\ifshorttables
\shorttablestrue

\usepackage{xcolor}
\usepackage{amsfonts}
\usepackage{mathrsfs}
\usepackage{amsmath}
\usepackage{amssymb}
\usepackage{amsthm}
\usepackage{bbm}
\usepackage[margin=1in]{geometry}
\usepackage{multirow}
\usepackage{enumitem}
\usepackage{array}

\usepackage[hidelinks]{hyperref}
\usepackage{orcidlink}
\usepackage{subcaption}
\usepackage{soul}

\biboptions{sort&compress}

\definecolor{darkgreen}{rgb}{0.0, 0.5, 0.0}

\newcommand{\TODO}[1]{{\bf{\textcolor{black}{#1}}}}

\newcommand{\revone}[1]{\textcolor{black}{#1}}
\newcommand{\revtwo}[1]{\textcolor{black}{#1}}

\newcommand{\revboth}[1]{\textcolor{black}{#1}}

\newcommand\Tstrut{\rule{0pt}{2.6ex}}         
\newcommand\Bstrut{\rule[-0.9ex]{0pt}{0pt}}   

\journal{Journal of Computational Physics}

\usepackage[symbol]{footmisc}
\usepackage{etoolbox}
\makeatletter
\def\ps@pprintTitle{\let\@oddfoot\relax \let\@evenfoot\relax}
\makeatother

\begin{document}

\begin{frontmatter}

\title{Active operator learning with predictive uncertainty quantification for partial differential equations}

\author[1]{Nick Winovich\footnote[2]{Equal contribution authors.}}
\author[2,3]{Mitchell Daneker$^{\dagger}$}
\author[2,4]{Lu Lu\footnote[1]{Corresponding authors. Email: lu.lu@yale.edu, guanglin@purdue.edu}}
\author[5,6,7]{Guang Lin$^{*}$}


\address[1]{Sandia National Laboratories, Albuquerque, NM 87123}
\address[2]{Department of Statistics and Data Science, Yale University, New Haven, CT 06511}
\address[3]{Department of Chemical and Biomolecular Engineering, University of Pennsylvania, Philadelphia, PA 19104}
\address[4]{Department of Chemical and Environmental Engineering, Yale University, New Haven, CT 06511}
\address[5]{Department of Mathematics, Purdue University, West Lafayette, IN 47907}
\address[6]{Department of Statistics, Purdue University, West Lafayette, IN 47907}
\address[7]{Department of Mechanical Engineering, Purdue University, West Lafayette, IN 47907}

\begin{abstract}

\textcolor{black}{
  With the increased prevalence of neural operators being used to provide rapid solutions to partial differential equations (PDEs), %
  understanding the accuracy of model predictions and the associated error levels is 
  necessary for deploying reliable surrogate models in scientific applications. Existing uncertainty quantification (UQ) frameworks employ ensembles or Bayesian methods, which can incur substantial computational costs during both training and inference. We propose a lightweight predictive UQ method tailored for Deep operator networks (DeepONets) that also generalizes to other operator networks. Numerical experiments on linear and nonlinear PDEs demonstrate that the framework's uncertainty estimates are unbiased and provide accurate out-of-distribution uncertainty predictions with a sufficiently large training dataset. Our framework provides fast inference and uncertainty estimates that can efficiently drive outer-loop analyses that would be prohibitively expensive with conventional solvers. 
    We demonstrate how predictive uncertainties can be used in the context of Bayesian optimization and active learning problems to yield improvements in accuracy and data-efficiency for outer-loop optimization procedures. %
    In the active learning setup, we extend the framework to Fourier Neural Operators (FNO) and describe a generalized method for other operator networks. To enable real-time deployment, we introduce an inference strategy based on precomputed trunk outputs and a sparse placement matrix, reducing evaluation time by more than a factor of five. Our method provides a practical route to uncertainty-aware operator learning in time-sensitive settings. 
}

\end{abstract}


\end{frontmatter}


\section{Introduction}

\revboth{Understanding and managing uncertainty is essential for data-driven models in high-consequence, real-time environments. This has motivated extensive research on uncertainty quantification (UQ), which estimates confidence in predictions based on data availability~\cite{smith2013uncertainty,sullivan2015introduction,karniadakis2021}. At the same time, scientific and engineering teams are quickly adopting neural surrogates to accelerate demanding tasks such as solving PDEs and performing adaptive simulations~\cite{ZHOU2025117990}. These models offer speed and flexibility, but ensuring their predictions are trustworthy via traditional UQ remains a challenge due to high computational costs and the black-box nature of neural surrogates. }

\revboth{Neural surrogates leverage the Universal Approximation Theorem, which states that neural networks can approximate any continuous function under mild conditions~\cite{cybenko1989approximation,hornik1989multilayer,leshno1993multilayer}. Building on this, operator networks such as DeepONet~\cite{lu2021learning} and Fourier Neural Operators (FNO)~\cite{li2020fourier} extend the idea by learning solution operators for entire families of PDEs with a single trained model~\cite{zhang2024d2no,zhang2024modno}. When using neural surrogates, however, accurate approximations from large datasets are not enough; without trustworthy uncertainty information, surrogates can mislead downstream decisions critical for creating effective control policies and guiding experimental design. This undermines the robustness of adaptive workflows, highlighting a pressing open question: how can these models be augmented with computationally efficient and reliable uncertainty estimates to support trustworthy decision-making~\cite{zhu2023_2,lee2024}?}

\revboth{Current UQ techniques for operator networks frequently use ensembles or Bayesian training schemes. Randomized-prior techniques like UQDeepONet~\cite{yang2022scalable} and Bayesian DeepONet~\cite{lin2022b} offer useful uncertainty estimates but have significant computational overhead, particularly during inference time when parallelism is constrained. Ensemble-based UQ evaluation costs scale linearly with ensemble size and as such, is generally impractical for real-time applications without access to considerable high-performance computing resources. As DeepONets have been promoted as fast surrogates~\cite{lin2023learning},
with work like POD-DeepONet~\cite{lu2022comprehensive} targeting even further speed improvements, %
introducing computational bottlenecks to facilitate UQ fundamentally undermines their practical utility.} 

\revboth{ These limitations are particularly important because the uses of uncertainty quantification extends beyond error estimation; it supports adaptive procedures such as model improvement via active learning (AL) and target optimization via Bayesian optimization (BO). Pickering et al.~\cite{pickering2022} demonstrated that DeepONet models equipped with UQ can guide data acquisition for AL, reducing costs in complex simulations and experiments~\cite{settles2009,gal2017,houlsby2011,karniadakis2021}. Related strategies, such as BO, leverage UQ to balance exploration and exploitation. The literature leaves a gap for lightweight, \emph{predictive} uncertainty methods that calibrate to model error in deterministic-training regimes and integrate seamlessly with outer-loop tasks.}

\begin{figure}[t]
\centering
\includegraphics[width=0.925\textwidth]{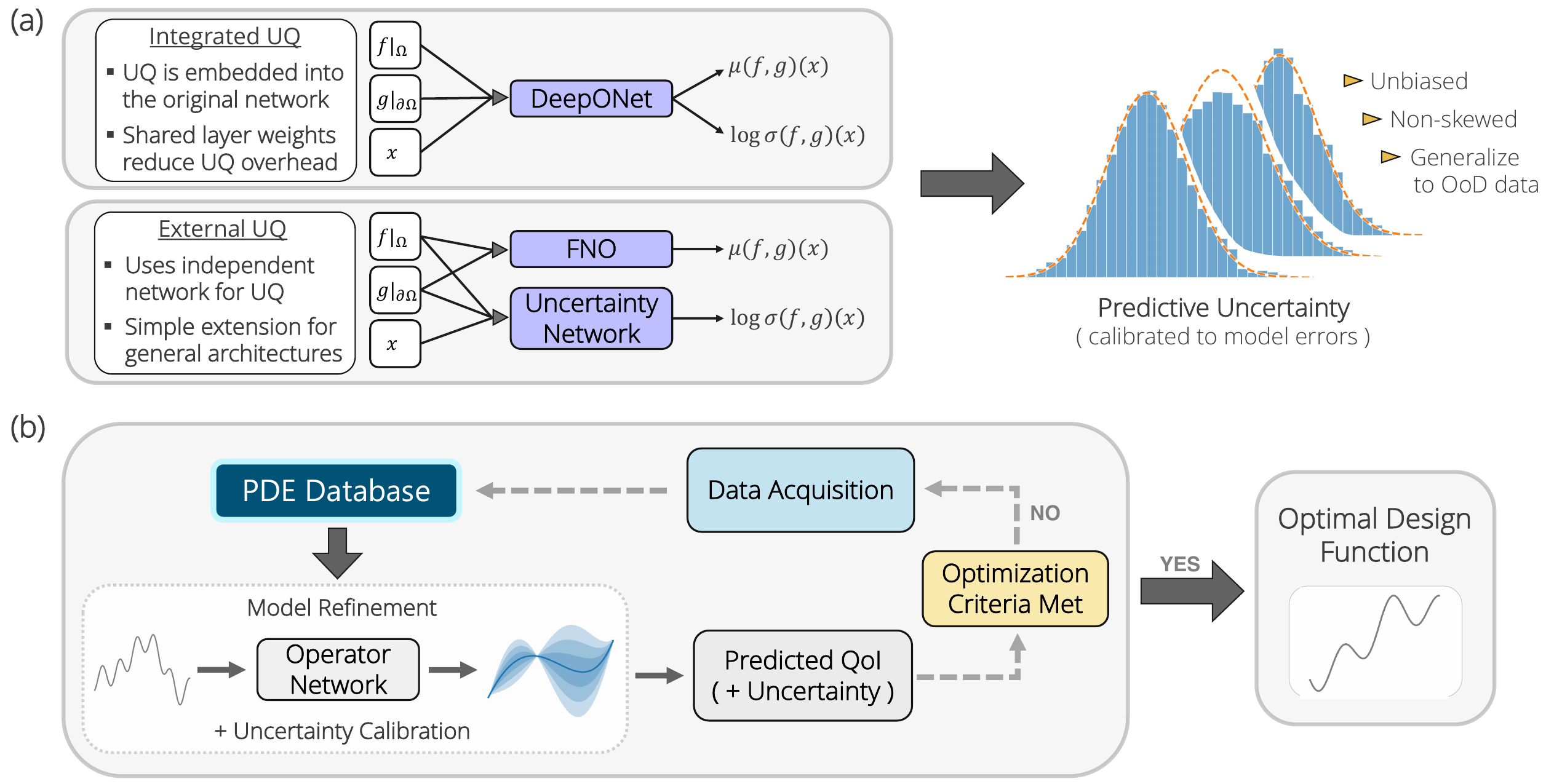}
\caption{{{(a)}} The uncertainty-equipped operator network architecture splits predictions into mean and variance estimates. \revboth{Depending on the operator network used, the uncertainty predictions could be integrated into the network like for our DeepONet, or an external (secondary) network such as for FNO. Which framework is used will depend on network architecture, but generalizes to any operator network.} We interpret network outputs as parameters for a predictive probability distribution, which are calibrated to the observed error distributions during training.  {{(b)}} The uncertainty estimates provided by the operator networks are employed to help guide outer-loop \revboth{data aquisition} procedures in {\emph{function spaces}}.  %
  The quantity of interest (QoI) and variance estimates are derived from the operator network predictions, and this information is used to guide data acquisition by balancing exploitation and exploration.} %
\label{fig:main_diagram}
\end{figure}

\revboth{In this work, we introduce a generalizable \emph{predictive} UQ framework for operator networks, based on ConvPDE-UQ networks~\cite{winovich2019convpde} and developed in parallel with the works of Moya et al.~\cite{moya2022deeponet,moya2025physicad}.  In contrast to Moya's 2025 work comparing conformalized DeepONets with Bayesian ensembles and other works in the literature, our framework extends predictive UQ to generalized operators for regression tasks. Additionally, we integrate our network into Bayesian-optimization and active-learning pipelines, with examples using DeepONet and FNO. We introduce a novel placement matrix for more efficient structured outputs during model deployment and provide a detailed architecture for DeepONet that delivers calibrated variance estimates using a \emph{single} network. Predictive uncertainty avoids ensembles and keeps overhead low during both training and inference. The framework interprets network outputs as parameters of a predictive distribution and calibrates these parameters to observed error distributions via a negative log-likelihood (NLL) loss. To separate mean and uncertainty predictions, we split the final layers of the branch and trunk and use a log-$\sigma$ parameterization for stability.}

\revboth{We validate our approach on linear and nonlinear PDEs and assess generalization on both in-distribution and out-of-distribution (OoD) function spaces. 
The predictive uncertainty estimates are well-aligned with observed errors near the tail-end of training, remain unbiased and non-skewed, and increase appropriately in more challenging OoD regimes while staying low in simpler cases. 
Finally, we demonstrate how predictive UQ accelerates outer-loop procedures in a visually intuitive BO example in 1D time with two peaks in parameter space as well as a more demanding AL setup for a 2D advection problem where we compare traditional UQ methods (Bayesian networks and Ensembles) to ours. We finally showcase our method in an AL scenario where traditional UQ is too expensive,  a 2D wave equation on a star-shaped domain, showing improved data efficiency and accuracy relative to baselines without UQ. An overview of the uncertainty-equipped networks and outer-loop workflows is provided in Fig.~\ref{fig:main_diagram}, with distinctions for our optimized \emph{single} network DeepONet (integrated UQ) and using secondary networks for architectures like FNO (external UQ).}

The main contributions of this work are summarized below.
\begin{itemize}
    \item \textbf{Reliable predictive UQ for DeepONet.} A single-network framework that calibrates uncertainty estimates to observed model errors without ensembles, yielding well-aligned uncertainty and mean predictions.
    \item \revboth{\textbf{Generalized UQ framework for operator networks.} Implementation details for how our framework can be extended to general operator learning models, with an example using FNO.} 
    \item \textbf{Efficient inference.} An optimized inference method with precomputed trunk outputs and a sparse placement matrix for structured outputs.
    \item \textbf{Generalization to OoD data.} Predictive uncertainties that reflect changing difficulty across function-space regimes, increasing on harder OoD examples and remaining low on simpler examples.
    \item \textbf{Accelerated outer-loop learning.} Integration of predictive UQ with BO and AL to improve data-efficiency and accuracy.
\end{itemize}

\revboth{The remainder of this paper is organized as follows. Section~\ref{sec:deeponet} reviews DeepONet. Section~\ref{sec:deeponetuq} details the predictive uncertainty framework. Section~\ref{sec:precomp} introduces the optimized inference workflow and placement matrix. Section~\ref{sec:datagen} describes the data generation procedure. Section~\ref{sec:validation} presents experimental results and uncertainty analysis. Sections~\ref{sec:bo} and ~\ref{sec:active} demonstrate BO and AL with predictive UQ, respectively. We conclude with limitations and future directions in Section~\ref{sec:conclusion}.}

\section{Uncertainty quantification in deep operator networks} 

\subsection{Deep operator networks}
\label{sec:deeponet}

DeepONet architectures are defined by %
two central components~\cite{lu2021learning,jin2022mionet,lu2022comprehensive}, referred to as the {\emph{branch network}} and {\emph{trunk network}}. %
The trunk network parses input coordinates corresponding to evaluation locations on the spatial (and/or temporal) domain $\Omega\subset\mathbb{R}^d$, and the branch network is responsible for processing information associated with the input functions, $(f,g) \in C(\Omega)\times C (\partial \Omega)$.
The branch and trunk components produce vectors $b\in\mathbb{R}^N$ and $t\in\mathbb{R}^N$ %
with features extracted by the respective networks; these features are then combined to form the final network prediction, which is defined by the inner product %
$\widehat{u}(f,g)(x) = \langle \, b \, , \, t \, \rangle$ for each location $x\in\overline{\Omega}$.  %

This predefined method of combining the outputs of the neural network components slightly restricts the expressivity of the resulting model, but it also provides a practical advantage with respect to interpretability. Since the output vector $t$ of the trunk network depends solely on the input coordinates $x$, it is natural to interpret the components of the trunk output as functions defined on the underlying spatial domain of the PDE:
\begin{equation*}
\operatorname{Trunk}(x) \,\,\, = \,\,\,  \big[ \,\, \raisebox{0.035in}{\rule{0.2in}{0.5pt}} \,\,\, t \,\,\, \raisebox{0.035in}{\rule{0.2in}{0.5pt}} \,\, \big]     \,\,\, = \,\,\,   \big[  \, \varphi_{1}(x)\, , \,\dots\, , \,\varphi_{N}(x) \,\big]
\end{equation*}

The component functions of the trunk network often bear a resemblance to the basis functions from finite element methods (FEM)~\cite{strang1973analysis,larsson2003partial}. %
For DeepONet models, these basis functions are learned by the trunk network automatically during training, as opposed to being specified beforehand in the case of  FEM calculations. The branch network assumes the role of determining the appropriate coefficients, $b$, for the family of basis functions %
proposed by the trunk. The resulting network approximation can then be expressed as
\begin{equation*}
\widehat{u}(f,g)(x) \, = \, \langle \,b \,,\, t \,\rangle \, = \,\, \sum\nolimits_{i=1}^N \, b_i \cdot \varphi_i(x)
 \hspace{0.125in} \mbox{where} \hspace{0.125in} b \, = \, \big[  \, b_{1}\, , \,\dots\, , \,b_{N} \,\big]  \, = \, \operatorname{Branch}(f,g).
\end{equation*}
This network prediction is evaluated by comparing it with the true PDE solution, $u(f,g)$, associated with the current input functions $(f,g)$, 
and the weights of the branch and trunk are updated to reduce discrepancies between the predicted and known solutions. %
More precisely, for a given input function pair $(f,g) \in C(\Omega)\times C (\partial \Omega)$, the DeepONet prediction $\widehat{u}(f,g)$  %
is assigned a loss for each $x\in\overline{\Omega}$ given by
\begin{equation*}
  \mbox{Loss}(\widehat{u} \, ; \, x, f, g, u) \,\, = \,\,
  |\widehat{u}(f,g)(x) - u(f,g)(x)|^2.
\end{equation*}
This loss is evaluated at randomly sampled locations on the interior and boundary of the domain to enforce both constraints of the target PDE.  %
In contrast to function-approximation networks, %
the model is evaluated on a family of distinct input function pairs simultaneously during training.  That is, the loss is averaged over spatial samples $x\in \overline{\Omega}$ and over function samples $(f,g) \in C(\Omega)\times C (\partial \Omega)$.  %
This allows the model to approximate solutions to new realizations of input functions $(f,g)$ after training, 
providing a surrogate model for the solution operator of the PDE.

\subsection{Equipping operator networks with predictive uncertainties}
\label{sec:deeponetuq}

In this section, we describe how DeepONet architectures can be extended to provide predictive uncertainties following a procedure analogous to that used in the ConvPDE-UQ framework~\cite{winovich2019convpde,winovich2021neural}.   %
Under this framework, the model provides its predictions in the form of probability distributions instead of pointwise estimates. %
More precisely, the network is designed to produce parameter estimates for a class of distributions (such as the mean and standard deviation for a normal distribution).
These parameters are calibrated during training to capture the distribution of observed errors; the resulting model then has the ability to provide uncertainty information along with its predictions. %

\begin{figure}[tb]
\centering
\hspace{-0.225in}\includegraphics[width=0.90\textwidth]{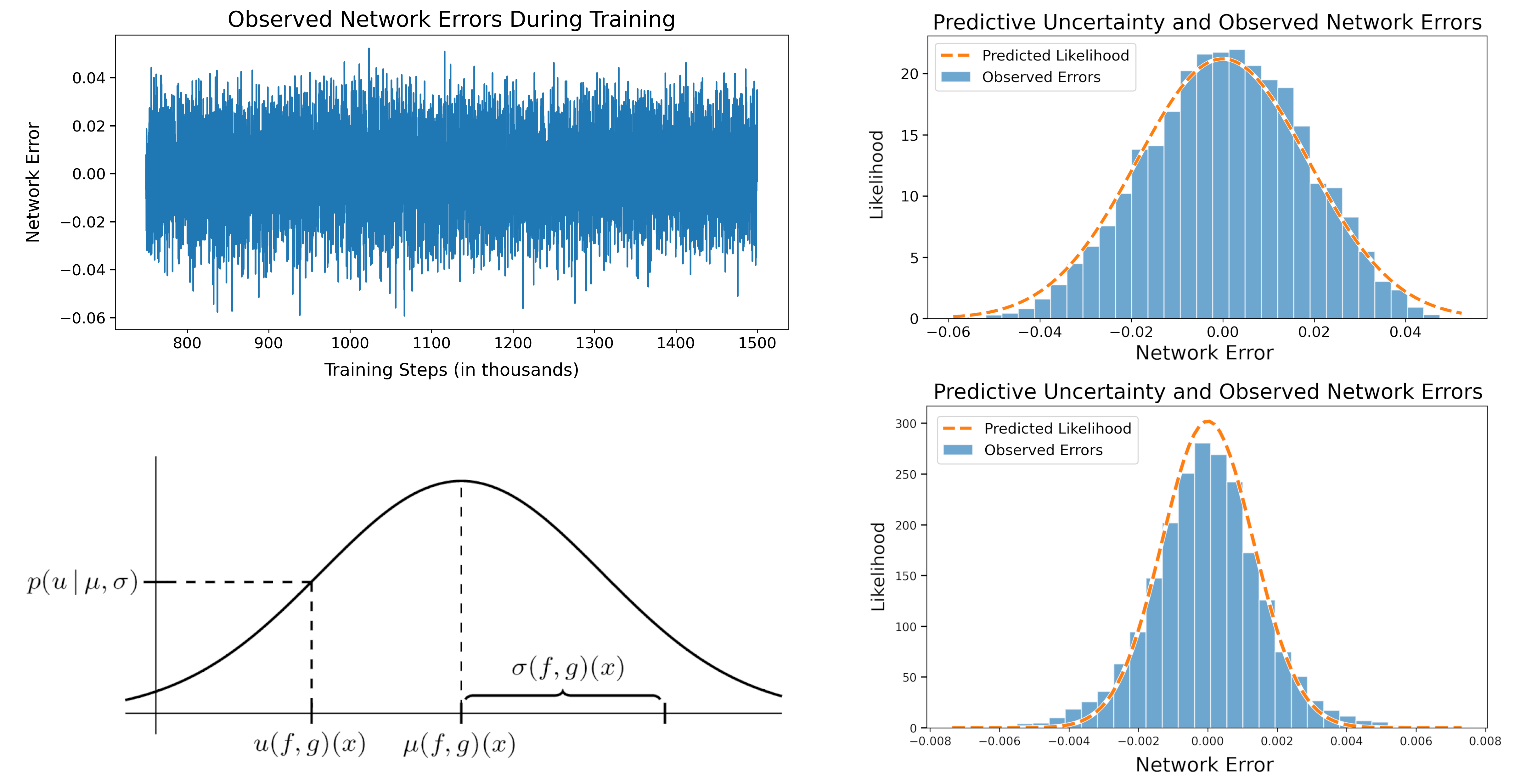}
\caption{Observed network errors during training (top left) and interpretation of network outputs as parameters for a normal distribution (bottom left). The means and standard deviations provided by the network are used to assign likelihoods to the observed target values, and the network aims to increase these likelihoods during the training process.  %
  Histograms of the prediction errors for two input examples are shown to the right with predictive uncertainties overlaid to illustrate how the network uncertainty matches the observed errors near the end of training.  }
\label{fig:pred_uq_fit_new}
\end{figure}

To train DeepONets using this probability-based framework, 
we first need to select an appropriate parameterized family of probability distributions.
In practice, this can be done by analyzing the empirical distribution of network errors observed during training. More precisely, a histogram of the observed network errors, such as the ones shown on the right of Fig.~\ref{fig:pred_uq_fit_new}, 
can be compared with the density functions of common families of distributions, such as normal distributions. %

Once an appropriate family of parameterized probability distributions is selected, the network loss is defined with respect to the associated log likelihood function, following the analogy of Gaussian process regression used in the ConvPDE-UQ framework~\cite{winovich2019convpde,winovich2021neural}. %
For example, when the network errors are observed to follow a normal distribution, the model's proposed likelihood for the true solution is $p(u; \mu, \sigma) = (2\pi\sigma^2)^{-1/2} \, \exp(-\tfrac{1}{2} |\mu(f,g)(x) - u (f,g)(x)|^2/\sigma^2)$, and the loss function is defined by:
\begin{equation}
  \mbox{Loss}(\mu, \sigma \, ; \, x , f, g, u)   \,\,  =  \, \,  \tfrac{1}{2} \big( \mu(f,g)(x) \, - \, u(f,g)(x) \big)^2 \big/ \big(\sigma(f,g)(x) \big)^2  \, + \, \tfrac{1}{2}\log\big(2\pi(\sigma(f,g)(x))^2\big). \label{eq:prob_loss}
\end{equation}
In this way, the network outputs $\mu(f,g)(x)$ and $\log \sigma(f,g)(x)$\footnote{To improve the stability of the numerical implementation, we interpret the uncertainty component of the network output on a logarithmic scale.} %
are interpreted as parameters for a predictive probability distribution placed on the values of the true solution, %
as illustrated in Fig.~\ref{fig:pred_uq_fit_new}. %

After conducting experiments with several architecture variations (described in \ref{sec:deeponet_boundary_conditions}), %
the network structure illustrated in Fig.~\ref{fig:main_diagram}(a) was found to provide the best performance. %
For the experiments presented in this work, the branch and trunk networks each consist of 5 fully-connected layers with 50 units and 30 units per layer, respectively. Each network layer uses Leaky ReLU activation functions~\cite{maas2013rectifier}, with the exception of the {{final layer of the branch network}} where no activation is applied. In order to equip the networks with predictive uncertainties, the final two layers of the branch and trunk are split to provide independent processing for the mean and uncertainty parameters. Specifically, two distinct pairs of 50 unit layers are used to produce the mean output $b$ and uncertainty output $b_\sigma$ of the branch network; similarly, two distinct pairs of 30 unit layers are used to produce the trunk outputs $t$ and $t_\sigma$. %
The final network outputs for the distribution parameters are then defined by the inner products: $\mu(f,g) = \langle  b, t\rangle$ and $\log \sigma(f,g) = \langle  b_\sigma, t_\sigma\rangle$. %
For the experimental results presented in this work, the number of basis functions $N$ was taken to be $150$; however, %
the performance of the models remain essentially unchanged for any choice of $N>50$.

\revboth{Although we initially developed our method for DeepONets, it is important to consider how this framework generalizes to other operator learning architectures such as FNOs~\cite{li2020fourier, wen2022UFNO,lehmann2024FFNO,zipeng2024neurips,Grady2023parallelFNO}, graph neural operators~\cite{li2020neural}, and transformer-based models~\cite{SHIH2025117560, NEURIPS2024_2cd36d32, calvello2024continuum, wu2024transolver}, which have seen extensive growth in the last few years~\cite{lin2021operator, wang2021learning,di2021deeponet,goswami2022deep, mao2024disk2planet}. In Sections~\ref{sec:advection} and~\ref{sec:active}, we focus on extending our framework to FNOs and leave other operator architectures for future research. Since FNOs operate in Fourier space, where evaluation grids constrain the learned mapping~\cite{lu2022comprehensive}, it is difficult to integrate predictive UQ estimates directly within FNO architectures.  We found that introducing a secondary network for UQ, which is jointly trained with the primary model, works best. The organization of the uncertainty prediction network provides room for exploration in future research; however, this is beyond the scope of this work, and we adopt a unified training scheme for our numerical experiments.}

\subsubsection{Uncertainty Types and Their Role in Operator Learning}

We note that traditional uncertainty quantification can be broken up into two types of uncertainty: aleatoric, which relates to inherent noise in the data, and epistemic, which reflects uncertainty due to limited knowledge or exploration of the function space. %
Explicit modeling of epistemic uncertainty typically requires ensembles or Bayesian approaches, as demonstrated in recent work~\cite{Guilhoto2024NEON}. %
Our method primarily captures aleatoric uncertainty; %
for problems involving Bayesian optimization (Section~\ref{sec:bo}) and active learning (Section~\ref{sec:AL_PDEs}), we leverage techniques from reinforcement learning to help supplement the exploration aspect of learning that would ordinarily be handled by pure epistemic uncertainty models. %
This strategy retains the efficiency required for practical inference while avoiding the challenge of disentangling the distinct types of uncertainty from a single deterministic model.  %
The work of \cite{valdenegro2022deeper} provides an analysis of different practical approaches for handling the disentanglement problem, but there are currently no mathematically rigorous frameworks for separating aleatoric and epistemic uncertainty, and this remains a key open problem for UQ techniques, as summarized in the conclusion of \cite{hullermeier2021aleatoric}.

\subsection{Precomputation to decrease computational cost}
\label{sec:precomp}

When deploying DeepONet models for real-time applications, it is important to modify the computational  workflow based on the target application. We consider two common use-cases where redundant calculations can be avoided to achieve significantly faster inference speeds. We first consider the construction of light-weight surrogate models for specific PDE solutions under the assumption of fixed input functions. This is followed by an overview of how inference can be performed efficiently on the full spatial domain for arbitrary input functions\revboth{, including a novel placement matrix, which, to our knowledge, has not been previously reported in literature, that enables fast and structured organization of DeepONet outputs for efficient visualization and downstream processing}.

\subsubsection{Precomputing branch weights}
\label{sec:precomputing_branch}

The first use-case we consider involves the construction of light-weight surrogate networks which can be used to approximate the solution function associated with fixed input data $(f,g)$. The optimization required for this type of application is simple, but yields a noticeable improvement in inference speed and helps motivate the more involved optimization steps introduced in Section~\ref{sec:precomputing_trunk}.

In this setting, we note that the branch output $b$ remains constant for any choice of evaluation location $x$, since the branch depends only on the fixed input data $(f,g)$. By precomputing the branch weights, the solution $\widehat{u}(f,g)$ can be evaluated at arbitrary evaluation locations $x$
using just a single forward-pass of the trunk network. %
In this context, the DeepONet framework can be seen as a natural generalization of the conventional function-approximator networks~\cite{chen95}; %
i.e., once the branch weights are fixed, the DeepONet framework yields fast neural network surrogates\footnote{The key practical difference is that function-approximator networks need to be re-trained whenever the input data $(f,g)$ is modified, while DeepONet surrogates can be obtained by simply recomputing the branch weights for the new inputs (requiring just one forward pass of the branch network).}.

These light-weight surrogates also provide access to %
gradient information with very minimal overhead. In particular, since the branch output $b$ is independent of the input coordinate $x$ 
and the basis functions $\varphi_i(x)$ are defined by the neural network architecture of the trunk component, a linear differential operator $\mathcal{L}_x$ can be applied to DeepONet surrogates by computing
\begin{equation*}
\mathcal{L}_x [\widehat{u}(f,g)(x)] \,\, = \,\,  \mathcal{L}_x \left[ \sum\nolimits_{i=1}^N  b_i \cdot \varphi_i(x)  \right]  
\,\, = \,\,  \sum\nolimits_{i=1}^N  b_i \cdot \mathcal{L}_x \left[ \varphi_i(x) \right],
\end{equation*}
where $\mathcal{L}_x\left[ \varphi_i(x) \right]$ is evaluated %
using automatic differentiation through the trunk network.

\subsubsection{Precomputing trunk weights}
\label{sec:precomputing_trunk}

A more common use-case for DeepONet models arises in applications where the input data $(f,g)$ is expected to vary, and we are interested in
evaluating %
solutions quickly at a fixed set of locations throughout the spatial domain. The optimization relevant for this form of inference is primarily based on precomputing the trunk weights. 
To achieve real-time speeds in practice, 
{{it is also necessary to make a few additional adjustments to the inference calculations.}} 

To begin, we review the computational steps involved with evaluating a DeepONet model 
with input data $(f,g)$ at a fixed set of evaluation locations $\{x_j\}_{j\in\mathcal{E}} \subset \overline{\Omega}$ with $|\mathcal{E}|=E$. First, we need to evaluate the trunk at each of the input locations to form the matrix
\begin{equation*}
  T \,\, = \,\,
  \begin{bmatrix}
    \hspace{0.1in} \raisebox{0.025in}{\rule{0.2in}{0.5pt}} & t_1 & \hspace{-0.05in} \raisebox{0.025in}{\rule{0.2in}{0.5pt}} \hspace{0.1in} \,\, \\
    \hspace{0.1in} \vdots & \vdots & \vdots \hspace{0.1in} \,\, \\
    \hspace{0.1in} \raisebox{0.025in}{\rule{0.2in}{0.5pt}} & t_E & \hspace{-0.05in} \raisebox{0.025in}{\rule{0.2in}{0.5pt}} \hspace{0.1in} \,\,
  \end{bmatrix},
 \hspace{0.25in} \mbox{where} \hspace{0.25in} t_j \, = \, \begin{bmatrix}  \, t_{j1}\, , \,\dots\, , \,t_{jN} \,\end{bmatrix}  \, = \, \operatorname{Trunk}(x_j).
\end{equation*}
The branch output vector is then computed for the current input data $(f,g)$, 
and the network prediction $\widehat{u}$ is given in vector form by
\begin{equation*}
  \begin{bmatrix}
    \widehat{u}_1 \\
    \vdots \\
    \widehat{u}_E
  \end{bmatrix}
   = 
  \begin{bmatrix}
    \hspace{0.1in} \raisebox{0.025in}{\rule{0.2in}{0.5pt}} & t_1 & \hspace{-0.05in} \raisebox{0.025in}{\rule{0.2in}{0.5pt}} \hspace{0.1in} \,\, \\
    \hspace{0.1in} \vdots & \vdots & \vdots \hspace{0.1in} \,\, \\
    \hspace{0.1in} \raisebox{0.025in}{\rule{0.2in}{0.5pt}} & t_E & \hspace{-0.05in} \raisebox{0.025in}{\rule{0.2in}{0.5pt}} \hspace{0.1in} \,\,
  \end{bmatrix}
  \begin{bmatrix}
    b_1 \\
    \vdots \\
    b_N
  \end{bmatrix},
  \hspace{0.075in} \mbox{where} \hspace{0.075in} b^T \, = \, [\, b_{1}\, , \,\dots\, , \,b_{N} \,] \, = \, \operatorname{Branch}(f,g).
\end{equation*}

To interpret the vector prediction $\widehat{u}$ as a function on a two-dimensional domain, we must then place each of the entries $\widehat{u}_j = \widehat{u}(x_j)$ in a data structure which specifies the associated position $x_j$ in the spatial domain.
For example, it is often useful to have a discretized, two-dimensional representation of the predicted solution; 
to achieve this, we can specify evaluation locations $\{x_j\}_{j\in\mathcal{E}}$ based on a uniform $R\times R$ grid covering the domain and omit points that fall outside the domain.  %
To efficiently convert the vector prediction into a structured array, we construct a sparse ``placement matrix'', $P$, defined by mapping each index $j\in\mathcal{E}$ to the appropriate array location.  More precisely, the placement matrix maps the $j^{th}$ entry of the vector $\widehat{u}\in\mathbb{R}^E$ to the column and row associated with position $x_j$ in the two-dimensional matrix $\widehat{U} \in \mathbb{R}^{R\times R}$ storing the structured prediction values.  %

\begin{figure}[thbp]
  \centering
  \includegraphics[width=0.75\linewidth]{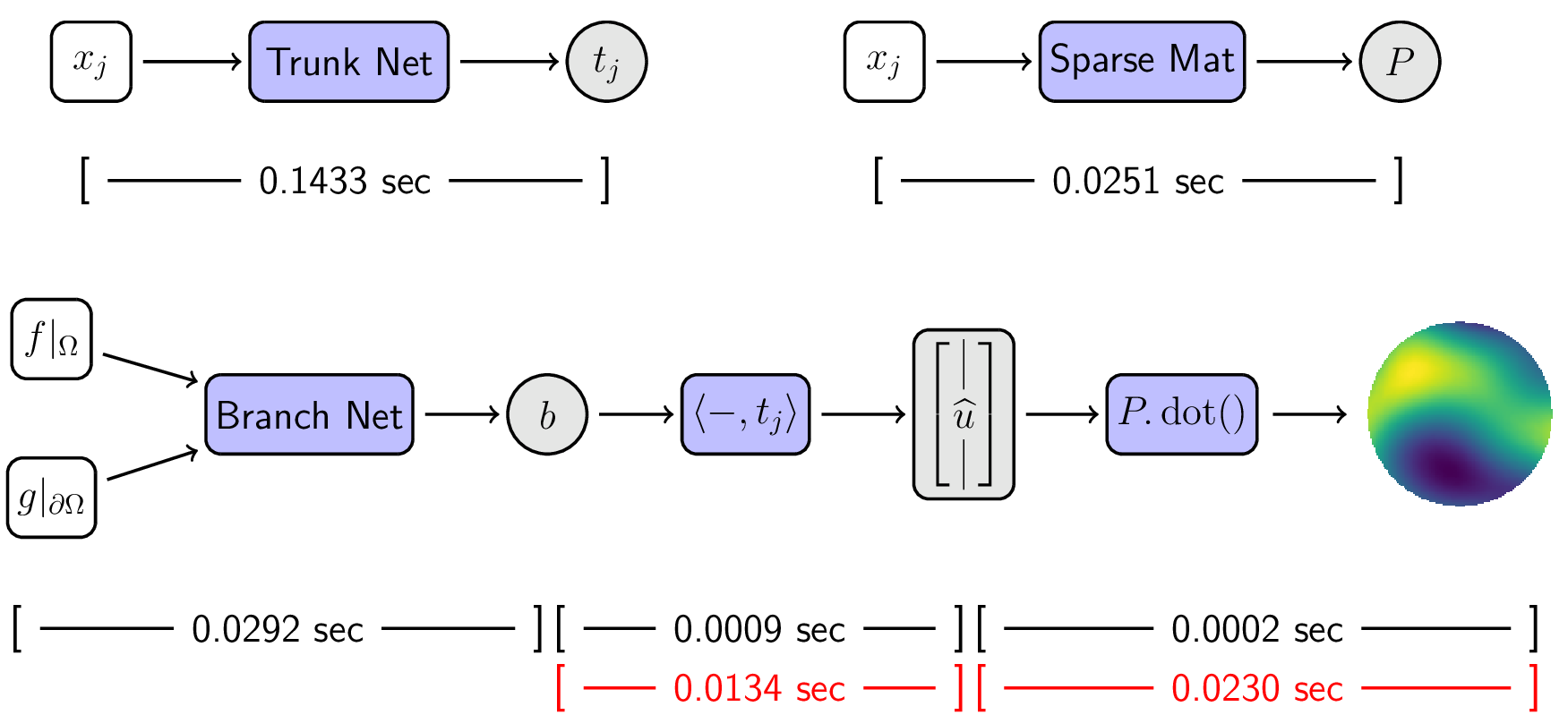}
  \caption[Optimized Inference Overview]{Inference is optimized for fast evaluations on a fixed grid by precomputing the trunk outputs for each grid location and constructing a sparse placement matrix to map vectorized network outputs to the correct array locations.  The timings reported in the top row are one-time computations performed after training. Red timings denote unoptimized speeds using loops and manual entry placement,  while black timings reflect the average runtime per input data pair $(f, g)$ processed in batches.  We note that the speedups reported use NumPy's $\texttt{einsum}$ operations applied to batched inputs and are not representative of single-example evaluations.  The $\log \sigma$ calculation has been omitted from the diagram for brevity.}
\label{fig:inference_1}
\end{figure}

Inference can be performed na\"{i}vely by evaluating both the branch and trunk networks for every input pair $(f, g)$ and evaluation location $x$, but doing so results in computation time ill-suited for real-time applications. 
By fixing evaluation locations and employing the inference strategy above, the inference time is significantly reduced.  %
Once training is complete, the trunk outputs $\{t_j\}_{j\in\mathcal{E}}$ and placement matrix $P$ only need to be computed once to facilitate inference for arbitrary input data $(f,g)$.
With these one-time calculations in hand, the computation time for evaluating the DeepONet solution across the full domain can be reduced to approximately $0.029$ seconds per problem (compared with $0.165$ seconds when computed na\"{i}vely), as shown in Fig.~\ref{fig:inference_1}. 

\subsection{Sampling procedure for randomized data generation}
\label{sec:datagen}

To construct the training datasets for the DeepONet models, we first employ Gaussian process sampling~\cite{rasmussen2003gaussian} to generate random realizations of the solution functions. In particular, we sample
\begin{align*}
  u \, \, \sim \, \, \mathcal{G}(0, k_{l}(x,y))
\end{align*}
from mean-zero Gaussian random fields with squared exponential covariance kernels $k_l(x,y) \, = \, \exp(-\|x-y\|^2/2l^2)$. Training data was generated using the length-scales  $l\in \{0.2000, 0.2333, 0.2667,$ $ 0.3000\}$, and additional test data was generated for length-scales $l\in \{0.1333, 0.1667, 0.3333, 0.3667\}$. %
The test length-scales were intentionally selected outside of the range used to generate training data so that we can assess how well the proposed models generalize to out-of-distribution data.

The associated interior data $f\in C(\Omega)$ and boundary data $g\in C(\partial\Omega)$ are then computed by applying
the forward differential operator and boundary operator directly to the sampled solution. 
This variation on the data generation allows us to construct training datasets without relying on conventional numerical solvers, such as FEM implementations; %
the datasets can be constructed whenever %
implementations of the forward differential operator and boundary operator are available.

\begin{figure}[htbp]
  \centering
  \includegraphics[width=0.85\textwidth]{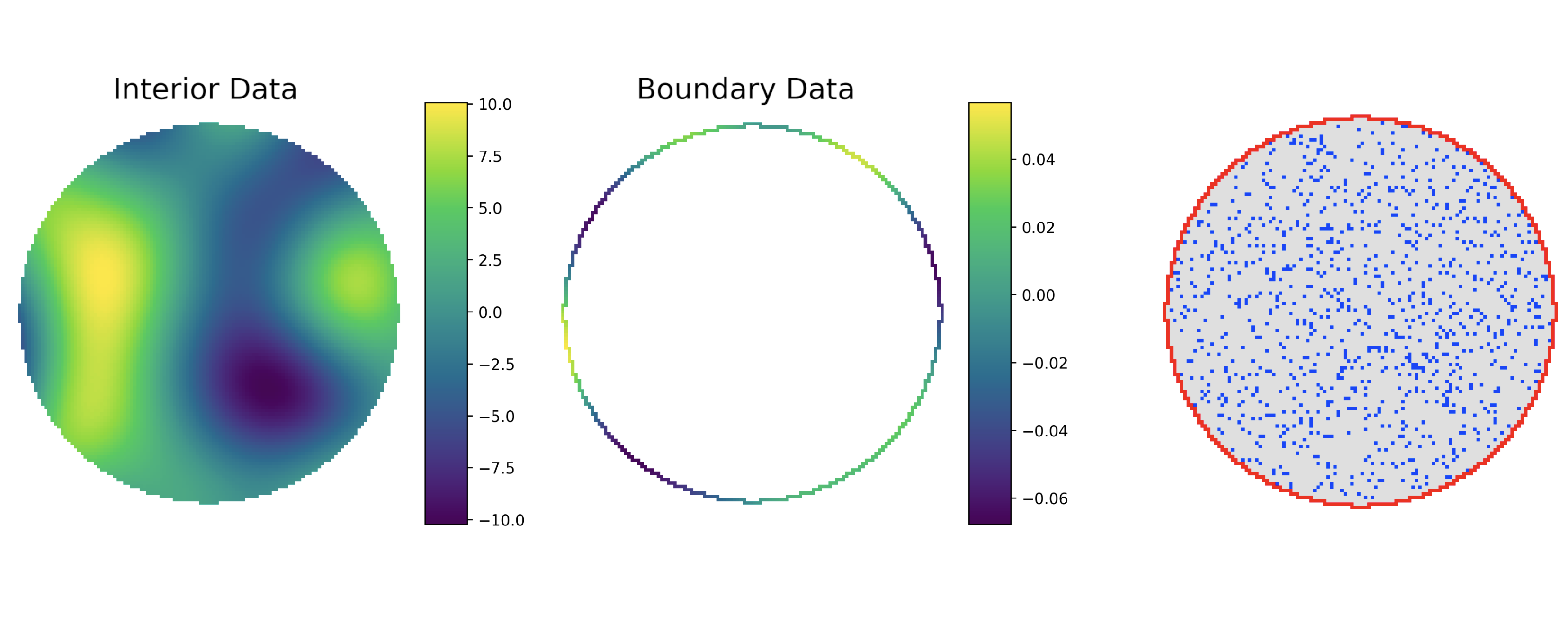}
  \caption[Input data for inhomogeneous boundary conditions.]{Example input data consisting of 2-dimensional interior data and 1-dimensional boundary data.  DeepONet sensor locations for the interior data (blue) and boundary data (red) are shown on the right.}
  \label{fig:input_data}
\end{figure}

The training data was initially generated on uniform spatial grids for simplicity, but this structured form of input data is not required for DeepONet models.  To increase the complexity of the problems, we subsampled an unstructured set of {\emph{sensor locations}} across the domain, $\{s_i\}_{i\in\mathcal{I}} \subset \Omega$, as illustrated on the right of Fig.~\ref{fig:input_data}.  The branch network is provided data from the interior functions at these unstructured sensor locations (along with a complete vector of boundary data at locations $\{s'_j\}_{j\in\mathcal{J}} \subset \partial\Omega$).  Training examples in the final dataset then take the form
\begin{equation*}
  \big(\, x \,, \,\, \{f(s_i)\}_{i\in\mathcal{I}} \,, \,\, \{g(s'_j)\}_{j\in\mathcal{J}} \,, \,\, u(f,g)(x) \,\big)  
\end{equation*}
with multiple evaluation locations $x\in\overline{\Omega}$ randomly sampled for each pair of input functions $(f,g)$.  %
For the experiments in this work, we found it was sufficient to sample $50$ evaluation locations per input function pair. %
We also conducted experiments with several %
different architectures for handling boundary data (as described in \ref{sec:deeponet_boundary_conditions}),
but found that %
simply appending boundary data to the branch network's input vector achieves the same performance as more structured approaches. %

\section{Validation of the proposed framework}
\label{sec:validation}

While the proposed network architecture and training procedure allow DeepONet models to produce uncertainty estimates, %
it is important to verify %
these estimates are unbiased and generalize beyond the scope %
of the training data before employing these models in practice. %
In addition to assessing potential biases in the network predictions, it is natural to consider two distinct forms of generalization when evaluating the proposed training procedure:
\begin{enumerate}
\item[1.] {Performance on unseen validation data sampled from length-scale classes used for training.}
\item[2.] {Performance on testing data sampled outside of the length-scale range seen by the network.} 
\end{enumerate}
The first case corresponds to the network generalization on in-distribution data, while the latter corresponds to the generalization for out-of-distribution (OoD) data~\cite{hendrycks2016baseline}. We first demonstrate that the network predictions are unbiased, and produce accurate predictions, on a set of linear problems.  %
This is followed by a detailed analysis of the predictive uncertainty estimates. %
We then assess the network architecture's generalization capabilities on two nonlinear problem setups.

The numerical results for the linear problems introduced in this section are summarized in  %
Table~\ref{tab:poisson}, and results for the nonlinear setups are summarized in Table~\ref{tab:nonlinear}. %
The length-scales \{0.2000, 0.2333, 0.2667, 0.3000\} correspond to the function spaces observed during training, while the length-scales $\{0.1333, 0.1667, 0.3333, 0.3667\}$ correspond to out-of-distribution data. %
We generated $5000$ function realizations for each training length-scale ({{with $500$ examples reserved for validation}}) along with $1000$ realizations for each test length-scale. %
We note that in general, the DeepONet models are capable of achieving close to a $1\%$ relative error for all problems, with performance worsening as the length-scale is reduced ({{as expected due to the increased data complexity in the low length-scale regime}}).

\newcommand\spacer{\hspace{0.05in}}
\newcommand\altspacer{\hspace{0.05in}}

\begin{table}[htbp]
\centering
\begin{tabular}{lllllllll}
\hline
\multicolumn{9}{l}{\Tstrut\altspacer{\bf{Poisson on Square with homogeneous BC}}\Bstrut} \\
\hline
\hline
\altspacer $l$&\spacer0.1333\spacer&\spacer0.1667\spacer&\spacer0.2000\spacer&\spacer0.2333\spacer&\spacer0.2667\spacer&\spacer0.3000\spacer&\spacer0.3333\spacer&\spacer0.3337 \altspacer \\
\hline 
\hline
\altspacer MSE\spacer&\spacer4.78e-4\spacer&\spacer1.45e-4\spacer&\spacer5.17e-5\spacer&\spacer2.82e-5\spacer&\spacer2.10e-5\spacer&\spacer1.78e-5\spacer&\spacer1.60e-5\spacer&\spacer1.50e-5 \hspace{0.0in}\altspacer \\
\altspacer MAE\spacer&\spacer1.71e-2\spacer&\spacer9.38e-3\spacer&\spacer5.60e-3\spacer&\spacer4.11e-3\spacer&\spacer3.51e-3\spacer&\spacer3.20e-3\spacer&\spacer3.01e-3\spacer&\spacer2.89e-3\altspacer \\
\altspacer $L^1$ Rel\spacer&\spacer3.33e-2\spacer&\spacer1.61e-2\spacer&\spacer8.60e-3\spacer&\spacer5.82e-3\spacer&\spacer4.62e-3\spacer&\spacer3.96e-3\spacer&\spacer3.63e-3\spacer&\spacer3.31e-3\altspacer \\
\altspacer $L^2$ Rel\spacer&\spacer3.19e-2\spacer&\spacer1.56e-2\spacer&\spacer8.36e-3\spacer&\spacer5.69e-3\spacer&\spacer4.55e-3\spacer&\spacer3.93e-3\spacer&\spacer3.62e-3\spacer&\spacer3.32e-3 \hspace{0.0in}\altspacer \\
\hline
\multicolumn{9}{l}{\Tstrut\altspacer{\bf{Poisson on Square with inhomogeneous BC}}\Bstrut} \\
\hline
\altspacer MSE\spacer&\spacer1.94e-2\spacer&\spacer4.17e-3\spacer&\spacer7.90e-4\spacer&\spacer1.67e-4\spacer&\spacer5.46e-5\spacer&\spacer3.22e-5\spacer&\spacer9.35e-5\spacer&\spacer8.83e-5\\
\altspacer MAE\spacer&\spacer8.78e-2\spacer&\spacer3.80e-2\spacer&\spacer1.59e-2\spacer&\spacer7.21e-3\spacer&\spacer4.09e-3\spacer&\spacer2.95e-3\spacer&\spacer2.91e-3\spacer&\spacer2.63e-3\\
\altspacer $L^1$ Rel\spacer&\spacer2.25e-1\spacer&\spacer9.87e-2\spacer&\spacer4.19e-2\spacer&\spacer1.92e-2\spacer&\spacer1.10e-2\spacer&\spacer8.04e-3\spacer&\spacer7.67e-3\spacer&\spacer6.91e-3\\
\altspacer $L^2$ Rel\spacer&\spacer2.83e-1\spacer&\spacer1.32e-1\spacer&\spacer5.88e-2\spacer&\spacer2.76e-2\spacer&\spacer1.59e-2\spacer&\spacer1.15e-2\spacer&\spacer1.07e-2\spacer&\spacer9.66e-3\\
\hline
\multicolumn{9}{l}{\Tstrut\altspacer{\bf{Poisson on Circle with inhomogeneous BC}}\Bstrut} \\
\hline
\altspacer MSE\spacer&\spacer3.85e-3\spacer&\spacer5.23e-4\spacer&\spacer7.04e-5\spacer&\spacer1.43e-5\spacer&\spacer6.45e-6\spacer&\spacer3.08e-6\spacer&\spacer3.69e-6\spacer&\spacer3.66e-6\\
\altspacer MAE\spacer&\spacer4.63e-2\spacer&\spacer1.66e-2\spacer&\spacer6.07e-3\spacer&\spacer2.70e-3\spacer&\spacer1.63e-3\spacer&\spacer1.22e-3\spacer&\spacer1.07e-3\spacer&\spacer9.37e-4\\
\altspacer $L^1$ Rel\spacer&\spacer1.20e-1\spacer&\spacer4.40e-2\spacer&\spacer1.64e-2\spacer&\spacer7.43e-3\spacer&\spacer4.51e-3\spacer&\spacer3.42e-3\spacer&\spacer3.06e-3\spacer&\spacer2.60e-3\\
\altspacer $L^2$ Rel\spacer&\spacer1.28e-1\spacer&\spacer4.79e-2\spacer&\spacer1.83e-2\spacer&\spacer8.30e-3\spacer&\spacer4.97e-3\spacer&\spacer3.75e-3\spacer&\spacer3.35e-3\spacer&\spacer2.86e-3\\
\hline

\end{tabular} 
\caption[Results for Linear Poisson on Square and Circle]{{Numerical results for the linear problem setups across each length-scale class. Mean squared errors (MSE), mean average errors (MAE), and $L^1/L^2$ relative errors are reported for each length scale, shown in columns.}} 
\label{tab:poisson}
\end{table}

\subsection{Poisson Equations with Homogeneous and Inhomogeneous BCs}

The first problem we consider is the Poisson equation on a square domain with homogeneous boundary conditions, as described by Eq.~\eqref{eq:pois-homo}. This is a simple problem, and if the proposed model performs poorly in this setting, %
we can conclude that the proposed addition of UQ is negatively impacting accuracy. %
\revone{The models were trained using the Adam optimization algorithm~\cite{kingma2014adam}, with a learning rate of $7.5e\mathrm{-}4$, a batch size of $512$, and exponential learning rate decay by a factor of $0.95$ every $25$ thousand learning steps.  Training was performed for $2$ million steps, and this setup was used to train all models evaluated in this section.} 
Overall, the network predictions for this problem setup are accurate, as shown in Table~\ref{tab:poisson}, with relative errors below $1\%$.  %
This allows us to move on to more complex problems with confidence that %
the UQ framework is not undermining the accuracy of network predictions. \\

\begin{subequations}\label{eq:poisson_equations}
  \noindent
    \begin{minipage}{0.05\textwidth}\centering
      \hspace{0.05in}
    \end{minipage}%
  \begin{minipage}{0.4\textwidth}
\begin{equation}
  \hspace{0.2in}\vspace{0.0in}\begin{cases}  \hspace{0.05in} \Delta u \, = \, f & \hspace{0.0in} \mbox{in} \hspace{0.2in} \Omega  \\
    \hspace{0.16in} u \, = \, 0 & \hspace{0.0in} \mbox{on} \hspace{0.1in} \partial\Omega   \end{cases}
 \label{eq:pois-homo}
\end{equation}
    \end{minipage}%
    \begin{minipage}{0.05\textwidth}\centering
      \hspace{0.05in}
    \end{minipage}%
    \begin{minipage}{0.4\textwidth}
\begin{equation}
    \hspace{-0.1in}\vspace{0.0in} \begin{cases}  \hspace{0.05in} \Delta u \, = \, f & \hspace{0.0in} \mbox{in} \hspace{0.2in} \Omega  \\   \hspace{0.16in} u \, = \, g & \hspace{0.0in} \mbox{on} \hspace{0.1in} \partial\Omega   \end{cases}
    \label{eq:pois-inhomo}
\end{equation}
    \end{minipage}
    \begin{minipage}{0.075\textwidth}\centering
      \hspace{0.05in}
    \end{minipage}%
    \vskip1.25em
\end{subequations}

\begin{figure}[thbp]
  \centering
  \begin{tabular}{c}
  \includegraphics[width=0.95\textwidth]{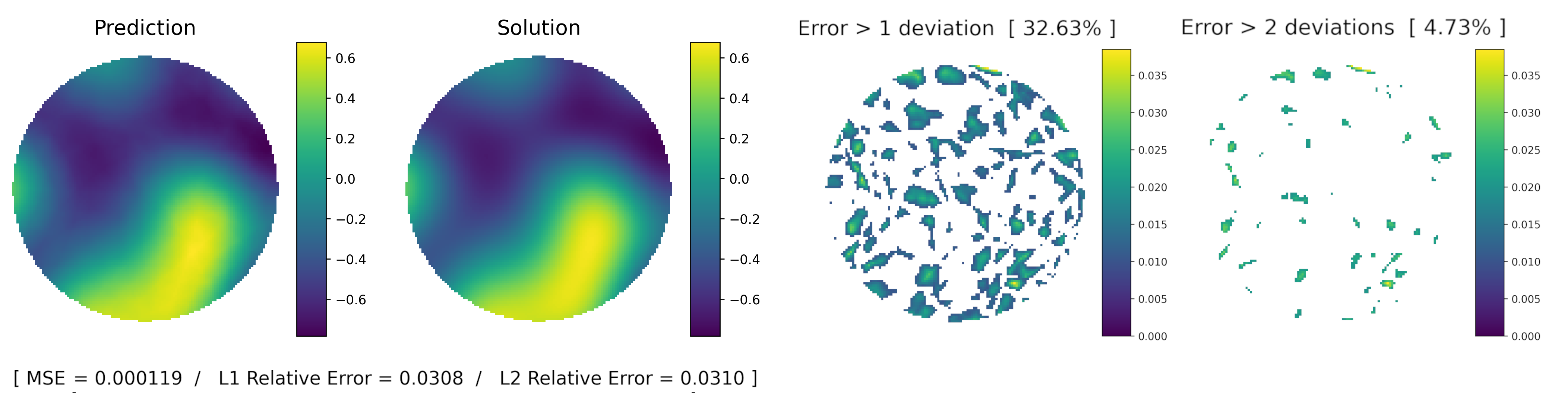} \\
  \includegraphics[width=0.95\textwidth]{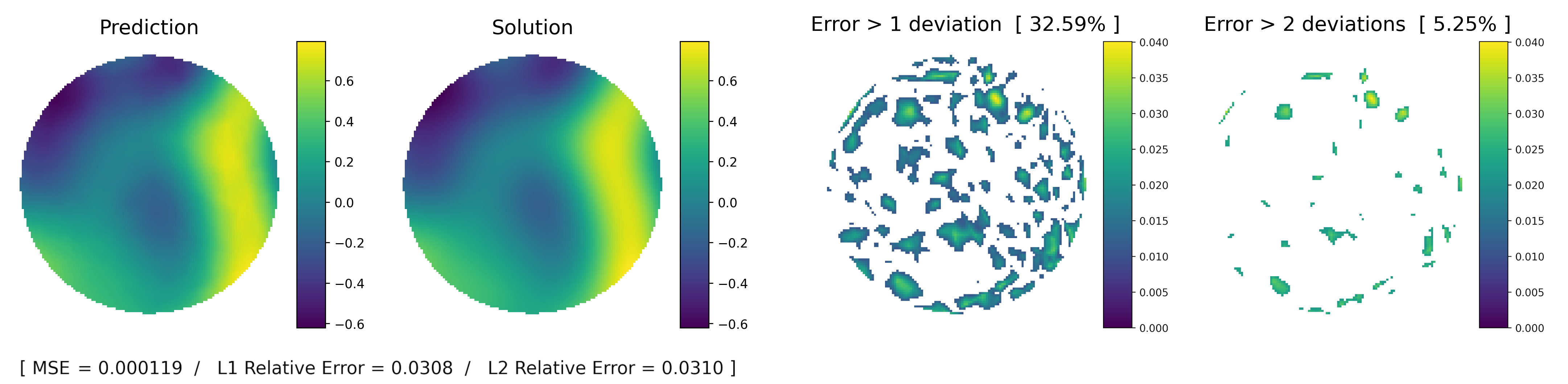}
  \end{tabular}
  \caption[Qualitative results for the Inhomogeneous Poisson setup]{Qualitative results for two examples of the Inhomogeneous Poisson setup show that 
    predictions provide accurate approximations to the true solutions.  The network's predictive uncertainty closely matches the empirical 68-95-99.7 rule, with around 32\% of the observed errors above 1 standard deviation and around 5\% above 2 deviations. } 
  \label{fig:circle_results}
\end{figure}

We next consider the Poisson equation with inhomogeneous boundary conditions in Eq.~\eqref{eq:pois-inhomo} for both a square and circular domain. The errors are comparable for the square and circular domains, but overall the circular domain yields lower errors than the square domain.  %
This difference is depicted in Table~\ref{tab:poisson}, 
and a qualitative assessment of the network results is provided in Fig.~\ref{fig:circle_results}. 
As an initial assessment of the predictive uncertainties, we note that the percentages of the dataset where errors exceed the predicted standard deviations align well with the empirical 68-95-99.7 rule for normally distributed data.  This suggests the predictive uncertainty may capture the model errors correctly and motivates a more thorough analysis of the uncertainty results.

\begin{table}[htbp]
\centering
\begin{tabular}{lllllllll}
\hline
\multicolumn{9}{l}{\Tstrut\altspacer{\bf{Nonlinear Poisson with inhomogeneous BC}}\Bstrut} \\
\hline
\hline
\altspacer $l$&\spacer0.1333\spacer&\spacer0.1667\spacer&\spacer0.2000\spacer&\spacer0.2333\spacer&\spacer0.2667\spacer&\spacer0.3000\spacer&\spacer0.3333\spacer&\spacer0.3337 \altspacer \\
\hline 
\hline
\altspacer MSE\spacer&\spacer4.46e-3\spacer&\spacer4.79e-4\spacer&\spacer6.71e-5\spacer&\spacer1.64e-5\spacer&\spacer9.30e-6\spacer&\spacer5.92e-6\spacer&\spacer1.97e-5\spacer&\spacer1.70e-5 \altspacer \\
\altspacer MAE\spacer&\spacer4.95e-2\spacer&\spacer1.67e-2\spacer&\spacer6.24e-3\spacer&\spacer3.01e-3\spacer&\spacer2.00e-3\spacer&\spacer1.58e-3\spacer&\spacer1.54e-3\spacer&\spacer1.41e-3 \altspacer \\
\altspacer $L^1$ Rel\spacer&\spacer1.29e-1\spacer&\spacer4.43e-2\spacer&\spacer1.67e-2\spacer&\spacer8.07e-3\spacer&\spacer5.31e-3\spacer&\spacer4.21e-3\spacer&\spacer3.94e-3\spacer&\spacer3.66e-3 \altspacer \\
\altspacer $L^2$ Rel\spacer&\spacer1.32e-1\spacer&\spacer4.61e-2\spacer&\spacer1.77e-2\spacer&\spacer8.58e-3\spacer&\spacer5.64e-3\spacer&\spacer4.48e-3\spacer&\spacer4.20e-3\spacer&\spacer3.89e-3 \altspacer \\
\hline
\multicolumn{9}{l}{\Tstrut\altspacer{\bf{Diffusion-Reaction with inhomogeneous BC}}\Bstrut} \\
\hline
\altspacer MSE\spacer&\spacer6.87e-3\spacer&\spacer1.28e-3\spacer&\spacer2.70e-4\spacer&\spacer8.83e-5\spacer&\spacer6.65e-5\spacer&\spacer5.77e-5\spacer&\spacer4.87e-5\spacer&\spacer5.67e-5 \altspacer \\
\altspacer MAE\spacer&\spacer6.50e-2\spacer&\spacer2.77e-2\spacer&\spacer1.22e-2\spacer&\spacer6.48e-3\spacer&\spacer4.57e-3\spacer&\spacer3.89e-3\spacer&\spacer3.61e-3\spacer&\spacer3.51e-3 \altspacer \\
\altspacer $L^1$ Rel\spacer&\spacer1.69e-1\spacer&\spacer7.27e-2\spacer&\spacer3.24e-2\spacer&\spacer1.69e-2\spacer&\spacer1.17e-2\spacer&\spacer9.85e-3\spacer&\spacer9.26e-3\spacer&\spacer8.97e-3 \altspacer \\
\altspacer $L^2$ Rel\spacer&\spacer1.71e-1\spacer&\spacer7.47e-2\spacer&\spacer3.37e-2\spacer&\spacer1.77e-2\spacer&\spacer1.23e-2\spacer&\spacer1.04e-2\spacer&\spacer9.84e-3\spacer&\spacer9.55e-3 \altspacer \\
\hline
\end{tabular} 
\caption[Results for Nonlinear Problem Setups]{{Numerical results for the nonlinear problem setups across each length-scale class.}}
\label{tab:nonlinear}
\end{table}

\subsection{Assessment of predictive uncertainty framework}

\begin{figure}[thb]
  \centering

  \begin{minipage}[t]{\textwidth}
    \centering
    \begin{subfigure}[t]{0.35\textwidth}
      \includegraphics[width=\textwidth]{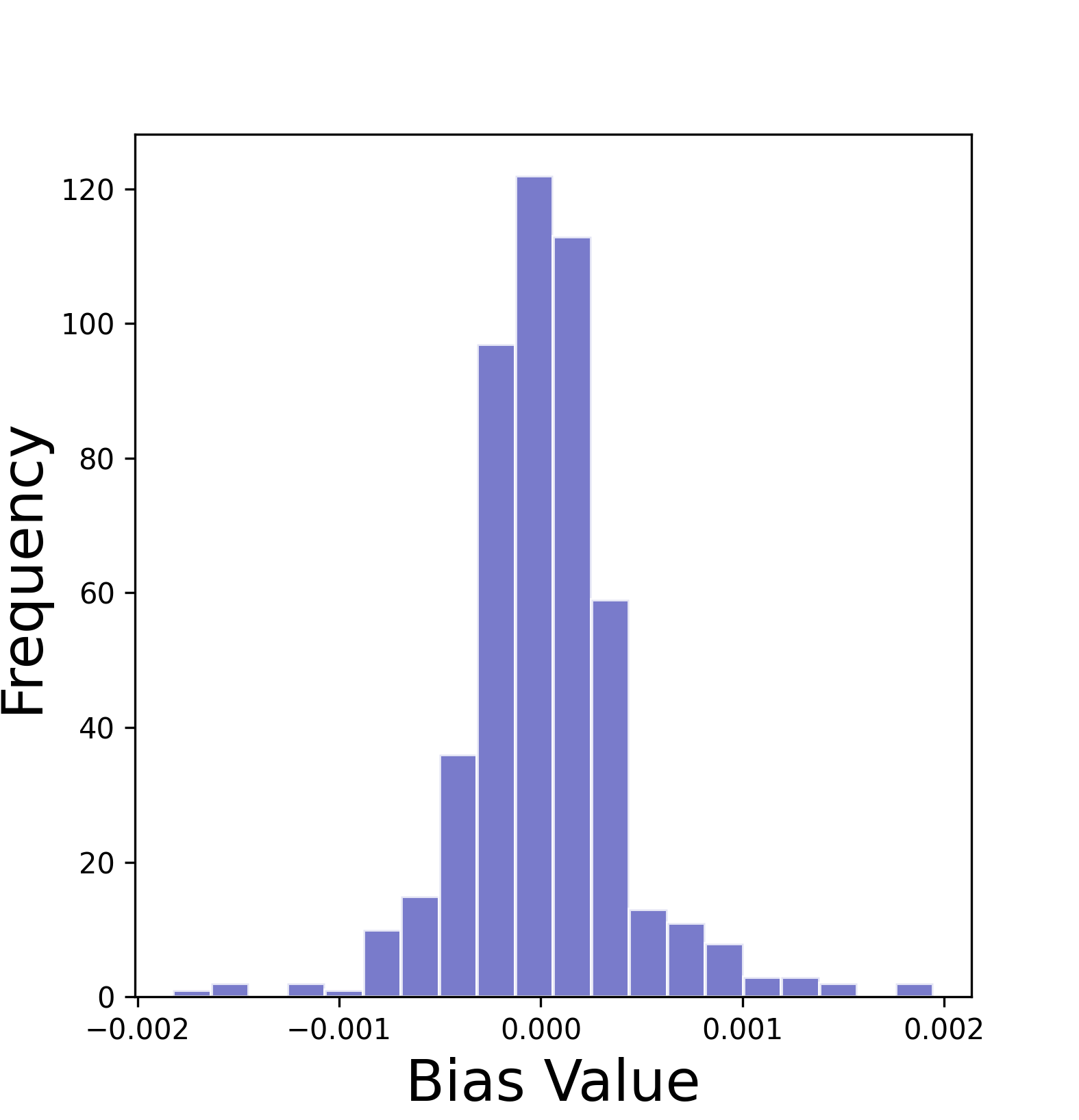}
    \end{subfigure}
    \hspace{-0.26in}
    \begin{subfigure}[t]{0.35\textwidth}
      \includegraphics[width=\textwidth]{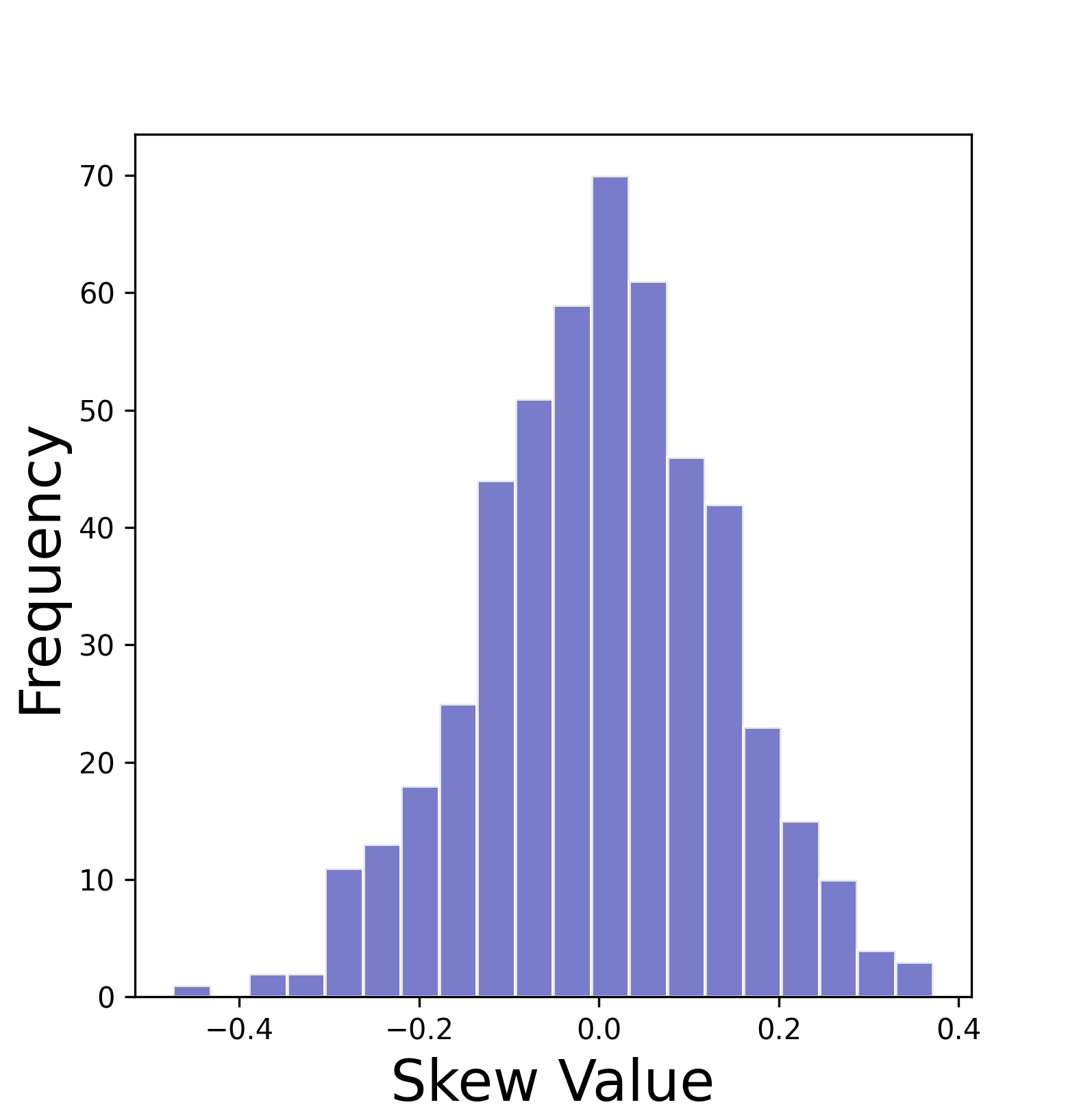}
    \end{subfigure}
    \hspace{-0.26in}
    \begin{subfigure}[t]{0.35\textwidth}
      \includegraphics[width=\textwidth]{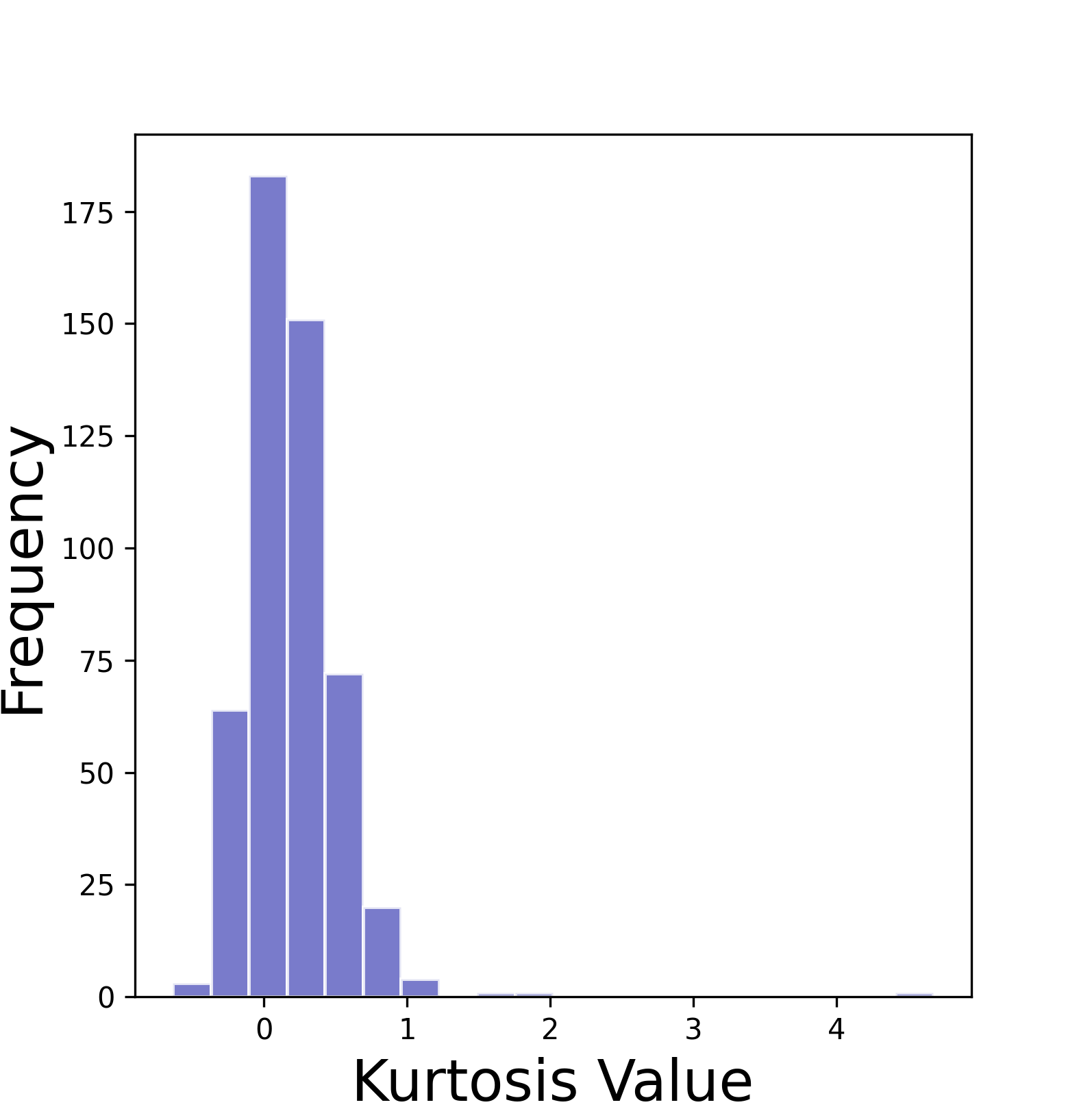}
    \end{subfigure}
  \end{minipage}

  \begin{minipage}[t]{\textwidth}
    \centering
    \begin{subfigure}[t]{0.45\textwidth}
      \includegraphics[width=\textwidth]{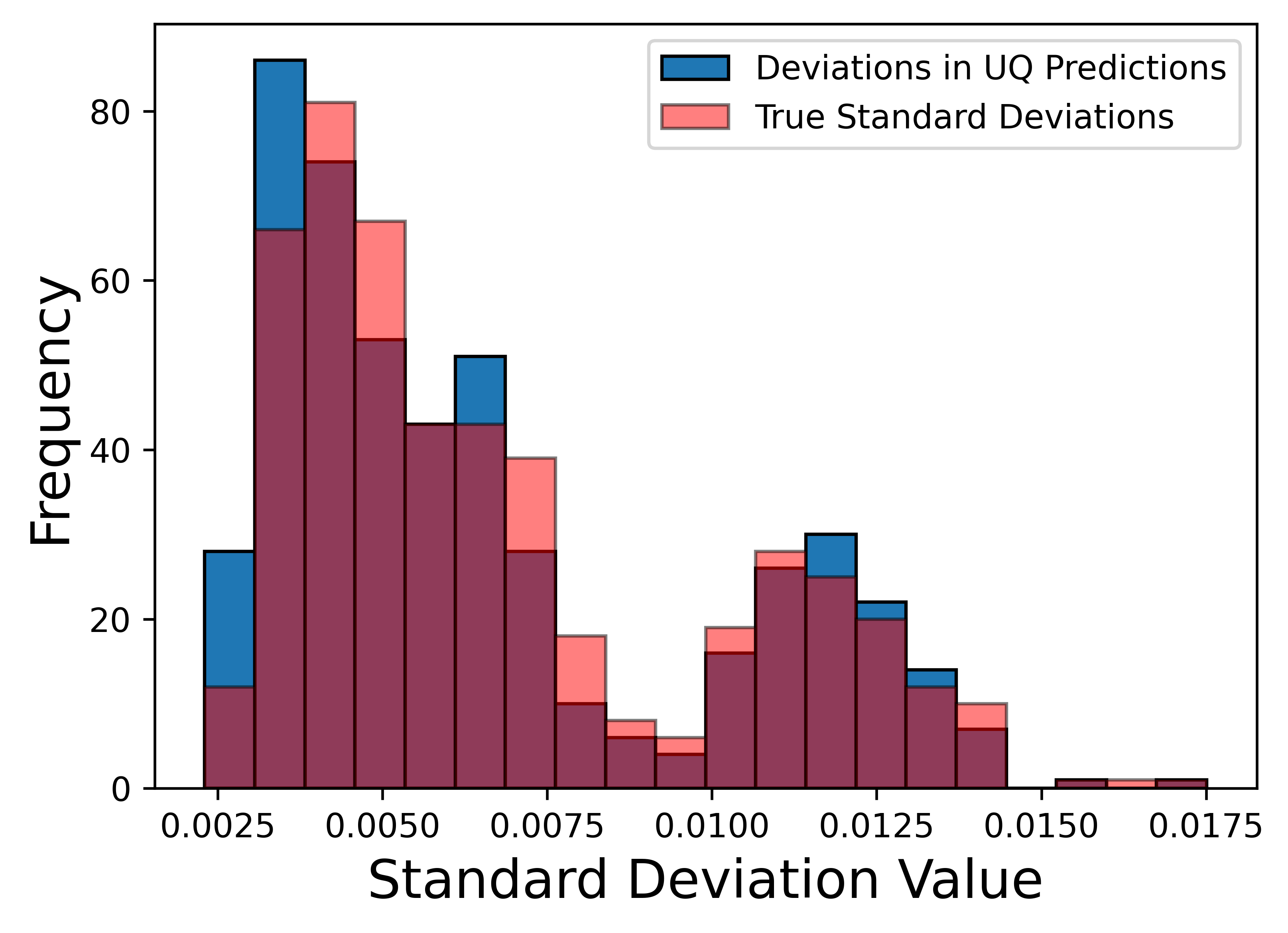}
    \end{subfigure}
    \hspace{0.05in}
    \begin{subfigure}[t]{0.45\textwidth}
      \includegraphics[width=\textwidth]{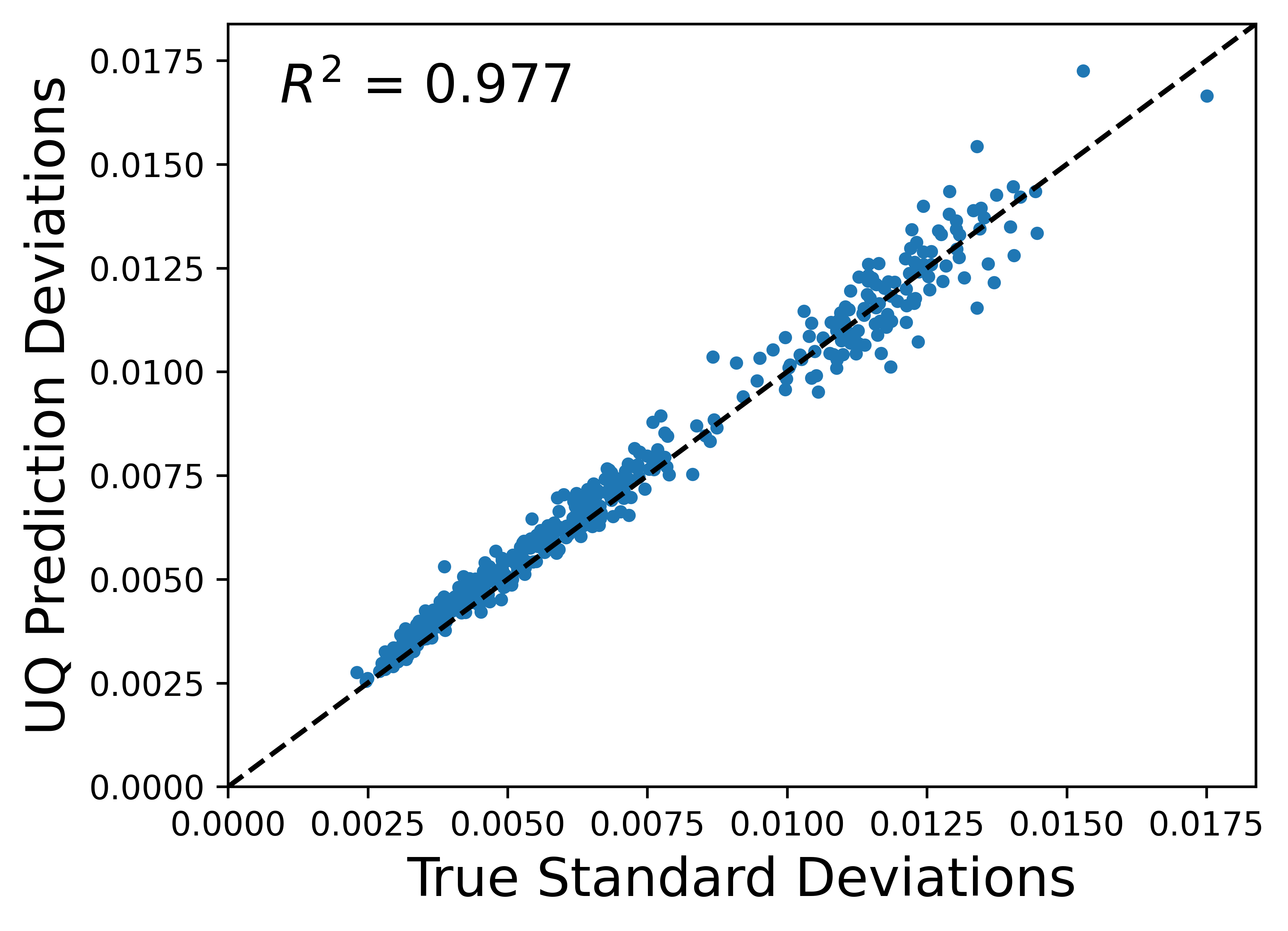}
    \end{subfigure}
  \end{minipage}

  \caption{Biases (top-left), skews (top-center), and kurtosis (top-right) of the observed error distributions at the tail-end of training for $500$ validation examples from the Inhomogeneous Poisson equation over a circular domain. We show the observed standard deviations (bottom-left) in errors at the end of training (blue) and the network's predicted uncertainties (pink) for 500 examples from the Inhomogeneous Poisson; the significant overlap (maroon) provides evidence that the network uncertainty is accurately capturing the model errors. \revtwo{ There is strong agreement between the true and predicted standard deviations with an $R^2$ of 0.977 (bottom-right).}}
  \label{fig:uq_summary}
\end{figure}

To evaluate the proposed uncertainty framework, we assess the model predictions for $500$ validation examples on the circular domain for the inhomogeneous Poisson equation. %
We evaluate the predicted uncertainties with respect to the distribution of observed model errors (see  Fig.~\ref{fig:pred_uq_fit_new}), %
and assess the overall distribution of predicted uncertainties compared with the observed standard deviations in the network errors at the tail-end of training. 
We also analyze the bias, skew, and kurtosis of the distributions of observed errors for each of the $500$ validation examples.  This analysis is used to verify that normal distributions are suitable candidates for modeling the observed error;  histograms summarizing these results are provided in Fig.~\ref{fig:uq_summary}.

The biases of the error distributions at the tail-end of training are consistently small, with values ranging from $-0.0018$ to $0.0019$. %
The histogram of biases is also symmetric and roughly centered at zero, %
indicating that the network does not have any tendency %
to over-estimate or under-estimate function values consistently across the dataset. %
The skew values for the observed error distributions are also approximately symmetric and centered at zero, %
with a minimum skew of $-0.4739$ and maximum skew of $0.3715$.  The majority of error distributions have skews with magnitudes below $0.2$, indicating the distributions are roughly symmetric.  Since we elected to use normal distributions to model uncertainty, it is important to verify %
the observed error distributions are sufficiently symmetric.  If the observed errors have skewed distributions, the loss function in Eq.~\eqref{eq:prob_loss} can be replaced with the negative log-likelihood for a different class of distributions (see Appendix B of \cite{winovich2019convpde}). 


A similar analysis of the kurtosis of the observed errors 
shows a generally positive and skewed right distribution. %
Kurtosis measures the ``tailedness'' of a distribution, with a higher value indicating more outliers~\cite{kline2023principles}.  %
Most of the kurtosis values are close to zero with a slight bias in the positive direction. %
While this means they will generally have slightly longer tails and sharper peaks, all kurtosis values are well within the standard {{reference interval, $[-2,2]$, for a normal distribution}}.   %

Finally, we consider how the predicted errors stand up to the true errors. %
The network provides standard deviation estimates which we compare with the true standard deviations of the error distributions observed at the tail-end of training. %
We evaluated the predicted and true standard deviations across $500$ validation examples and found the average relative error to be $0.0718$. %
The histogram on the right of  Fig.~\ref{fig:uq_summary} shows how the distribution of predicted uncertainties compares with the observed standard deviations.  There is a considerable overlap between the histograms (shown in maroon), and we see that the overall distribution is accurately captured by the UQ procedure. %
This provides evidence that the predictive uncertainty framework is operating correctly and can accurately predict the variance in the network predictions. %
Our assessments of the predictive uncertainty for the other problem setups were consistent with the analysis presented above. %

\subsection{Nonlinear Poisson Equation and Diffusion-Reaction Equation}

\begin{figure}[bhtp]
  \centering
  \begin{tabular}{c}
    \includegraphics[width=0.9\textwidth]{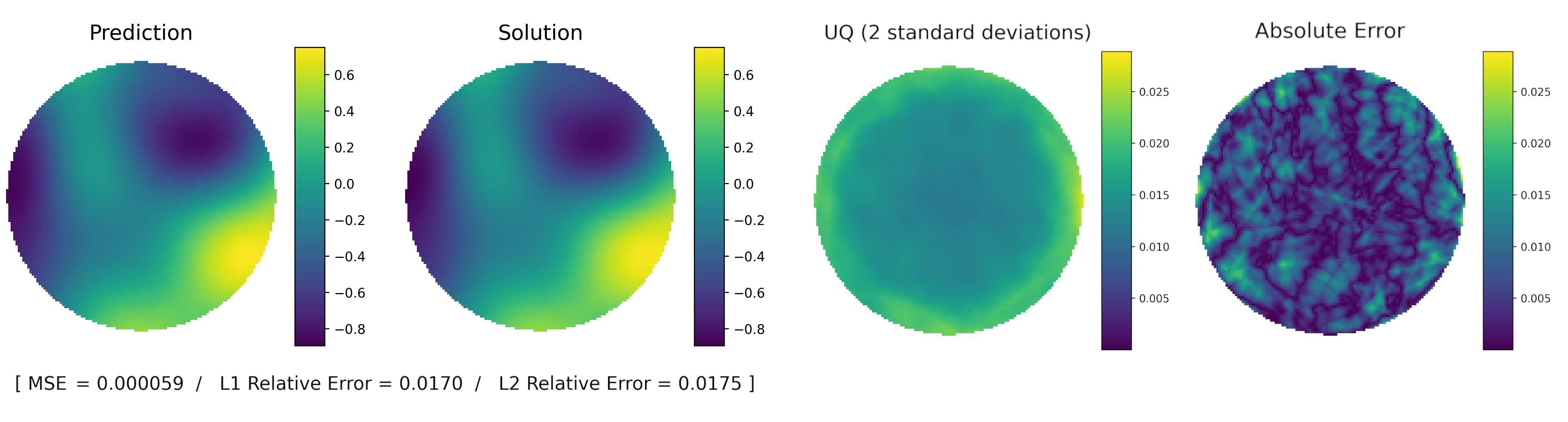} \\
    \includegraphics[width=0.9\textwidth]{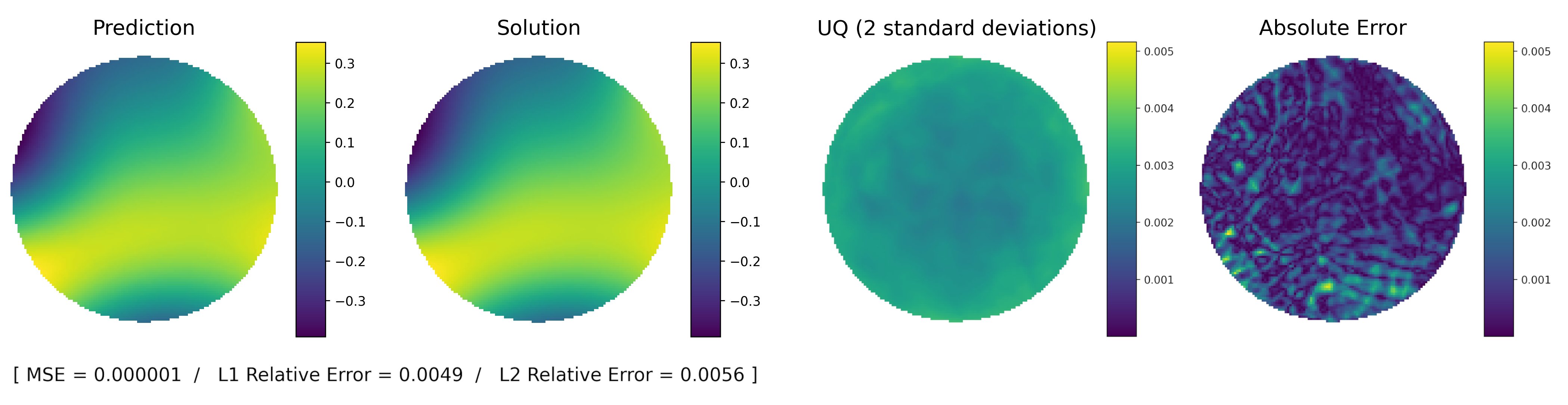}
  \end{tabular}
  \caption[Qualitative results for Nonlinear Poisson setup]{DeepONet results on validation dataset for Nonlinear Poisson equation on unit circle with predicted and true solutions shown on left. The observed error profiles (right) are seen to be unstructured and resemble random noise.  The network's predictive uncertainty is relatively flat across the domain and accurately captures the scale of the network errors. While the predictive uncertainty does not capture the precise locations of errors, the percentages falling within the uncertainty bounds still align well with the empirical rule, similar to the results shown in Fig.~\ref{fig:circle_results}.  }
  \label{fig:deeponet_nonlinear_predictions}  
\end{figure}

Nonlinearity adds additional complexity in traditional solvers and provides a more challenging benchmark for evaluating DeepONet models. %
For our first nonlinear problem setup, we consider the nonlinear Poisson formulation described in Eq.~\eqref{eq:non-linearpoisson}.  A summary of the network prediction errors across different length-scales is provided in Table~\ref{tab:nonlinear}. %
The predictions for this setup are also accurate, with errors comparable to the linear setups, %
and examples of the DeepONet predictions are shown in Fig.~\ref{fig:deeponet_nonlinear_predictions}. %
We see that the predictions accurately approximate the true solutions and yield noisy, unstructured error profiles with maximum values on the same order as the predictive uncertainties. %

\begin{subequations}\label{eq:nonlinear_equations}
  \noindent
    \begin{minipage}{0.025\textwidth}\centering
      \hspace{0.025in}
    \end{minipage}%
  \begin{minipage}{0.475\textwidth}
\begin{equation}
  \vspace{0.0in} \begin{cases} \hspace{0.05in} \operatorname{div}\big( (1 + u^2) \, \nabla u  \big) \, = \, f & \hspace{0.0in} \mbox{in} \hspace{0.2in} \Omega  \\   \hspace{0.95in} u \, = \, g  & \hspace{0.0in} \mbox{on} \hspace{0.1in} \partial\Omega  \end{cases}
\label{eq:non-linearpoisson}
\end{equation}
    \end{minipage}%
    \begin{minipage}{0.045\textwidth}\centering
      \hspace{0.025in}
    \end{minipage}%
    \begin{minipage}{0.4\textwidth}
\begin{equation}
  \vspace{0.0in}   \begin{cases} \hspace{0.05in} \Delta u \, + \, u^2 \, = \, f & \hspace{0.0in} \mbox{in} \hspace{0.2in} \Omega  \\    \hspace{0.49in} u \, = \, g  & \hspace{0.0in} \mbox{on} \hspace{0.1in} \partial\Omega  \end{cases}
\label{eq:diffreac}
\end{equation}
    \end{minipage}
    \begin{minipage}{0.025\textwidth}\centering
      \hspace{0.025in}
    \end{minipage}%
    \vskip1.25em
\end{subequations}

The final problem setup we consider is a nonlinear diffusion-reaction prescribed by Eq.~\eqref{eq:diffreac}. %
A summary of the accuracies of the DeepONet model for each length-scale is provided in Table~\ref{tab:nonlinear}. %
Overall, the prediction errors are slightly higher than those encountered in previous setups, %
likely stemming from the increased complexity of the system. %
We also note that the larger relative errors are more pronounced in the smaller length-scales, especially the length-scales $0.1333$ and $0.1667$, which were withheld during training. %
Since the errors are higher for this setup, it is natural to ask whether the predictive uncertainty correctly reflects the change in accuracy.  %
Ideally, we would also like the uncertainties to provide an indication of the drop in performance at lower length-scales. %

\subsection{Generalization to Out-of-Distribution Data}
\label{subsec:generalization_extrapolation}

\begin{figure}[htbp]
  \centering
  \includegraphics[width=0.999\textwidth]{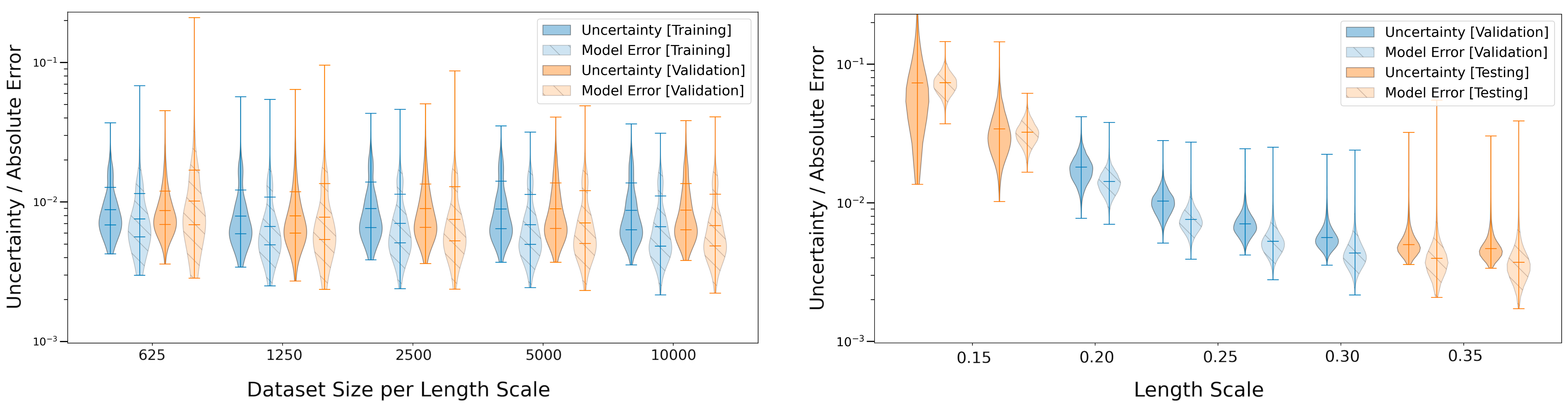}
  \caption[Comparison of DeepONet performance on training and validation datasets.]{Comparison of DeepONet performance on training and validation examples for different training dataset sizes for the Nonlinear Diffusion-Reaction problem setup (left). The network converges well even with limited training data, but the performance is also observed to level off quickly as the number of training examples increased indicating data saturation and a possible limit to the network's accuracy. %
    To the right, we provide an analysis of predictive uncertainty produced by a DeepONet model on the Nonlinear Diffusion-Reaction problem setup.  When evaluated on in-distribution validation data (blue) as well as OoD test data (orange) the network's uncertainty is seen to accurately capture the observed model error. %
  }
\label{fig:train_test_analysis_new}
\end{figure}

To assess the proposed uncertainty framework's generalization capabilities, we evaluate the model's performance on unseen in-distribution and OoD data. %
In the presence of sufficient training data, the prediction accuracy of the DeepONet models on in-distribution validation data matches the accuracy on training examples across all length-scales. The associated uncertainty estimates are also well-calibrated to the validation %
errors when sufficient training data is available, as shown in Fig.~\ref{fig:train_test_analysis_new}. %

Of greater interest is the fact that the predictive uncertainties generalize quite well to out-of-distribution data, as shown in the second plot of Fig.~\ref{fig:train_test_analysis_new}. As noted earlier, the lower length-scale classes correspond to more difficult problems; accordingly, the network performance gradually decays as the length-scale is reduced. Notably, the predictive uncertainties for the smallest length-scales are also observed to increase, reflecting the rising network errors associated with the out-of-distribution data. However, this rise in uncertainty does not occur for the out-of-distribution at the other extreme (i.e., length-scales larger than those seen during training), which correctly reflects the model performance for these simpler problems.

This shows that the network can, to some extent, automatically determine the difficulty associated with a given problem. %
More precisely, the accuracy of the network's predictive uncertainties extends beyond the scope of the function classes observed during training. %
The network is never provided any labeling information regarding the length-scale values, so the network's predictive uncertainty is based solely on features extracted from the raw input function values. %

\section{Accelerating \revone{Bayesian optimization} with predictive uncertainties}
\label{sec:bo}

In this section, we demonstrate how the proposed uncertainty framework can be used to accelerate outer-loop \revone{BO} procedures. %
\revtwo{As a proof-of-concept problem,} %
we consider a \revtwo{visually intuitive nonlinear} system of time-dependent functions parameterized by a vector $\theta=(\alpha, \beta)$ with $\alpha \in [-7,4]$ and $\beta \in [0,6]$. %
The system consists of an acceleration $f_\theta(t)$, velocity $u_\theta(t)$, and final position $q_\theta$ defined by
\begin{eqnarray*}
f_\theta(t) \, &=&  -(1+\alpha)\cdot\alpha \cdot\sin(\alpha t) \,- \, \cos^2(\beta) \cdot e^{t \cos(\beta)},  \\ 
u_\theta(t) \, &=& \hspace{0.12in}  (1+\alpha)\cdot\cos(\alpha t) \, - \, \cos(\beta) \cdot e^{t \cos(\beta)},  \\  
q_\theta\, &=&   \int_0^1 u_\theta(t) \, dt \,\,\, = \,\,\frac{1+\alpha}{\alpha} \,\sin(\alpha) \,-\, e^{\cos(\beta)}.
\end{eqnarray*}

Our goal is to identify the parameters $\theta = (\alpha, \beta)$, along with the corresponding velocity function $u_\theta(t)$, that maximize the final position of the system, $q_\theta$. %
We consider two formulations of this problem, 
with both leveraging DeepONet models to approximate the system's velocity function based on acceleration information. %
In the first formulation, we provide the DeepONet model with the underlying system parameters $(\alpha, \beta)$ as input; we refer to this as the {\emph{parameter-to-function}} setup. %
For the second formulation, we provide the model with time-series data containing acceleration information; we refer to this as the {\emph{function-to-function}} setup. %
In both cases, the network is trained to predict the velocity function $u_{\theta}(t)$, along with the associated uncertainty estimates. %
We then derive the quantity of interest (QoI) from the predicted velocity using the quadrature rule
\begin{equation*}
\widehat{q}_\theta  
\,\, = \,\,
\int_0^1 \widehat{u}(t) \, dt  \,\,\, \approx \,\,\, \frac{1}{2n} \left( \widehat{u}(t_0) \,+\, \widehat{u}(t_n) \, + \, 2\, \sum\nolimits_{i=1}^{n-1} \,\widehat{u}(t_i)\, \right)
\end{equation*}
This quadrature rule is also applied to the predictive uncertainties in the velocity function, allowing us to derive an implied uncertainty, $\sigma(\widehat{q}_\theta)$, on the quantity of interest $\widehat{q}_\theta$. %

\begin{figure}[htbp]
  \centering
  \begin{tabular}{cc}
    \begin{tabular}{cc}
    \hspace{-0.125in}\includegraphics[width=0.255\textwidth]{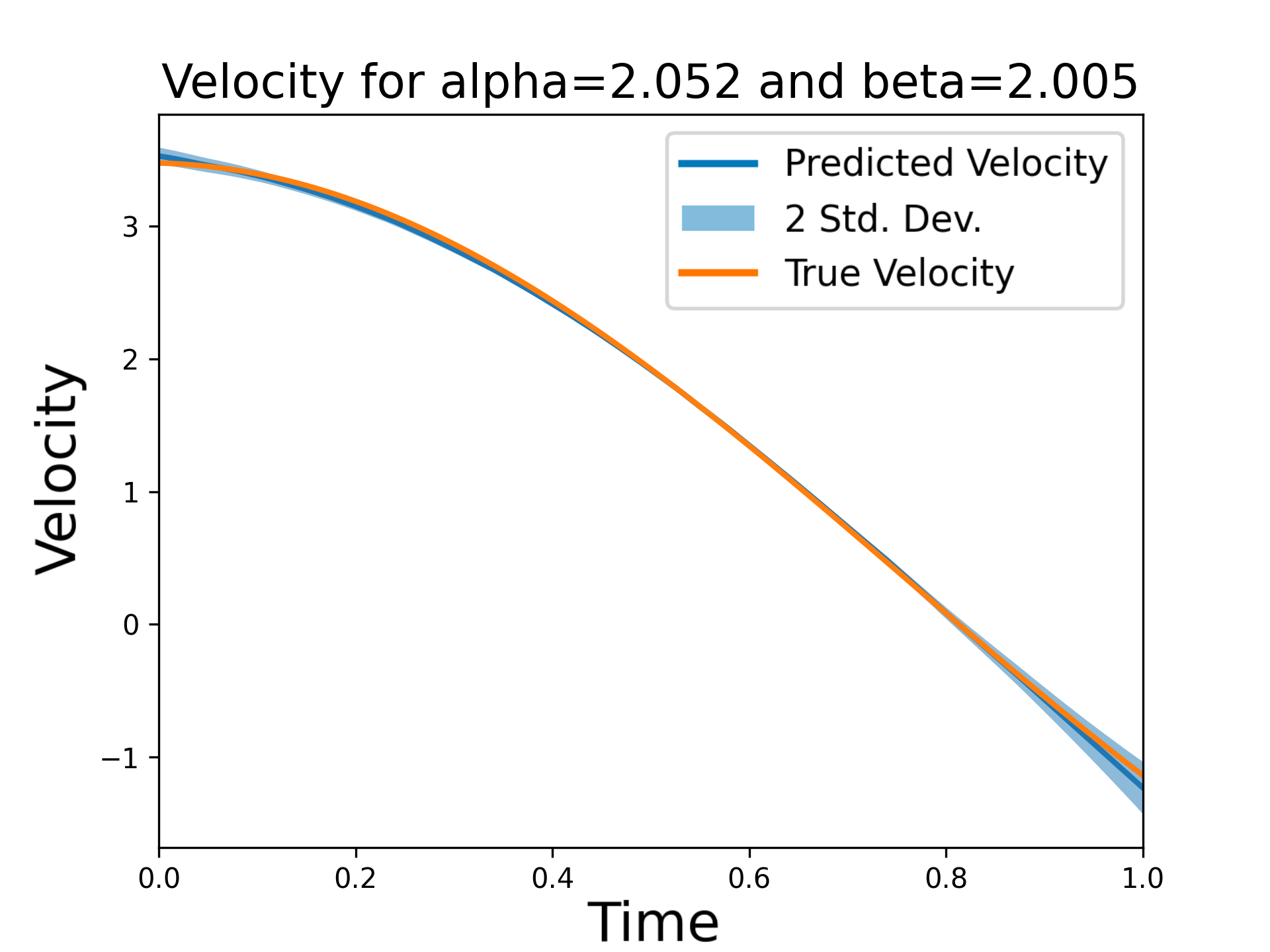} &
    \hspace{-0.15in}\includegraphics[width=0.255\textwidth]{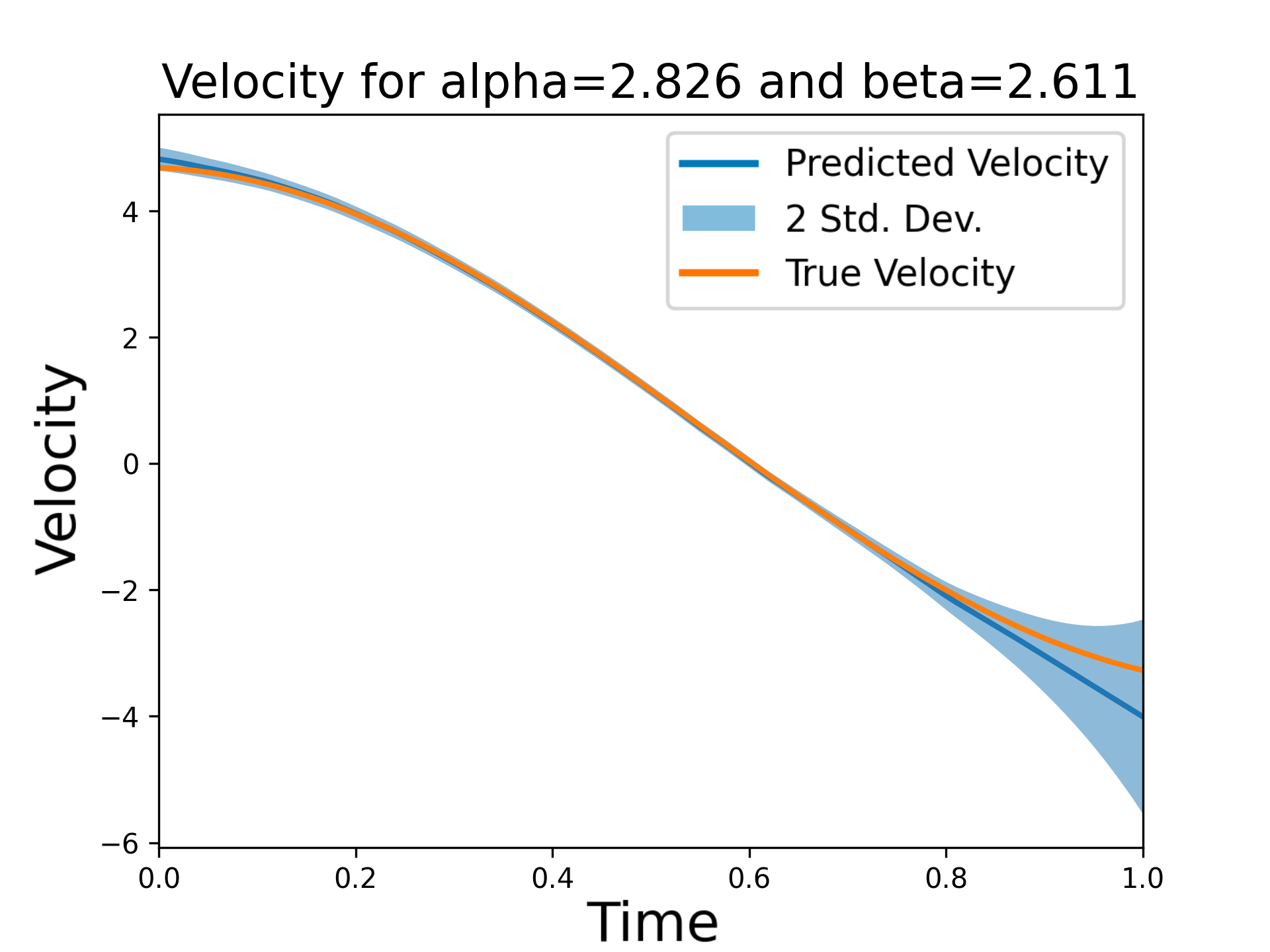} \\
    \hspace{-0.125in}\includegraphics[width=0.255\textwidth]{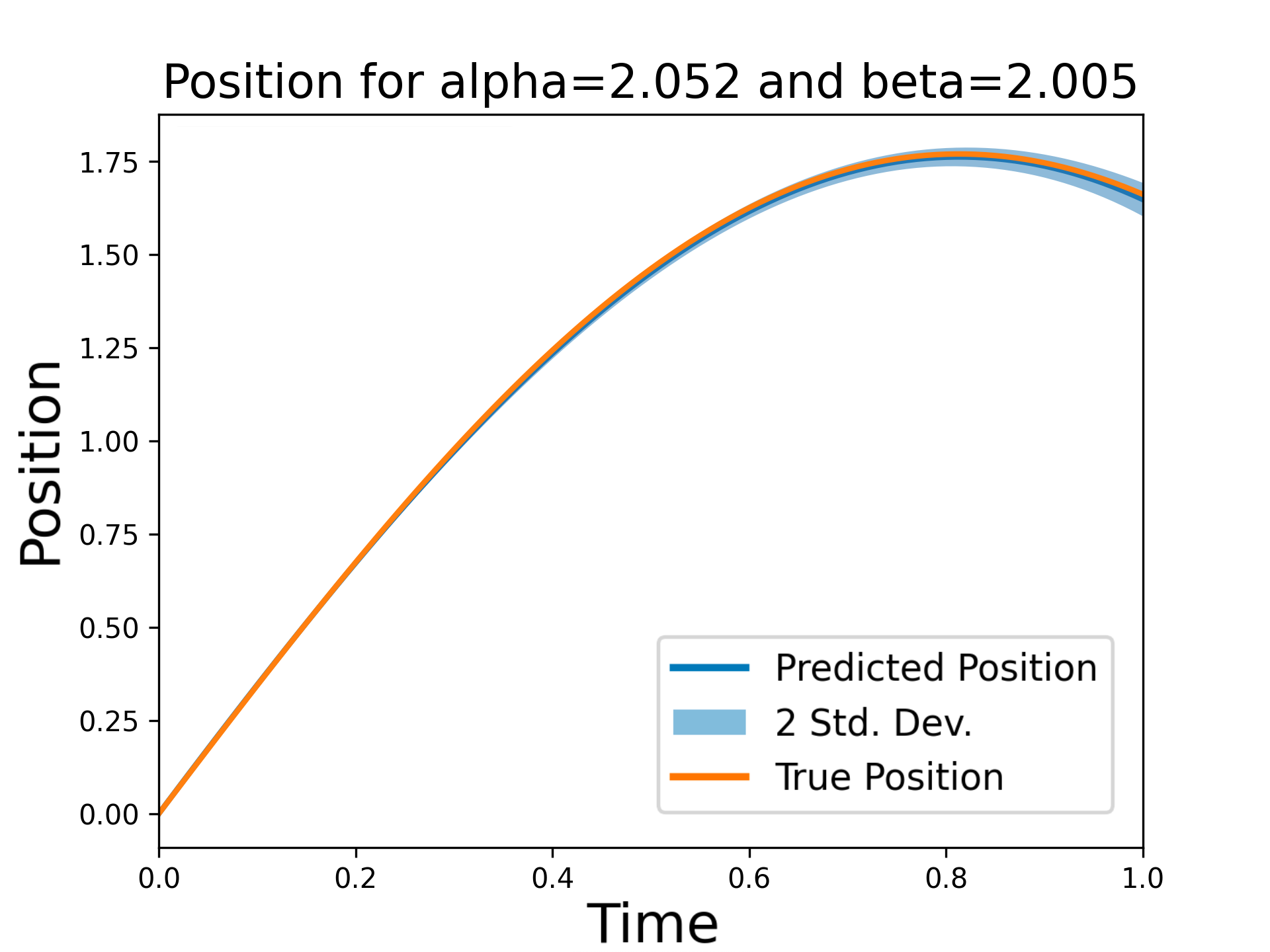} &
    \hspace{-0.2in}\includegraphics[width=0.255\textwidth]{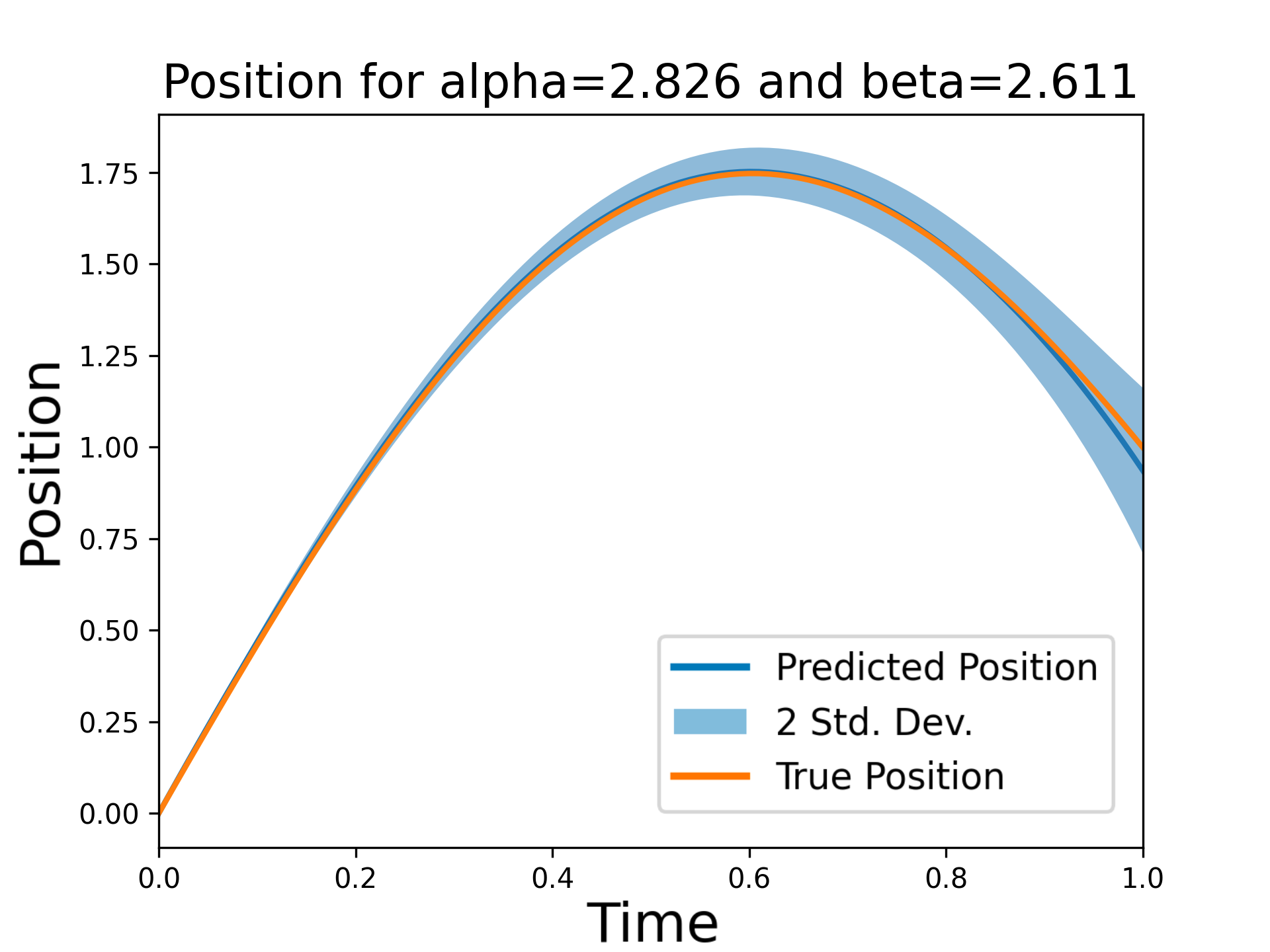} 
    \end{tabular} &
    \hspace{-0.10in}\raisebox{-1.025in}{\includegraphics[width=0.425\textwidth]{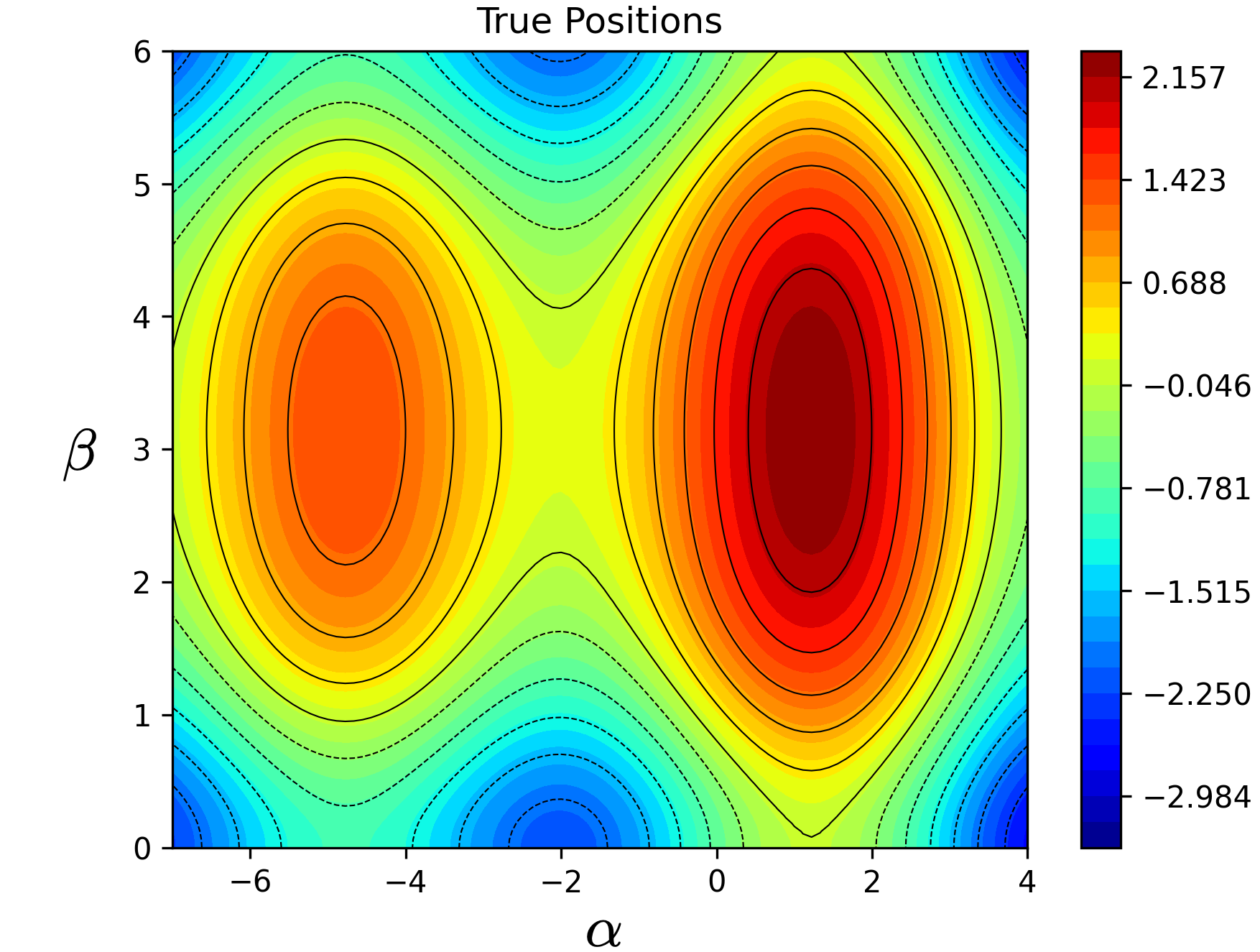}}
  \end{tabular}
  \caption{
    Example trajectories for velocity (upper-left) and position (lower-left) along with network predictions at the end of function-to-function learning. %
    The true position field at time $t=1$ is plotted on the right for parameter values $\alpha\in[-7,4]$ and $\beta\in[0,6]$. %
    The position field contains two peaks: %
    a local maximum on the left and a global maximum on the right. We sample the initial training locations around the local maximum to assess the framework's ability to avoid local trapping.}
    \label{fig:position_field}
\end{figure}

We will identify the optimal velocity function using as few system queries as possible; this is a common requirement for practical applications where data collection is expensive and/or time-consuming and QoI estimates must be made using a limited budget of system interrogations. %
For improved data efficiency, %
we employ {\revone{BO}} to iteratively refine the DeepONet model and %
direct our search for the optimal system configuration. %
The governing equations for this problem were specifically designed to include two peaks inside the constrained parameter space: a local maximum at $(-4.7680, \pi)$ and a global maximum at $(1.2150, \pi)$, as shown in Fig.~\ref{fig:position_field}. 
We start with a random selection of points with $\alpha < -2.0$, so that the initial observations are confined to the left half of the domain with a bias toward the local maximum.  %

The proposed \revone{BO} procedure leverages the uncertainty estimates of the DeepONet models to {\emph{explore}} the parameter space in regions of high uncertainty while also {\emph{exploiting}} %
the mean predictions to search regions with high predicted QoI values. %
To assess the performance of the {\revone{BO}} procedure, we compare its results with greedy (exploitation-driven) and random (exploration-driven) alternatives. %
Details for the three strategies are summarized below:
\begin{itemize}[leftmargin=0.375in]
\item \textbf{\revone{Bayesian optimization} with UQ:}
  New query locations are proposed based on the upper confidence bound induced by the network's predictive uncertainty.  %
  We evaluate the network on a grid of parameter values, $\Lambda \subset [-7,4]\times[0,6]$, and integrate the network's velocity predictions to obtain mean and standard deviation estimates for the QoI.  We use the acquisition values $\widehat{q}_\theta + \varepsilon \cdot \sigma(\widehat{q}_\theta)$ to rank the query locations $\theta\in\Lambda$, where $\varepsilon>0$ isused to weight the impact of uncertainty. %
  The system is then evaluated at the locations with the highest acquisition values, and the network is re-trained with the augmented dataset before the next iteration.  %
\item \textbf{Greedy:}
  We train a standard DeepONet model, without uncertainty, to approximate the velocity functions and integrate the predictions to obtain QoI estimates. %
  We select the parameters $\theta\in\Lambda$ with the highest predicted QoI values, $\widehat{q}_\theta$, and perform additional system queries at these locations.
  The network is then re-trained before starting the next iteration. %
\item \textbf{Random:} %
  We train an uncertainty-equipped DeepONet model using query locations sampled uniformly from the parameter grid $\Lambda$ at each iteration. %
\end{itemize}

For both the \revone{BO} and greedy strategies, a degree of random exploration is introduced to balance exploitation and exploration. %
During the initial iterations, we supplement the model's proposed query locations with additional points sampled uniformly from the parameter grid $\Lambda$. 
We found this significantly improved the consistency of both approaches and helped prevent instances of local trapping\footnote{Both the {\revone{BO}} and greedy strategies occasionally converged to the local maximum during our initial experiments.  This problem was resolved by incorporating random exploration to supplement both search strategies. }.  %
{\revone{We provide detailed summaries of the network architectures and training procedures in \ref{appendix:bo_training}.}}

\subsection{Parameter-to-Function Learning}

In the first formulation, 
we feed the parameters $(\alpha, \beta)$ directly into the branch network of the DeepONet and train the trunk network using randomly sampled time locations on the interval $[0,1]$. %
An initial training dataset is created by randomly sampling $50$ parameter locations in %
$[-7,-2] \times [0,6]$ and computing the associated velocity functions.  
The DeepONet model is tasked with approximating the mapping $\theta \mapsto u_\theta(t)$, and we use the predictive uncertainty estimates to {{compute the value of the upper confidence bound acquisition function~\cite{cox1992statistical}.}}  This acquisition function is used to select the next batch of queries for each learning iteration.  %
To obtain batches of distinct query locations, we define a selection distance, $\delta_\tau > 0$, that specifies the minimum allowable distance between queries proposed at the current iteration, $\tau$. 
We achieve this by selecting query locations sequentially according to the following rule: 
\begin{equation}
  \theta^*_{\tau, k}  \,\, = \,\, \operatorname{argmax}_{\theta\in \Lambda_{\tau, k}} \, \widehat{q}_{\theta} \, + \, \varepsilon_{\tau} \cdot \sigma(\widehat{q}_\theta),  \hspace{0.2in} \mbox{where}   \hspace{0.2in}  \Lambda_{\tau,k} \, = \, \big\{ \theta\in\Lambda \, : \, |\theta - \theta^*_{\tau,j}| > \delta_\tau \,\, \forall \, j < k \big\}. 
  \label{eq:active_eq}
\end{equation}
We gradually decay the values of the exploration weight, $\varepsilon_\tau$, and selection distance, $\delta_\tau$, by setting $\varepsilon_\tau = 2\cdot(0.9)^\tau$ and $\delta_\tau = 1\cdot (0.9)^\tau$. %
This allows the model to propose more precise (and greedier) query locations in later iterations to refine its final prediction for the optimal parameters. %

At each learning iteration, we evaluate the system at $10$ query locations selected using Eq.~\eqref{eq:active_eq} for the {\revone{BO}} setup,  the highest predicted QoI for the greedy strategy, and uniformly sampled locations for the random approach. %
To assist with exploration, we also include $5$ randomly sampled query locations during the initial iteration.  We reduce the number of random points added by a factor of $0.9$ for subsequent iterations, rounding decimal values to the nearest integer.  %
We perform $20$ learning iterations for each strategy, resulting in a total of $294$ system queries.  We conducted $50$ independent trials for each strategy to assess performance across different initial sample locations.  The initial data locations were sampled randomly for each of the $50$ trials, and each learning strategy used the same initial samples to provide a fair comparison of the results.

\begin{figure}[bht]
    \centering
    \hspace{-0.1in}\includegraphics[width=0.85\textwidth]{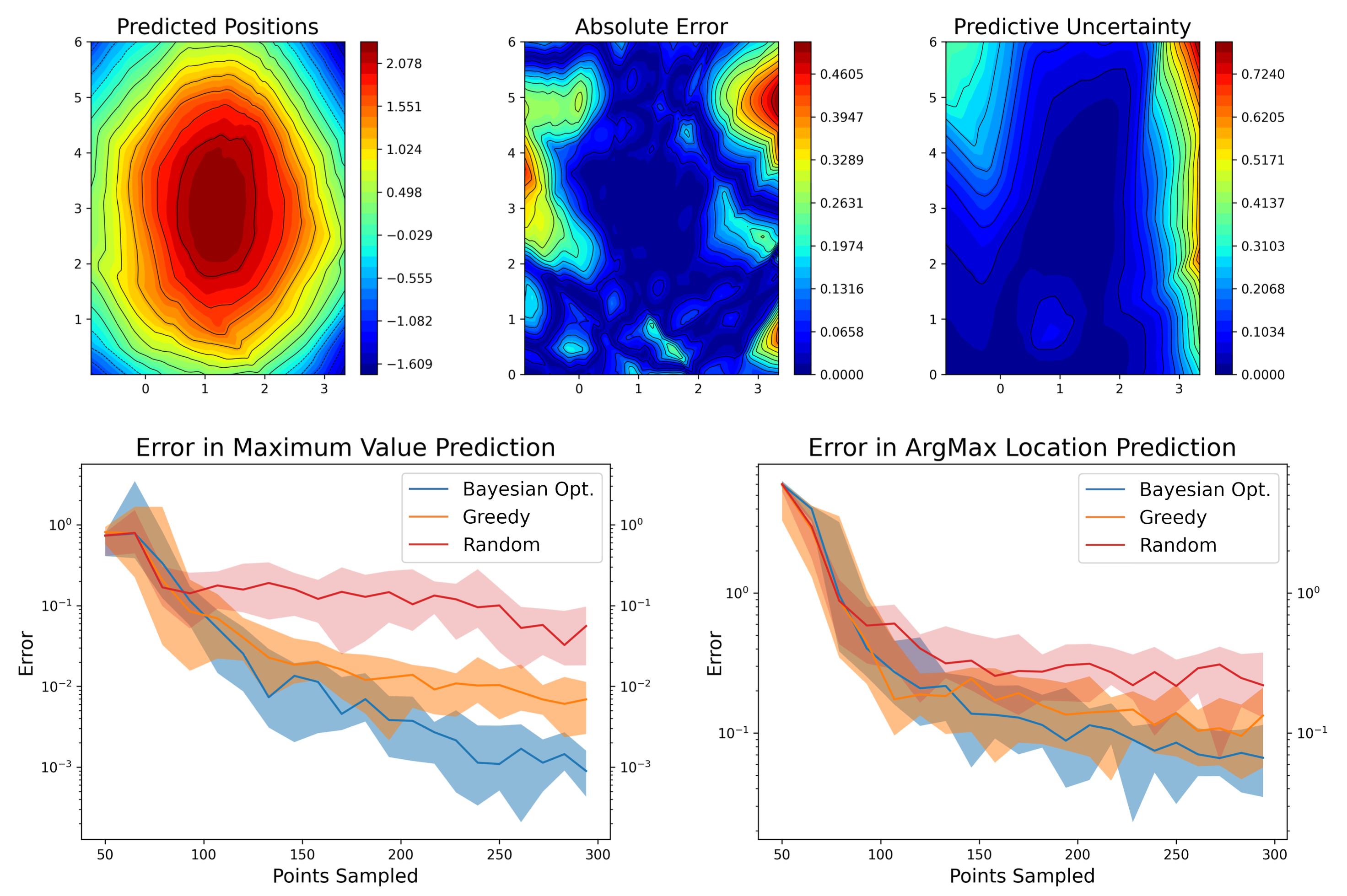}
    \caption{Results for parameter-to-function {\revone{BO}} problem.  %
      Plots of the final predicted position field at time $t=1$ (top left), the absolute error relative to the ground truth (top center), and the model's predicted uncertainty in position (top right). %
      By construction of the {\revone{BO}} setup, the model is most accurate near the peak location, $(1.2150,\pi)$.
      The uncertainty in position is derived from the model's UQ predictions for velocity %
      and give a reasonable approximation of the observed errors. %
      We conducted $50$ trials for each learning strategy, with each trial consisting of $20$ learning iterations.  The median and spread between the $0.2$-quantile and $0.8$-quantile are plotted (bottom) for each iteration to summarize the results across trials.  The errors in the predicted maximum function value are shown on the left, and errors in the predicted maximal location are shown to the right. %
      }
        \label{fig:active_error_plots_new}
\end{figure}

Examples of ground-truth and predicted velocity functions are shown in Fig.~\ref{fig:position_field}, along with the network's predicted uncertainty bounds.  These mean and uncertainty functions are integrated over time to obtain estimates for the final position (and uncertainty in position). 
A qualitative summary of the 
position predictions {{produced}} by the {\revone{BO}} procedure are shown in Fig.~\ref{fig:active_error_plots_new}. %
We note that the error is lowest in a neighborhood around the global peak, indicating the learning procedure correctly targeted this region for refinement. The uncertainty estimates also align relatively well with the observed errors; however, the network uncertainty underestimates errors in the region $\alpha<0$, possibly due to the limited amount of training data available.  %

We assess the performance of each learning strategy using two metrics: the error in the predicted maximum QoI value and the error in the maximizing parameters for the QoI.  The ground truth solution for the maximum value is %
$q_{\theta^*} \approx 1.3410$, which occurs at $\theta^* = (1.2150, \pi)$. %
A quantitative comparison between the {\revone{BO}}, greedy, and random strategies is provided in Fig.~\ref{fig:active_error_plots_new}, which summarizes the results from $50$ trials of each learning strategy. %
The uncertainty-equipped DeepONet model achieves the lowest errors in both the maximum value and maximizing parameter estimates. %
For the maximum value metric, the {\revone{BO}} setup achieves a median final error of $8.96\mathrm{e-}4$ over $50$ trials, while the greedy and random strategies had median errors of $6.90\mathrm{e-}3$ and $5.59\mathrm{e-}2$, respectively.    %
The performance difference for parameter estimation was less pronounced, but the {\revone{BO}} setup still achieved the lowest median error, $0.067$, compared with median errors of $0.133$ and $0.219$ for the greedy and random approaches.

\subsection{Function-to-Function Learning}

For the second experiment, we modify the branch input of the DeepONet to use time-series observations of acceleration instead of the parameters $\alpha$ and $\beta$. %
We create an initial training dataset by randomly sampling $10$ parameter locations in the interval $[-7,-2] \times [0,6]$ and computing the associated acceleration and velocity functions.  %
The DeepONet model is then tasked with predicting velocity functions based on time-series information regarding acceleration.

Initial experiments showed the direct function-to-function mapping was too simple to benefit from {\revone{BO}}.  The mapping from acceleration to velocity amounts to a straight-forward integral estimate, and the models provided accurate approximations with minimal system interrogations.  
To increase the problem complexity, we modify the acceleration information passed to the network using the following transformation: 
\begin{equation*}
  f_\theta^{input}(t) \,\, = \,\, \left( \, 1 \, + \, \exp(-\tfrac{1}{2} f_\theta(t))  \, \right)^{-1}  \, - \, \tfrac{1}{2}.
\end{equation*}
We compute the modified acceleration values at $33$ temporal locations on the interval $[0,1]$, and use these values as the function inputs to the network.  
Since the velocity function has an initial value that is independent of acceleration, we assume the initial velocity, $u_\theta(0) = 1+\alpha - \cos(\beta)$, is known for each parameter $\theta$.  We then sum the known initial velocity with the network's output before computing the network's loss relative to the true velocity function. %

\begin{figure}[tb]
  \centering
  \hspace{-0.2in}\includegraphics[width=0.85\textwidth]{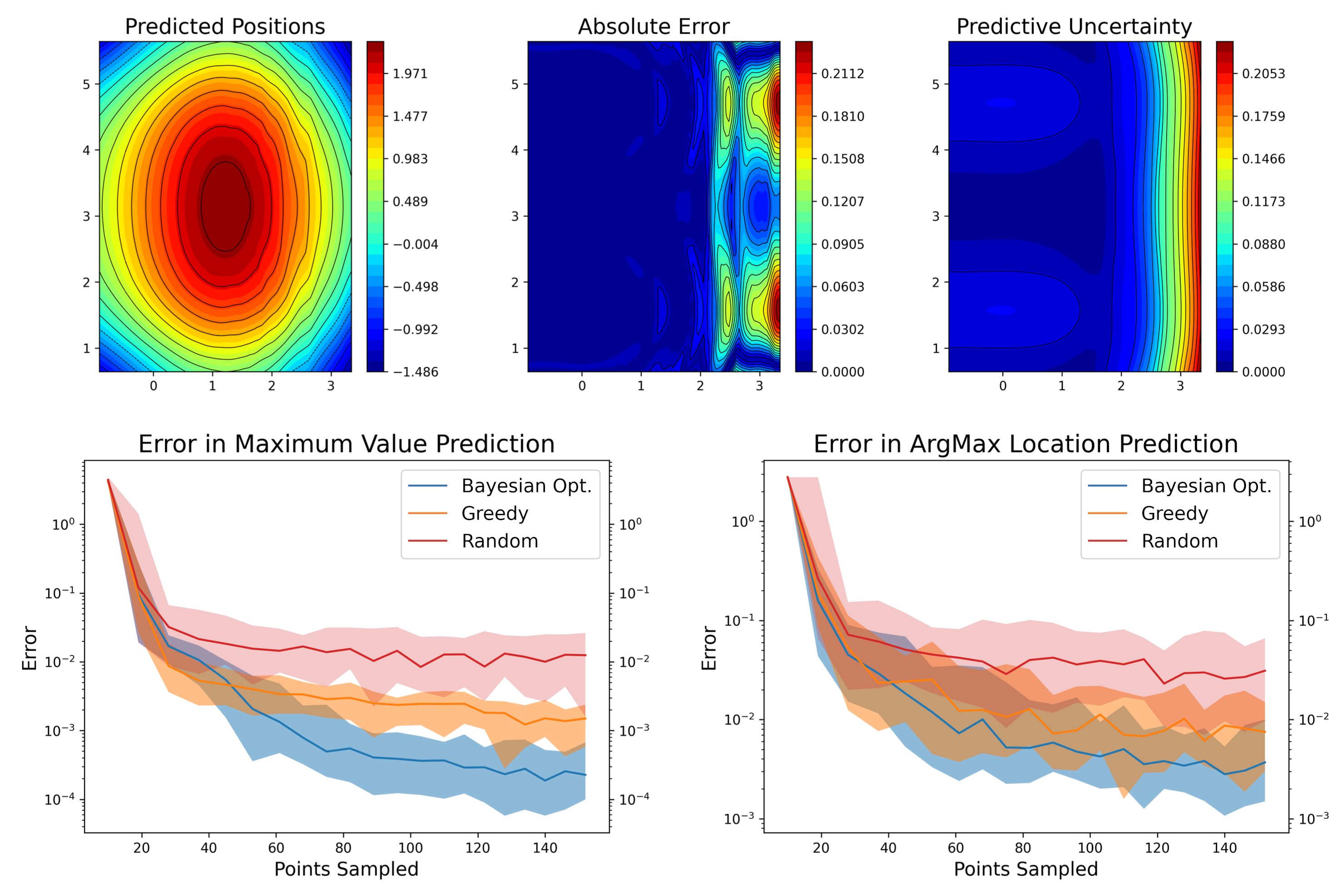}  
  \caption{Results for function-to-function {\revone{BO}} problem. %
    The final predicted position field at time $t=1$ for the function-to-function setup (top left).  %
    The predicted positions accurately reflect the ground truth values everywhere except the far-right side of the domain, as seen in the plot of observed errors (top center), which is consistent with the predicted uncertainty (top right). %
    We conducted $50$ trials for each learning strategy, with each trial consisting of $20$ learning iterations.  The median and spread between the $0.2$-quantile and $0.8$-quantile are plotted for each iteration to summarize the results across trials. %
    The error in the predicted optimal value (bottom left) and predicted optimal location (bottom right) are plotted for each learning strategy.  The {\revone{BO}} strategy enabled by predictive uncertainty estimates yields an order of magnitude improvement in optimal value prediction. %
  } %
    \label{fig:active_error_plots_function_new}
\end{figure}

At each learning iteration, we evaluate the system at $5$ query locations proposed by each learning strategy.  We also include $4$ randomly sampled query locations during the initial iteration, and we reduce the number of random points added by a factor of $0.9$ for subsequent iterations. %
We perform $20$ learning iterations for each strategy, resulting in a total of $152$ system queries. We conducted $50$ independent trials for each strategy using the same initialization procedure as before. %

A qualitative summary of the 
position predictions for the function-to-function setup are shown in Fig.~\ref{fig:active_error_plots_function_new}. %
We note that the predicted positions are more accurate, and better structured, than those obtained in the parameter-to-function setup (which had access to nearly twice as many system interrogations).  The predicted uncertainty also aligns well with the observed errors, although it still underestimates errors in some regions (likely due to the limited data available).

We assess the performance of each learning strategy using the same metrics as the parameter-to-function setup, and we again find that {\revone{BO}} with the uncertainty-equipped DeepONet model outperforms the greedy and random alternatives. %
A quantitative comparison of each strategy's performance over $50$ trials is provided in Fig.~\ref{fig:active_error_plots_function_new}. %
For the maximum value metric, the {\revone{BO}} setup achieves a median final error of $2.27\mathrm{e-}4$ over $50$ trials, while the greedy and random strategies had median errors of $1.50\mathrm{e-}3$ and $1.25\mathrm{e-}2$, respectively.    %
For the maximizing parameter estimates, the {\revone{BO}} setup achieved the lowest median error, $3.69\mathrm{e-}3$, compared with median errors of $7.51\mathrm{e-}3$ and $3.11\mathrm{e-}2$ for the greedy and random approaches.  %



\section{\revboth{Accelerating active learning with predictive uncertainties}}
\label{sec:AL_PDEs}

\subsection{\revboth{Active learning for advection-diffusion PDEs}}
\label{sec:advection}

\begin{figure}
  \centering
  \begin{tabular}{cccc}
    \includegraphics[width=0.22\textwidth]{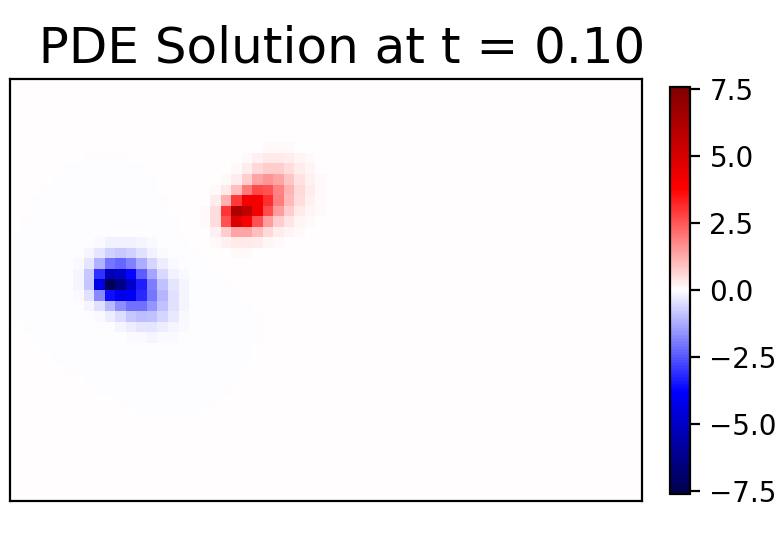} & 
    \includegraphics[width=0.22\textwidth]{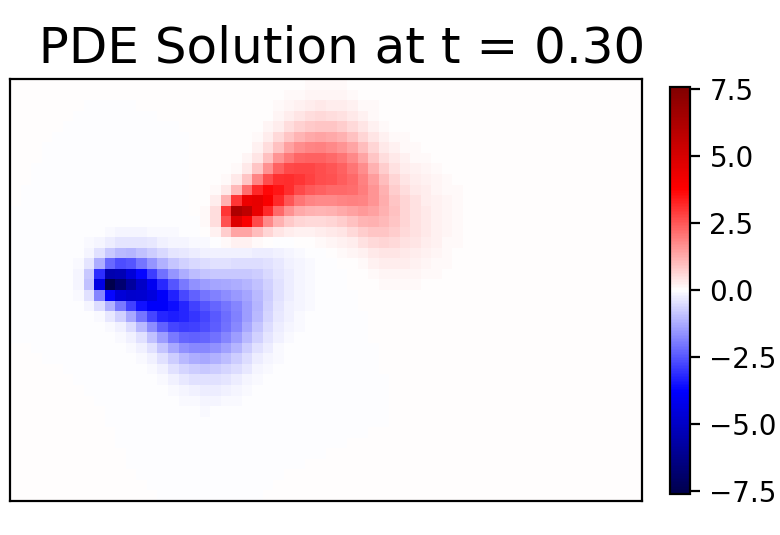} & 
    \includegraphics[width=0.22\textwidth]{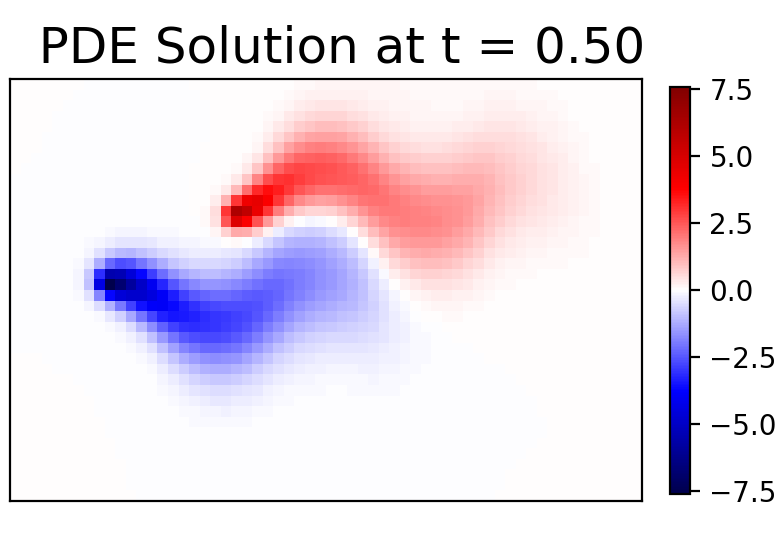} & 
    \includegraphics[width=0.22\textwidth]{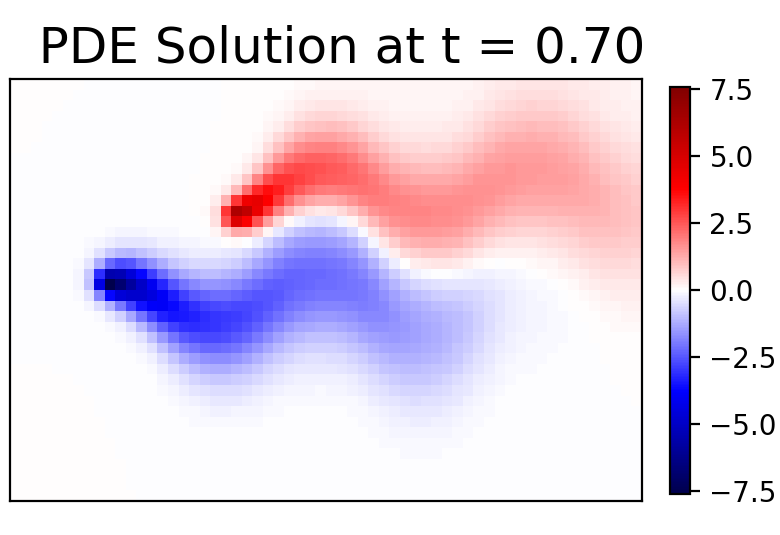} 
  \end{tabular}
  \caption{\revboth{Example solution to the advection-diffusion equation with randomized sink (blue) and source (red) locations in an advection dominant system.}}
  \label{fig:advection_domain}
\end{figure}

{\revboth{To evaluate how predictive uncertainty accelerates active learning, we benchmark our framework on a realistic PDE setting. While Section~\ref{sec:bo} introduced the method in a visually intuitive context, operator learning is most impactful for high-dimensional PDEs. We consider a 2D advection-diffusion equation with time dependence in an active learning environment:}}
\begin{equation}
  \tfrac{\partial}{\partial t} u(x,t) \, - \, \operatorname{div}(\kappa \nabla u(x,t)) \, + \, \nu(x) \cdot \nabla u(x,t)  \,\,=\,\, f_{source}(x) - f_{sink}(x) \hspace{0.15in} \forall \,\, x\in\Omega \,\, , \,\, t\in (0,0.7) 
  \label{equation:advection}
\end{equation}
{\revboth{with diffusion coefficient $\kappa = 0.01$ and velocity field given by:}}
\begin{equation}
  \nu(x_1,x_2) \,\, = \,\, 
  \begin{bmatrix}
    (1+x_1) \cdot \sqrt{0.9^2 - (0.75\cdot\cos(1.1-x_1) \cdot \sin( 4\pi  x_1) )^2} \\
    \, -0.9\cdot \cos(1.1-x_1) \cdot \sin( 4\pi x_1)
  \end{bmatrix}
\end{equation}
{\revboth{We solve the PDE on the 2D rectangular domain $\Omega = (0.0,1.5)\times(0.0,1.0)$, with the diffusion coefficient and velocity field chosen to produce an advection-dominant system, as shown in Figure~\ref{fig:advection_domain}. %
The source term and sink terms are randomized for each problem realization and given by:}}
\begin{equation}
  f_{i}(x_1, x_2) \,\, = \,\,
  8.5/\sigma \cdot \operatorname{exp}\left( \sqrt{(x_1-c_{i,1})^2 + (x_2-c_{i,2})^2} / \sigma^2 \right)
  \hspace{0.135in} \mbox{for} \hspace{0.135in}  i \, \in \, \{source,\, sink\} 
\end{equation}
{\revboth{where the center locations $c_{i,1}$ and $c_{i,2}$ are sample uniformly from the interval $(0.1, 0.9)$ and the shared width parameter $\sigma$ is sampled from $(0.075, 0.125)$.  We constructed a dataset consisting of solutions to $5,000$ random realizations of this advection diffusion setup.  Each solution was evaluated at $36$ equally spaced time locations between $t=0.0$ and $t=0.7$, and snapshots for each time-step were saved to arrays of shape $60\times 40$ corresponding to the function values on a uniform grid covering the rectangular domain.}}

{\revboth{
The operator networks are passed a single $60\times 40$ array as input, reflecting the spatial values of the sink and source terms on the right-hand-side of the PDE, and are tasked with predicting the time evolution of the associated PDE solution.
The goal of the active learning problem is to minimize the overall error on the full dataset of $5,000$ PDE examples using as few examples as possible during training.}}

{\revboth{For this problem setup, we assess the predictive uncertainty framework on another class of operator learnings models, FNOs, and show that FNO models can also be extended to provide predictive uncertainty estimates.  We then compare the performance of uncertainty-equipped FNO models with two popular alternative uncertainty frameworks: ensemble models and Monte Carlo dropout~\cite{gal2016dropout,seoh2020qualitative}.}}  %
{\revboth{The architectures for each uncertainty framework were built on top of a common FNO backbone, consisting of spectral convolutional layers followed by a fully convolutional decoder.  The precise implementation details for each network architecture are provided in Section~\ref{appendix:advection_architectures} of \ref{appendix:al_architectures}.}}

\revboth{We perform $10$ iterations of AL, starting with an initial set of $250$ randomly sampled PDE realizations. %
  At the start of each AL iteration, the models are initialized and fit to the current dataset.  The networks are then evaluated on all $5,000$ examples from the complete dataset, and $50$ new PDE examples are selected for inclusion in the subsequent training iteration.  For the uncertainty-based models, the PDE examples with the highest uncertainty are used to augment the dataset at each iteration.  For the standard FNO model, we select examples randomly at each iteration and use these results as a baseline for comparison.} %

\begin{figure}[t]
  \centering
  \begin{tabular}{cc}
    \hspace{-0.2in} \includegraphics[width=0.48\textwidth]{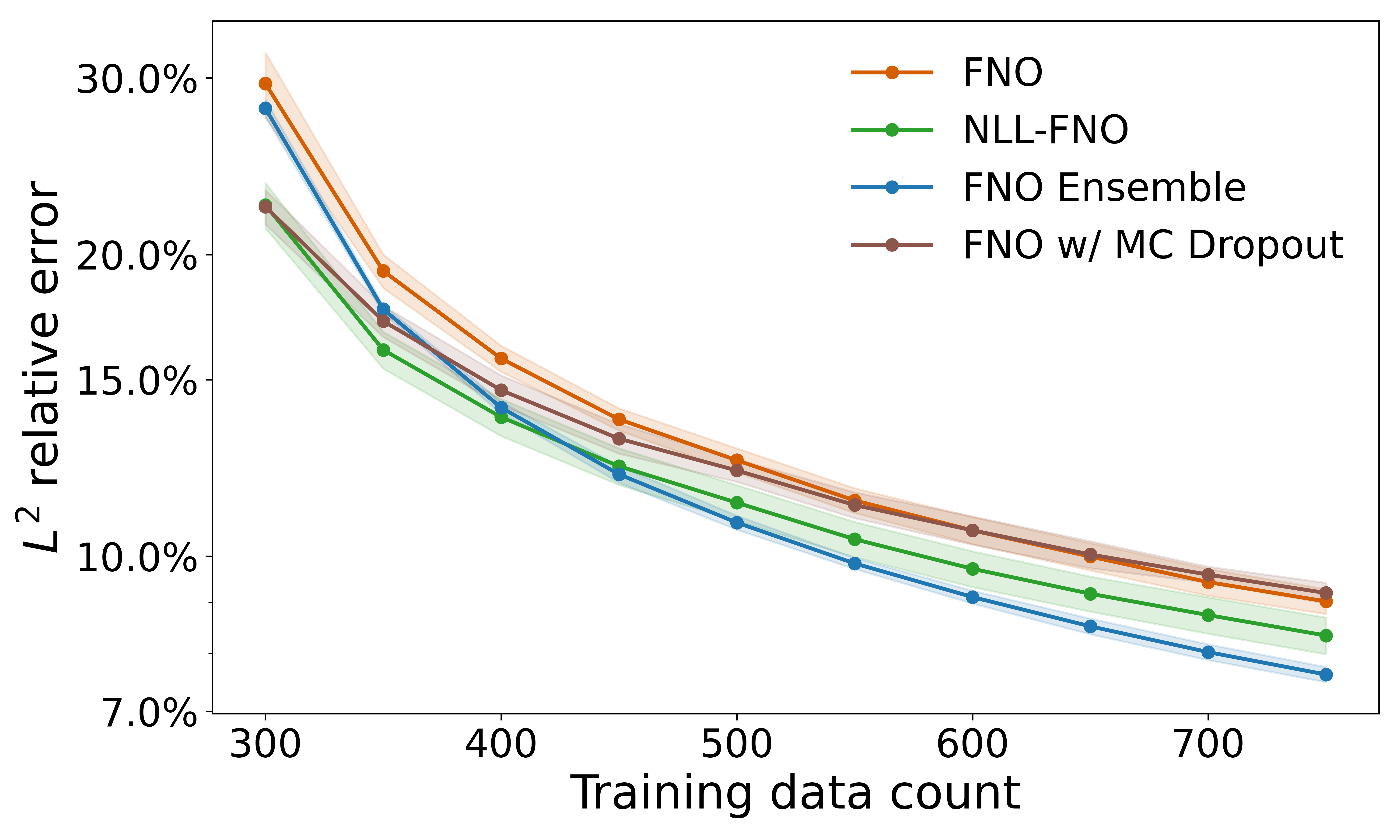} & 
    \includegraphics[width=0.48\textwidth]{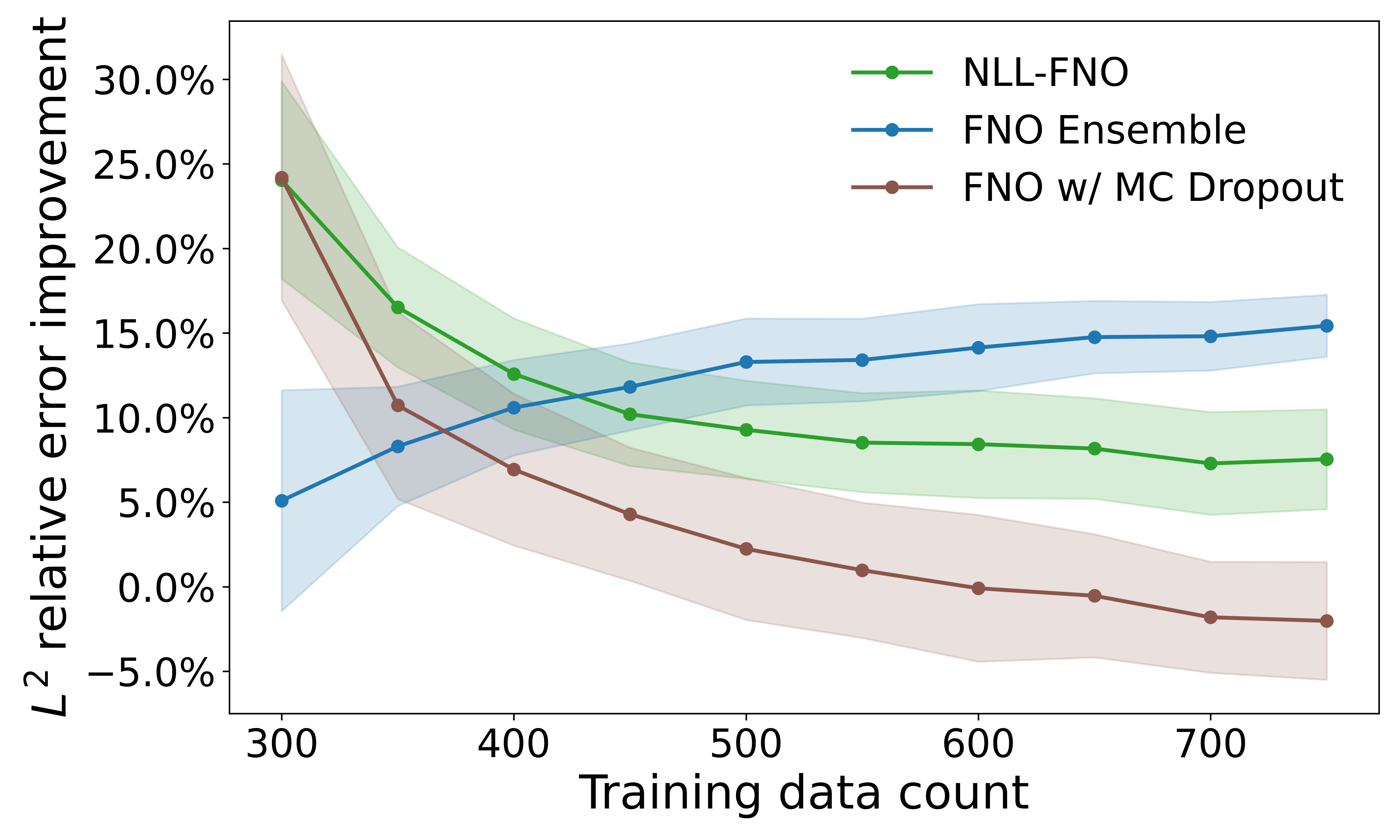} 
  \end{tabular}
  \caption{\revboth{$L_2$ relative error as a function of active learning (AL) training dataset size (left), averaged over 10 trials (outliers removed); one standard deviation is indicated by a lighter shaded band. Improvement metric using the same 10 trials (right), comparing FNO to FNO with NLL and DeepONet to DeepONet with NLL. The improvement is computed as the relative error improvement, $(L_2^{\operatorname{FNO}} - L_2^{\operatorname{FNO}_{\operatorname{NLL}}}) \,/\, L_2^{\operatorname{FNO}}$ (and analogously for others).}}
  \label{fig:advection_results}
\end{figure}

\revboth{To assess how the methods perform across different random initializations, we performed $10$ independent trials of the active learning procedure, with each initialized using a distinct set of $250$ PDE examples.  A complete table of the numerical results for each trial and active learning iteration is provided in Table~\ref{table:advection_combined_results} in \ref{appendix:al_table_results}, and the results are summarized in Figure~\ref{fig:advection_results}. }

%
%




\begin{table}[t]
\centering
{\fontsize{10}{10}\selectfont{
\begin{tabular}{ccccccccccc}
Iteration & 1 & 2 & 3 & 4 & 5 & 6 & 7 & 8 & 9 & 10 \\ 
 \hline \Tstrut 
FNO & 0.7  & 1.4  & 2.1  & 3.0  & 3.9  & 4.9  & 5.9  & 7.0  & 8.2  & 9.4  \\ 
NLL-FNO\, & 0.7  & 1.5  & 2.4  & 3.3  & 4.3  & 5.4  & 6.5  & 7.8  & 9.1  & 10.4  \\ 
MC Dropout & 1.4  & 3.0  & 4.7  & 6.5  & 8.6  & 10.8  & 13.2  & 15.6  & 18.3  & 21.1 \\
Ensemble FNO & 8.6  & 18.5  & 29.1  & 40.9  & 53.9  & 67.5  & 82.5  & 98.1  & 114.9  & 132.4  
\end{tabular}
}}
\caption{\revboth{Summary of cumulative training time (minutes) for advection-diffusion setup averaged across runs.}}
\label{table:results_average_runtimes_advection}
\end{table}

\begin{table}[t]
\centering
{\fontsize{10}{10}\selectfont{
\begin{tabular}{ccc}
  Model &  $L^2$ relative error & Runtime \\
  \hline\Tstrut
  FNO           &  3.34\%   &  19.3 min  \\
  NLL-FNO      &  3.25\%   &  21.8 min  \\
  MC Dropout    &  5.26\%   &  47.5 min  \\
  Ensemble FNO  &  3.30\%   &  138.7 min  
\end{tabular}
}}
\caption{\revboth{Summary of training results for advection-diffusion  using full dataset of $5,000$ PDE examples.}}
\label{table:results_full_dataset_advection}
\end{table}

\revboth{The FNO ensemble model consistently provided the best performance, with $L^2$ relative error improving on average by a factor of $15.4\%$ relative to the baseline FNO model by the final active learning iteration.  The NLL-FNO model produced more modest improvements by a factor of $7.5\%$, while the MC Dropout approach ultimately had a negative impact on accuracy, reducing accuracy by a factor of $2.0\%$.  To understand how uncertainty calibration may have impacted these final accuracies, we also computed the UQ coverage associated with each model's predictive uncertainty: that is, the percentage of data points that fall within 1, 2, and 3 standard deviations of the model's prediction.  The coverage percentages for the ensemble models were $86.1\%$, $97.1\%$, and $99.1\%$, which are generally in agreement with the target empirical values of $68.0\%$, $95.0\%$, and $99.7\%$.  The NLL-FNO model had similar coverage percentages of $86.5\%$, $94.0\%$, and $95.6\%$, while the MC dropout model percentages were $55.6\%$, $78.8\%$, and $90.3\%$.  Both the ensemble and NLL approaches had noticeable over-coverage for the single standard deviation metric, but were reasonably well calibrated for 2 and 3 standard deviations.  In contrast, the MC dropout model exhibits under-coverage for all three metrics, which likely contributed to the weak performance of this model for active learning. }

\revboth{While the ensemble approach yields the largest improvement in accuracy, and was observed to have the best uncertainty coverage, it introduces significantly more computational overhead in comparison with the alternatives. The average runtimes for each active learning iteration are reported in Table~\ref{table:results_average_runtimes_advection}.  The cumulative training time using the standard FNO model was $9.4$ minutes.  Introducing the uncertainty predictions for the NLL-FNO model increase the runtime by $10.6\%$ on average, and using the MC dropout approach increased the runtime by $124.5\%$\footnote{The increase in runtime for MC dropout is primarily due to the increased number of training epochs at each iteration, which we found necessary for the model to achieve comparable accuracy to the baseline model.}.  The overhead for training the ensemble model is noticeably higher, with an average increase in runtime by $1308.5\%$. }

\revboth{Depending on the application, introducing this level of overhead during training may or may not be acceptable.  For smaller problems, like the one considered here, the overall training time is still manageable and can be justified by the improved model accuracy.  For larger problems, the computational demands can grow considerably, as we shall see in the next section, and training ensembles may become prohibitively expensive.  Moreover, both the ensemble and MC dropout approaches introduce additional overhead during the inference stage.  The average time required to evaluate the baseline FNO model on the complete set of $5,000$ PDEs was $1.2$ seconds.  The evaluation time was virtually unchanged for NLL-FNO, which took $1.2$ seconds, while the ensemble model required $9.6$ seconds and MC dropout took $16.2$ seconds\footnote{The computational costs at inference time are directly proportional to the number of models used for the ensemble approach (set to $15$ for our experiments) and the number of samples used for MC dropout (set to $25$ in this work). }.  These increased costs at inference time can be more difficult to justify than the overhead introduced during training.  In many cases, the primary motivation for training operator networks is to provide lightweight surrogate models that can accelerate outer-loop tasks and analyses.  If the inclusion of UQ results in surrogates that are an order-of-magnitude slower during inference, the benefits of any added accuracy or uncertainty coverage may be undermined by the reduced number of queries that can be made to the surrogates.  This will ultimately depend on the target application, but it is a tradeoff that should be carefully considered when selecting an uncertainty framework. }

\revboth{It is also important to consider whether the active learning procedure has any true advantage over a standard training procedure. Generating solutions for a single PDE takes approximately $9.09$ seconds; generating the complete set of $5,000$ examples took $12.6$ hours, while only $1.9$ hours were required to generate the $750$ examples used for active learning.  The accuracies and runtimes for training models on the complete dataset are provided in Table~\ref{table:results_full_dataset_advection}.  While data generation dominates the overall compute time, the models trained using the full dataset achieve noticeably lower losses.  We will return our attention to this tradeoff with a more detailed discussion in the next section. 
}




\subsection{\revboth{Active learning for wave equation on an irregular domain}}
\label{sec:active}

\begin{figure}
  \centering
  \begin{tabular}{cccc}
    \includegraphics[width=0.22\textwidth]{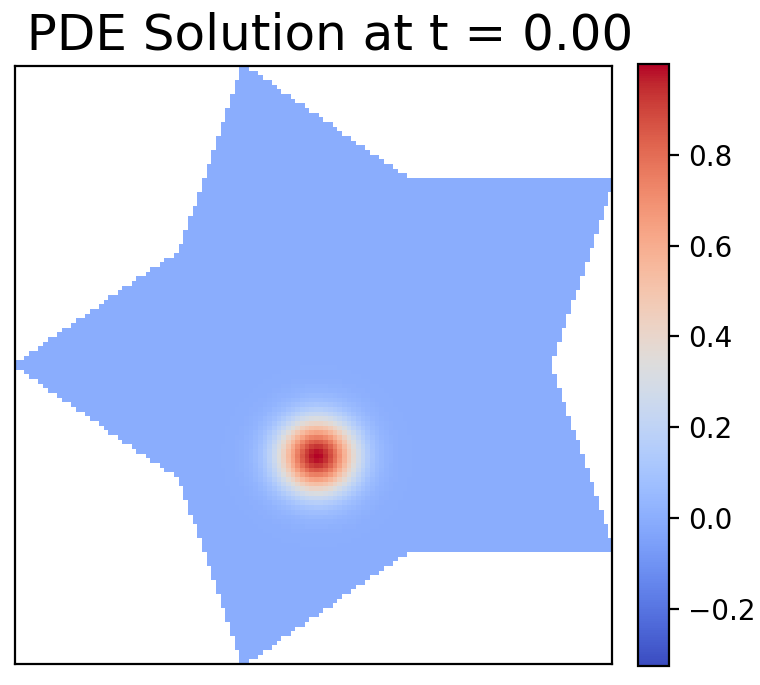} & 
    \includegraphics[width=0.22\textwidth]{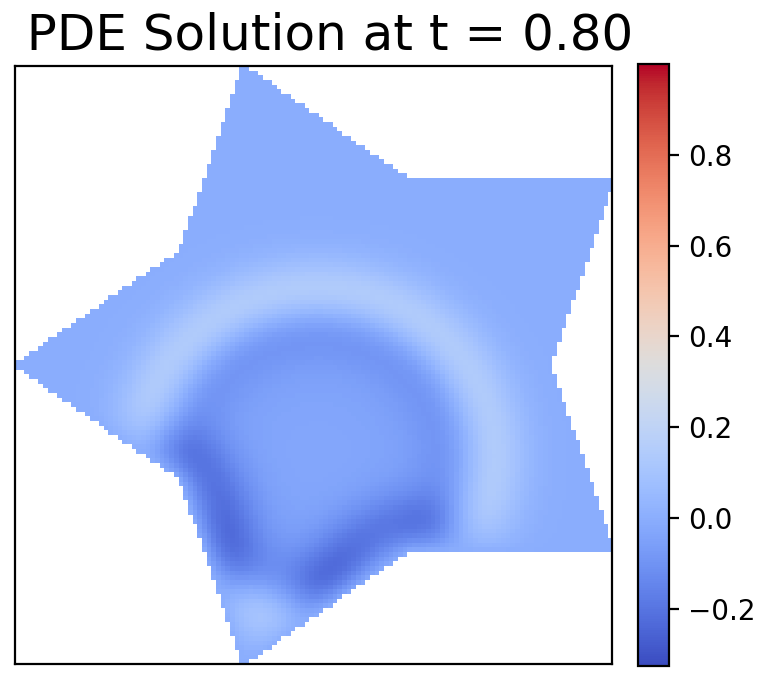} & 
    \includegraphics[width=0.22\textwidth]{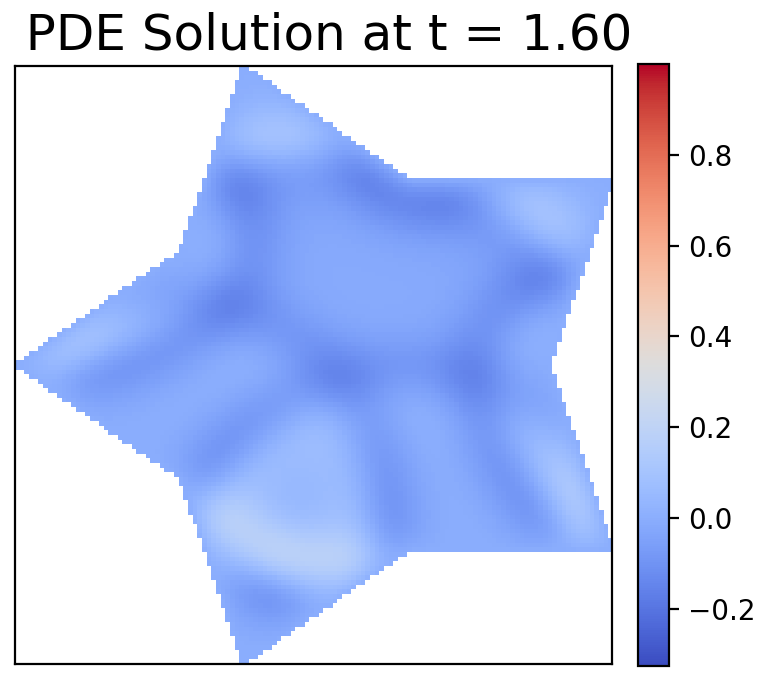} & 
    \includegraphics[width=0.22\textwidth]{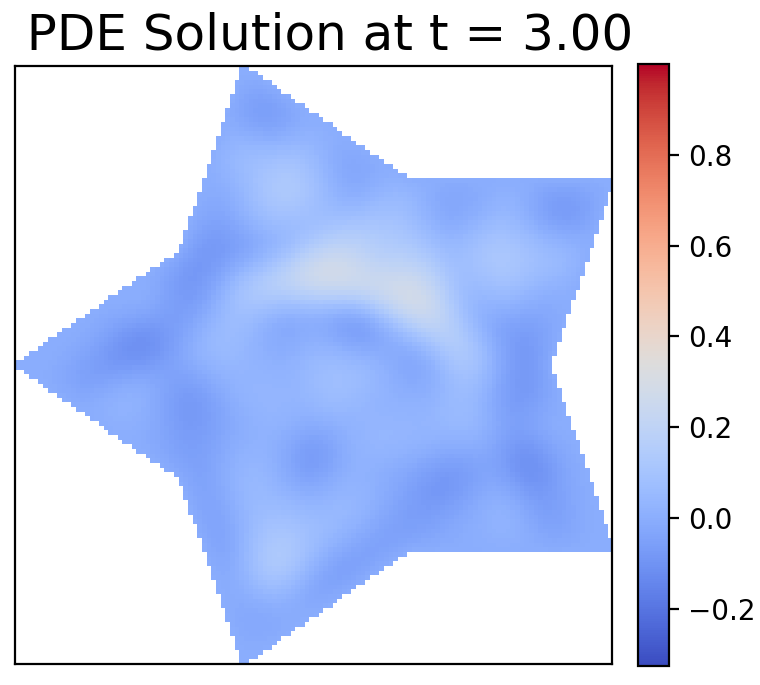} 
  \end{tabular}
  \caption{\revboth{Illustration of the star-shaped domain used for AL experiments, and an example PDE solution evolving in time.  The location and width of the initial concentration are varied to generate a family of problems, and operator networks are tasked with predicting the time evolution of the system.}}
  \label{fig:al_domain}
\end{figure}

\revboth{The results from \ref{sec:advection} highlight the potential for predictive uncertainties to improve active learning with operator networks for relatively simple time-dependent systems. For our final numerical experiments, we aim to assess how these results extend to more complex systems.  To that end, we consider an active learning problem involving a family of 2D time-dependent wave equations:}
\begin{equation}
  \tfrac{\partial^2}{\partial t^2} u(x,t)  \,\,=\,\, \nabla^2 u(x,t) \hspace{0.15in} \forall \,\, x\in\Omega \,\, , \,\, t\in (0,3) \hspace{0.2in} \mbox{with} \hspace{0.2in}   u(x,0) \,\, = \,\,  \operatorname{exp}(-L||x-c||^2) \,,
  \label{equation:active_learning}
\end{equation}
\revboth{where $c$ is a randomly sampled center, $L$ is a length scale determining the width of the initial peak, and natural boundary conditions, $\tfrac{\partial u}{\partial n} = 0$, are enforced. To assess operator networks' ability to handle non-standard geometries (e.g., rectangular or circular domains), we selected a two-dimensional star-shaped domain $\Omega$ with five equally spaced points placed on the unit circle.}  %
\revboth{This PDE was selected because it introduces time dependence, a non-standard geometry, and hyperbolic governing equations, making it significantly more challenging than the proof-of-concept example in Section~\ref{sec:bo}.} %
\revboth{We generated solutions to 20,000 realizations of the PDE using MFEM~\cite{anderson2021mfem} with uniform random samples for center locations $c\in[-0.5, 0.5]^{\times 2}$ and length scales $L\in[27.0, 33.0]$.  The FEM solutions were then projected to a discrete spatial grid with resolution $128\times 128$ at $76$ linearly spaced time locations spanning the interval $[0,3]$. Fig.~\ref{fig:al_domain} depicts the domain and an example reference solution.}

\revboth{We assessed the performance of DeepONets and FNOs using an AL framework for solving the PDE family described in Equation~\ref{equation:active_learning}.} %
\revboth{The operators were passed a discrete $128\times 128$ array containing the initial values $u(x,0)$ on a uniform grid as input and tasked with predicting the evolution of the system from $\,t\!=\!0\,$ to $\,t\!=\!3$.} %
\revboth{We perform $10$ iterations of AL, starting with an initial set of $250$ randomly sampled PDE realizations. %
At the start of each AL iteration, the models are initialized and fit to the current dataset.  The networks are then evaluated on all $20,000$ examples from the complete dataset, and $50$ new PDE examples are selected for inclusion in the subsequent training iteration.} %
\revboth{The objective of the AL procedure is to achieve the highest accuracy on the full dataset using as few examples as possible during training.} %
\revboth{For uncertainty-equipped models, the examples with the highest predicted uncertainty are used to augment the training dataset.  For models without uncertainty, the points are selected randomly.} %
\revboth{We provide the complete details for the AL training procedure in \ref{appendix:al_training}.} %

\begin{figure}
  \centering
  \begin{tabular}{cc}
    \hspace{-0.2in} \includegraphics[width=0.48\textwidth]{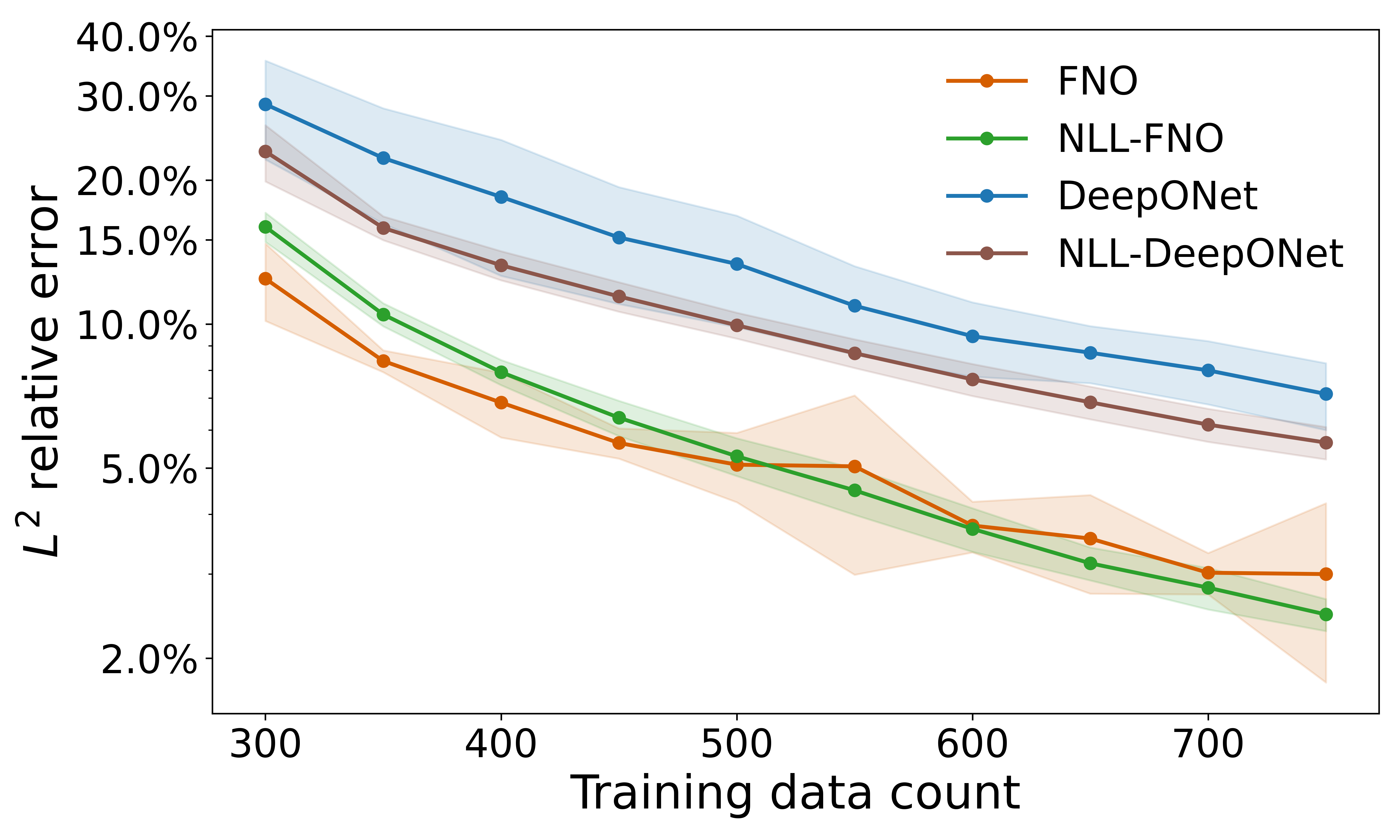} & 
    \includegraphics[width=0.48\textwidth]{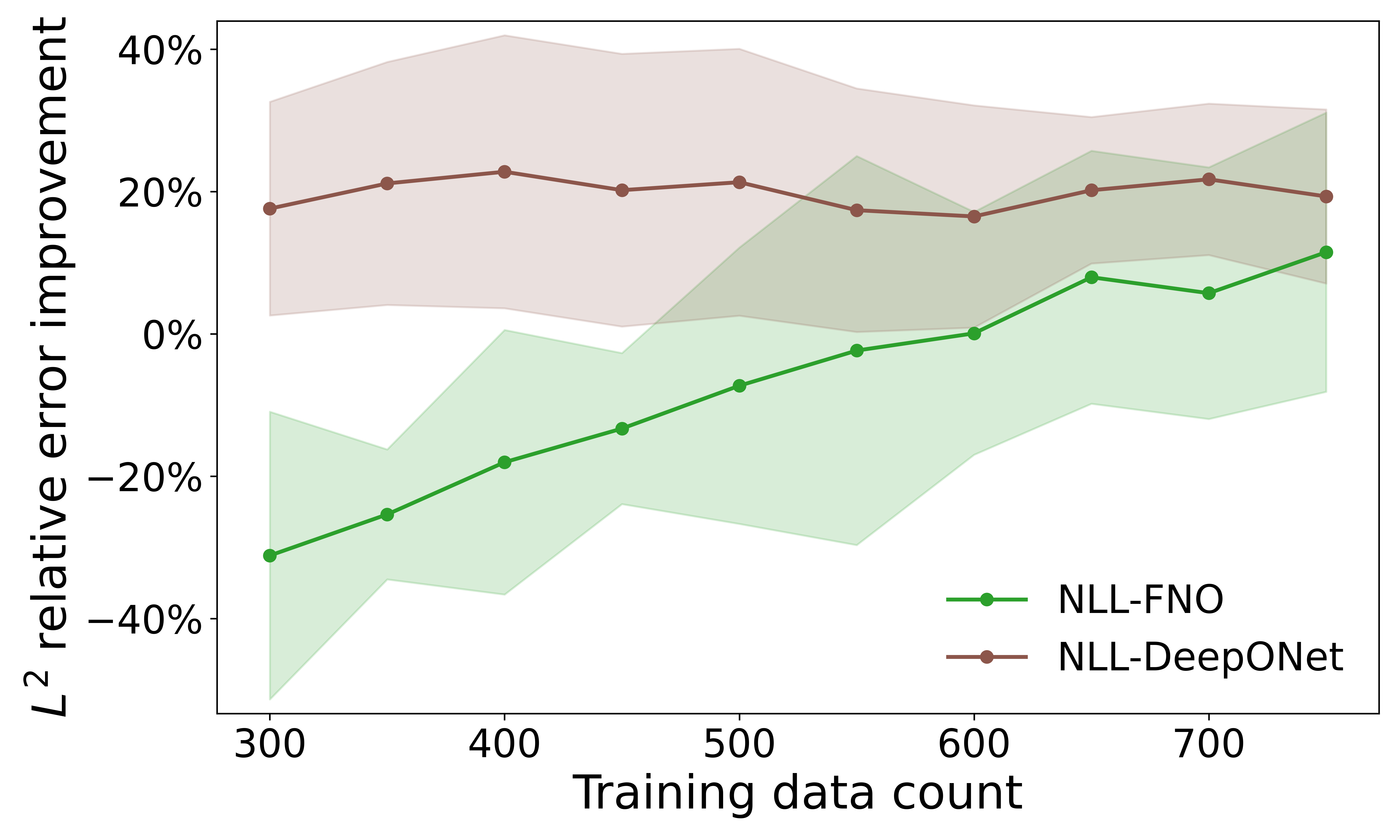} 
  \end{tabular}
  \caption{\revboth{$L_2$ relative error as a function of active learning (AL) training dataset size (left), averaged over 10 trials (outliers removed); one standard deviation is indicated by a lighter shaded band. Improvement metric using the same 10 trials (right), comparing FNO to FNO with NLL and DeepONet to DeepONet with NLL. The improvement is computed as the relative error improvement, $(L_2^{\operatorname{FNO}}-L_2^{\operatorname{FNO}_{\operatorname{NLL}}}) \, / \, L_2^{\operatorname{FNO}}$ (and analogously for DeepONet).}}
  \label{fig:wave_results}
\end{figure}

\revboth{To ensure robustness, we repeat this process with $10$ different initializations of the starting dataset.} 
\revboth{Figure~\ref{fig:wave_results} shows accuracy as a function of dataset size for each AL iteration.} 
\revboth{DeepONet with NLL exhibits a consistent improvement ratio of approximately 20\% over DeepONet in all AL iterations. On the other hand, FNO with NLL performs worse than FNO in the first few AL iterations; however, once it has 500 datapoints, it shows improvement over random selection. We likely see poorer FNO performance due to limited data and challenges in uncertainty calibration, which requires a UQ calibration period that, once completed, provides an improved model compared to the standard FNO baseline.  Overall, the UQ models exhibit lower standard deviations compared to random selection, indicating greater stability among the models.}
\revboth{The performances of models trained using the complete set of $20,000$ PDE examples are provided in Table~\ref{table:results_full_dataset} for comparison.}

%
%

%
%

%
%

%
%

\subsubsection{\revboth{Modified DeepONet Architecture}}
\label{sec:al_deeponet_implementation}

\revboth{To provide a fair comparison between methods, both DeepONet and FNO use similar spatio-temporal discretizations for their predictions, sacrificing some of the advantages of mesh-free predictions for improved efficiency.} %
\revboth{While calibrating the training hyperparameters and network architecture for the original DeepONet models, we observed that a relatively large number of basis functions were required for training to converge below $10\%$ $L^1$ relative error.  Specifically, we found values of $N>4096$ yielded the best performing models.  However, the memory footprint of evaluating the DeepONet trunk at all $128\times 128\times 76$ domain locations during evaluation proved prohibitively large.  Our experiments were conducted on Tesla V100 GPUs with $32$Gib of memory.  Storing the trunk outputs alone required $19$GiB of memory, and the GPU was unable to store both the model weights and trunk outputs at the same time.  This resulted in very inefficient evaluation times, and rendered the architecture impractical without further optimizations or a multi-GPU implementation.}

\revboth{To overcome this memory issue, and place the DeepONet models on more equal footing with respect to the FNOs, we elected to use a variant of the standard DeepONet architecture that leverages a discretized trunk.  Instead of training the trunk network in the form of a spatio-temporal function (i.e., $\operatorname{Trunk}: (x,t) \mapsto \{\varphi_n(x,t)\}_{n=1}^N$), we train the trunk network to create fixed basis functions on a predefined spatio-temporal grid: $ \operatorname{Trunk} \,\,  \sim \,\, \{\widehat{\varphi}_n\}_{n=1}^N \hspace{0.05in}$ with  $\hspace{0.05in} \widehat{\varphi}_n \, \in \, \mathbb{R}^{128\times 128 \times 76}$.}

\revboth{A complete description of the precise network architectures used for the AL experiments is provided in \ref{appendix:al_architectures}.}



\subsubsection{\revboth{Discussion on UQ Performance}}
\label{sec:al_uq}

\revboth{The uncertainty estimates for the DeepONet models remained fairly well correlated with the true errors throughout the AL process, but the correlation was noticeably weaker than that observed for the simpler elliptic problems in Section~\ref{sec:validation}.  At the end of the final learning iteration, the DeepONet models uncertainties were correlated with the observed errors on the full dataset with an $R^2$ value of $0.73$. The uncertainty estimates for the FNO models were considerably less reliable, with a typical $R^2$ value of $0.29$ at the final iteration.  The relatively poor calibration of the FNOs uncertainty estimates align with the accuracy results reported in Table~\ref{table:al_combined_results}, where the uncertainty-equipped FNOs achieve roughly the same accuracies as the standard models, although there is a small and consistent advantage observed across runs.  For the DeepONet models, the advantage is more pronounced, and this could be improved on further with better uncertainty-to-error calibration, as observed in the simpler setups.  The reduction in UQ accuracy likely stems from two fundamental factors: (i) the increase in problem complexity and (ii) the modifications made to the DeepONet architecture.} 
\revboth{As such, we plan to direct future research toward understanding} %
\revboth{the causes of the observed performance drop-off more fully.}

\begin{table}
\centering
{\fontsize{10}{10}\selectfont{
\begin{tabular}{ccc}
  Model &  $L^2$ relative error & Runtime \\
  \hline\Tstrut
  FNO      &  1.35\%   &  10.6 hr  \\
  NLL-FNO  &  1.41\%   &  11.9 hr  \\
  DeepONet &  2.53\%   &  6.0 hr  \\
  NLL-DeepONet  &  2.46\%   &  12.9 hr  
\end{tabular}
}}
\caption{\revboth{Summary of training results for wave equation using full dataset of $20,000$ PDE examples.}}
\label{table:results_full_dataset}
\end{table}

\subsubsection{Discussion on Runtimes}
\label{sec:al_timings}


\revboth{This PDE is complex enough to challenge memory and GPU resources, making uncertainty quantification (UQ) methods computationally demanding.} %
\revboth{Traditional ensemble-based UQ approaches are generally infeasible for problems of this scale: training a single model for one iteration takes on the order of an hour using GPU acceleration, and training an ensemble of 50+ models for UQ is unrealistic.} %

\revboth{Other frameworks like Bayesian (B)-DeepONet~\cite{lin2022b} offer rigorous theoretical frameworks for UQ. Unfortunately, B-DeepONets are computationally intensive for training and inferences, especially in the context of high-dimensional operators and AL. Even with recent advances in approximate inference for reduced training costs, such as variational inference (VI)~\cite{GARG2023105685, ortega2024variational} and Laplace approximations~\cite{ortega2024variational,deng2022accelerated}, reliance on expensive Monte Carlo sampling to estimate the posterior predictive distributions is prohibitively expensive. Much like ensembles, 50+ forward passes per input using a modified B-DeepONet architecture renders the `lightweight' aspect of our surrogates to be slower than the PDE solver itself.  Furthermore, standard mean-field approximations in VI often struggle to capture the complex spatial correlations inherent in operator outputs without imposing restrictive computational overheads. Consequently, our framework prioritizes a deterministic, single-pass approach that balances calibration quality with the strict latency constraints of real-time outer-loop applications.}

\revboth{While the AL results for both DeepONets and FNOs show consistent improvements over models trained with purely randomized samples, there is also a significant computational cost associated with training at each learning iteration.}  %
\revboth{The cumulative runtimes for performing AL with each model (averaged across runs) are shown in Table~\ref{table:results_average_runtimes}.  The uncertainty-equipped FNO models take roughly $30.7$ hours to complete, while the uncertainty-equipped DeepONet models take around $13.7$ hours.}
\revboth{Each PDE solve takes approximately $10$ seconds to complete using $8$ OpenMP threads, and the full dataset of $20,000$ examples took approximately $55.5$ hours to generate.} %
\revboth{For the AL setup, the models used a total of $750$ PDE solves, which require $2$ hours of compute time in addition to the model training times.}   

\revboth{The FNO models are able to achieve accuracies comparable to that of the model trained using the full dataset, while using only $750$ PDE examples during training.  The difference in the end-to-end runtimes between the active and full-dataset approaches are not as drastic, however, with the active setup taking approximately $30.3$ hours (including data generation) and the full-dataset model requiring between $62.8$ and $65.2$ hours, for standard and uncertainty-equipped models, respectively.}

\revboth{For the DeepONet models, the difference in runtimes was more noticeable, with AL taking around $15.6$ hours and the full-dataset model requiring between $59.8$ and $62.2$ hours.  But the performance of the full-dataset models were considerably better than that of the AL models, suggesting more AL iterations (and additional data generation) would need to be performed to produce models with comparable performance to those trained using the full dataset.}  %

\revboth{Balancing the tradeoffs between computational budgets and surrogate model accuracy is a key practical factor for many real-world applications.  The numerical results presented here show some potential advantages of employing an AL strategy for training operator networks on families of PDEs, but the results are not as clear-cut as one may have hoped for.  It is possible that the framework would have more pronounced advantages for more complex PDE systems, where the solver times become more significant.  We also not that for experimental setups with humans in the loop the additional computational costs required for training would almost certainly be offset by reducing the number of experiments that need to be conducted.  
}





\begin{table}
\centering
{\fontsize{10}{10}\selectfont{
\begin{tabular}{ccccccccccc}
Iteration & 1 & 2 & 3 & 4 & 5 & 6 & 7 & 8 & 9 & 10 \\ 
 \hline \Tstrut 
FNO & 1.6  & 3.5  & 5.7  & 8.3  & 11.2  & 14.2  & 17.3  & 20.8  & 24.7  & 28.5  \\ 
NLL-FNO\, & 1.6  & 3.4  & 5.4  & 7.9  & 10.6  & 13.5  & 16.7  & 20.3  & 24.2  & 28.3  \\ 
DeepONet & 0.6  & 1.2  & 1.8  & 2.6  & 3.4  & 4.3  & 5.2  & 6.2  & 7.2  & 8.3  \\ 
\!NLL-DeepONet\! & 0.8  & 1.7  & 2.7  & 3.9  & 5.2  & 6.7  & 8.2  & 9.9  & 11.7  & 13.6  
\end{tabular}
}}
\caption{\revboth{\revboth{Summary of cumulative training time (hours) for wave equation AL procedures averaged across runs.}}}
\label{table:results_average_runtimes}
\end{table}

\section{Conclusions}
\label{sec:conclusion}

In this work, we conducted a series of experiments investigating a light-weight uncertainty quantification framework for deep operator networks.  %
We trained models using a log-likelihood loss function to calibrate predictive uncertainty estimates to the errors observed at the tail-end of training. %
We evaluated the models on a variety of linear and non-linear PDE problems and
explored several architecture variants to identify the best configurations for incorporating boundary conditions and uncertainty estimates. %
We also introduced an optimized computational graph that yields an order-of-magnitude reduction in evaluation time and enables fast operator inference for real-time applications. %
We presented a detailed statistical analysis of the UQ framework %
and verified that the predictive uncertainties accurately reflect the observed error distributions. %
We showed the uncertainty estimates also generalize well to out-of-distribution data by evaluating the models on function spaces unseen during training. %

{\revboth{
We demonstrated how the uncertainty-equipped operator networks can be used to accelerate outer-loop BO procedures.  
We first considered a simple, easy to visualize, proof-of-concept problem and demonstrated how the predictive uncertainties can significantly improve data-acquisition processes for certain outer-loop processes. %
This resulted in nearly an order-of-magnitude improvement over standard DeepONet models. %
Finally, we conducted an extensive set of numerical experiments for an AL problem involving systems governed by hyperbolic PDEs on an irregular domain.  We showed that both DeepONet and FNO models can be effectively extended to provide predictive uncertainties that improve AL performance relative to the baseline models.  %
In future work, we hope to provide a more comprehensive analysis of the predictive uncertainties for more operator surrogate architectures that includes assessments of the uncertainty calibration and coverage.  We also plan to investigate more complex PDE systems including variable coefficients, as well as discontinuities and shocks, to assess how well the AL procedures examined here generalize to more challenging real-world problems.
}}


%

\vspace{0.1in}
\noindent  \textbf{Acknowledgments}\quad\footnotesize This paper describes objective technical results and analysis. Any subjective views or opinions that might be expressed in the paper do not necessarily represent the views of the U.S. Department of Energy or the United States
Government. Sandia National Laboratories is a multimission laboratory managed and operated by National Technology and Engineering Solutions of Sandia LLC, a wholly owned subsidiary of Honeywell International, Inc., for the U.S. Department of Energy's National Nuclear Security Administration under contract DE-NA-0003525 (SAND2026-15586O). 
GL would like to thank the support of National Science Foundation (DMS-2533878, DMS-2053746, DMS-2134209, ECCS-2328241, CBET-2347401 and OAC-2311848), and U.S.~Department of Energy (DOE) Office of Science Advanced Scientific Computing Research program DE-SC0023161, the SciDAC LEADS Institute, the Uncertainty Quantification for Multifidelity Operator Learning (MOLUcQ) project (Project No. 81739), and DOE–Fusion Energy Science, under grant number: DE-SC0024583.
LL was supported by the U.S. DOE Office of Advanced Scientific Computing Research under Grants No.~DE-SC0025593 and No.~DE-SC0025592 and the U.S. National Science Foundation under Grants No.~DMS-2347833 and No.~DMS-2527294.

\appendix

\section{Implementation details}
\label{sec:implementation_appendix}

\setcounter{figure}{0}
\renewcommand{\theequation}{A\arabic{equation}}
\setcounter{equation}{0}

DeepONet models can be equipped with predictive uncertainties by  %
(1) selecting a parameterized family of distributions, (2) extending the network outputs to provide estimates for each parameter, and (3) defining a loss function based on the negative log marginal likelihood. %
Once these modifications have been made, uncertainty-equipped DeepONet models are trained using the same procedure as standard DeepONet models.  %
In practice, some additional adjustments to the input data are sometimes required to prevent the training procedure from diverging at the start of training due to numerical overflow.  %
For example, when using a normal distribution to model the network's output, small variance predictions can lead to extremely large loss values (which can result in overflow and stall training completely).   %
This can be avoided by rescaling input data so that the initial outputs of the networks are on a reasonable scale (e.g., on the order of $-10$ to $10$ for log standard deviation predictions).  %
This rescaling step is generally sufficient, and it was all that was required for the experiments in this work. %
For other applications, it is possible that training could destabilize later on in the training procedure due to excessively low variance predictions. %
In this case, clipping can be used to impose an absolute limit on the minimum variance value output by the network. %

\subsection{Network variants for boundary conditions and uncertainty quantification}
\label{sec:deeponet_boundary_conditions}

\begin{figure}[htb]
  \centering
  \includegraphics[width=0.985\textwidth]{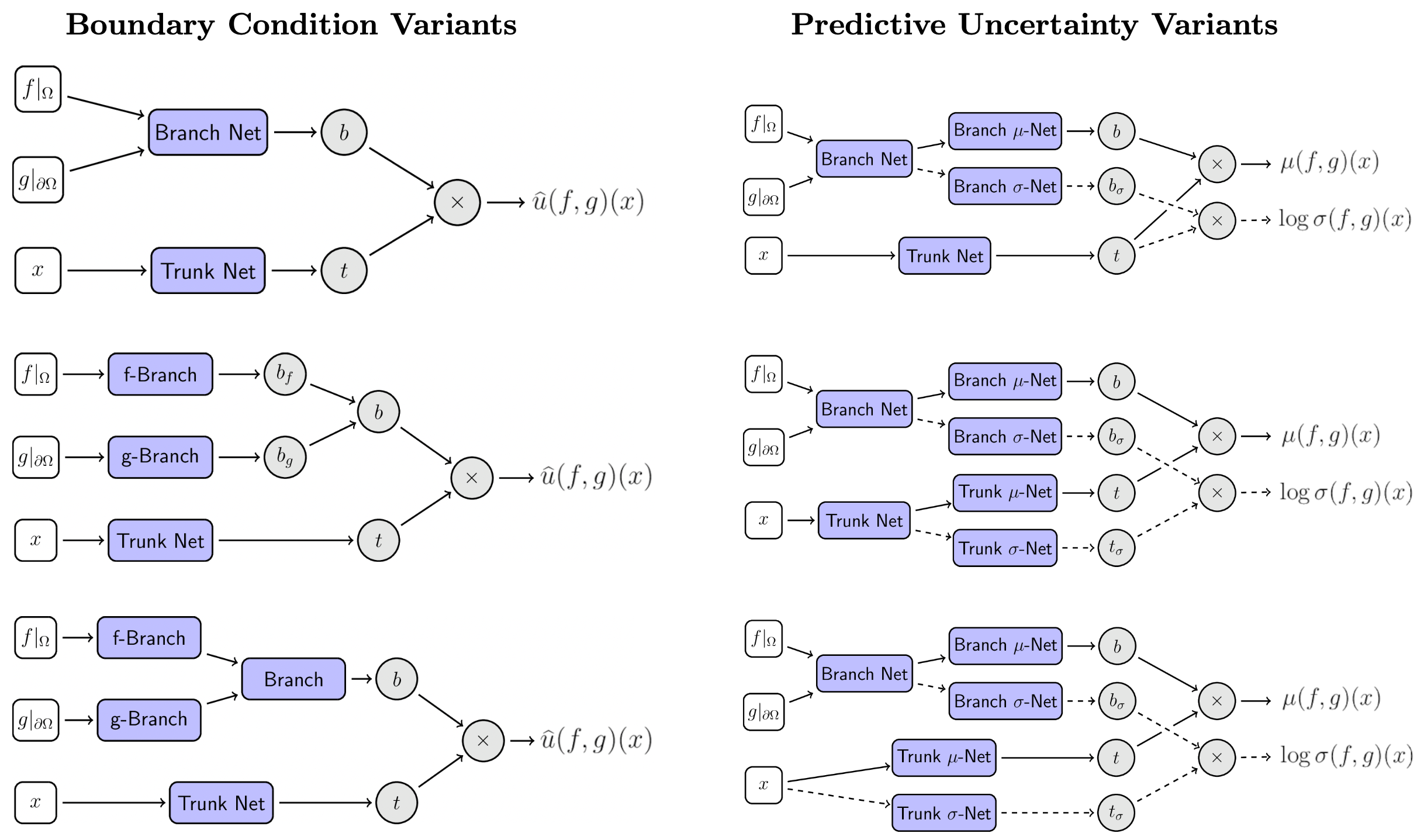} 
  \caption[DeepONet BC and UQ architecture variations.]{DeepONet architecture variants for the incorporation of boundary conditions (left) and the prediction of uncertainty parameters (right).  The simplest variant for handling boundary conditions (top-left) was found to match the performance of more structured architectures. %
    For uncertainty-equipped operator networks, we found that decoupling the trunk predictions for mean and uncertainty parameters consistently improved the performance of the network.  The second variant on the right was selected since it introduces far less trainable parameters than independent uncertainty networks and was observed to achieve similar levels of performance. 
  }
  \label{fig:architecture_variants}
\end{figure}

One technical challenge associated with 
neural network approximations for PDEs is the incorporation of boundary conditions. %
Handling input data associated with inhomogeneous boundary conditions is complicated
by the fact that the dimensionality of boundary data is strictly less than that of the interior data, as shown in Fig.~\ref{fig:input_data}. %
This leads to a sparse representation of the boundary data in the ambient space and poses a significant challenge for convolutional architectures in practice. 
For fully-connected layers, the boundary can simply be appended to the list of sensor values associated with the interior data. %
We conducted a series of ablation studies to determine the appropriate network architecture for handling inhomogeneous boundary conditions. The architecture illustrated at the top left of Fig.~\ref{fig:architecture_variants}  %
was ultimately chosen due to its simplicity; while the other two variations achieved similar performance, 
there was no observable benefit associated with separate processing components used for the 
interior data, $f\in C(\Omega)$, and boundary data, $g\in C(\partial \Omega)$.

Once the architecture for the inclusion of boundary conditions was selected, additional ablation experiments were conducted in order to identify the best network design regarding uncertainty predictions. %
We hypothesized that the performance of the uncertainty framework could be improved by decoupling the mean and standard deviation predictions during the final layers of the branch and trunk networks.  %
This is motivated by the fact that the output of the trunk network serve as basis functions for the final model predictions.  By separating the trunk at the end, it is possible for the network to learn a distinct set of basis functions for its uncertainty predictions (instead of relying on the basis functions learned for the mean predictions). %
We considered three architecture variations for decoupling the uncertainty predictions, which are shown in the right column of Fig.~\ref{fig:architecture_variants}. 
The first variant uses a shared trunk network while incorporating a split head in the branch network to decouple the mean and uncertainty predictions.  This architecture consistently performed the worst among the three variants.  %
The remaining architectures also decouple predictions in the trunk network, with the first doing so in the final layers of the network and the latter incorporating an entirely separate network for trunk uncertainty predictions. 
We observed similar performance with both variants and elected to use the simpler architecture (shown in the center-right of Fig.~\ref{fig:architecture_variants}) since it requires less trainable parameters (and significantly less overhead) than the fully-decoupled alternative. %

\clearpage
\section{Bayesian Optimization Training Procedure}
\label{appendix:bo_training}

\setcounter{figure}{0}
\renewcommand{\thefigure}{B\arabic{figure}}
\setcounter{table}{0}

\revboth{{\bf{Parameter-to-Function Problem Setup. }}  {\emph{Network architecture.}} The DeepONet model used for the parameter-to-function problem setup used a standard, fully-connected network architecture with node counts of $[40, 40, 40]$ for both the trunk and branch networks and ReLU activation functions.  For the NLL-DeepONet models, the final layer was split into two independent layers, each with $40$ nodes, to provide mean and standard deviation outputs for the branch and trunk.}

\revboth{{\emph{Active Learning Setup. }} The active learning setup begins with a randomized set of $50$ training examples. The models are trained using an Adam optimizer with learning rate $1.0e\mathrm{-}4$ for $15$ thousand epochs, trained for an additional $15$ thousand epochs after dropping the learning rate to $5.0e\mathrm{-}5$, and then trained for $15$ thousand epochs with the final learning rate of $1.0e\mathrm{-}5$.  At the end of each training iteration, the operator networks are evaluated on a collection of $50,000$ candidate input parameter pairs; the $10$ parameters that are assigned the highest acquisition values are added to training dataset to complete the active learning iteration, and this process is repeated for $20$ iterations.  To avoid potential complications with local minima and lingering influences of early training iterations on the final results, we reinitialized the network weights and optimizers between each active learning iteration. }

\vspace{0.1in}
\revboth{{\bf{Function-to-Function Problem Setup. }}  {\emph{Network architecture.}} The DeepONet model used for the parameter-to-function problem setup used a standard, fully-connected network architecture with node counts of $[40, 40, 40, 40, 50]$ for both the trunk and branch networks and ReLU activation functions.  For the NLL-DeepONet models, the final layer was split into two independent layers, each with $50$ nodes, to provide mean and standard deviation outputs for the branch and trunk.}

\revboth{{\emph{Active learning setup.}} The active learning setup begins with a randomized set of $10$ training examples. The models are trained using an Adam optimizer with learning rate $1.0e\mathrm{-}4$ for $5$ thousand epochs, trained for an additional $5$ thousand epochs after dropping the learning rate to $5.0e\mathrm{-}5$, and then trained for $5$ thousand epochs with the final learning rate of $1.0e\mathrm{-}5$.  At the end of each training iteration, the operator networks are evaluated on a collection of $50,000$ candidate input functions; the $5$ input functions that are assigned the highest acquisition values are added to training dataset to complete the active learning iteration, and this process is repeated for $20$ iterations.  To avoid potential complications with local minima and lingering influences of early training iterations on the final results, we reinitialized the network weights and optimizers between each active learning iteration.  }

%
%
%
%


\section{Active Learning Training Procedure}
\label{appendix:al_training}

\setcounter{figure}{0}
\renewcommand{\thefigure}{C\arabic{figure}}
\setcounter{table}{0}

\subsection{Advection-Diffusion Setup}

\revboth{For each of the models trained for the active learning results, we used the Adam optimization algorithm~\cite{kingma2014adam} to calibrate model weights.  Each model was trained using a batch size of $32$, an initial learning rate of $2.5e\textrm{-}4$, and exponential decay by a factor of $0.99$ applied to the learning rate until reaching a minimum limit set to $5.0e\textrm{-}5$.  The models were trained for $150$ epochs during each active learning iteration, for a total of $10$ training iterations.  The MC dropout model was trained for $350$ epochs each iteration in an attempt to improve the accuracy of these models; however, even with the extended training period, the models provided minimal improvements compared with the alternatives. }

\revboth{After each active learning iteration (i.e., once new data points are selected and added to the training data), the model is initialized with the same weights from the previous iteration.  The optimizer state and learning rates are reset before training resumes, as we found this led to better performance than that obtained by resuming training with the same optimizer.  }

\subsection{Wave Equation Setup}
\revboth{For each of the models trained for the active learning results, we used the Adam optimization algorithm to calibrate model weights.  For the uncertainty-equipped DeepONet model, the initial learning rate was set to $2.5e\textrm{-}4$; for all other models, the initial rate was set to $5.0e\textrm{-}4$.}

\revboth{The FNO models were trained using a batch size of $4$ (which we found resulted in considerably better performance compared with larger batch sizes) and were trained for $1000$ epochs during each active learning iteration.  }

\revboth{The DeepONet models were trained using a batch size of $32$ and were trained for $1500$ epochs; training was observed to level off after this number of epochs, and training times were still substantially shorter than FNOs despite the increased number of epochs. }

\revboth{After each epoch, the learning rate was decayed exponentially until reaching a minimum value, after which it remained fixed.  For the uncertainty-equipped models the decay rate was set to $0.99$ with a minimum limit set to $5.0e\textrm{-}5$.  For models without uncertainty, the decay rate was set to $0.985$ with a minimum limit set to $1.0e\textrm{-}4$.  This difference was motivated by the observation that UQ-based training tends to work better at slightly lower learning rates for the final stage of training, but takes slightly longer to make its initial descent (which we account for by slightly reducing the initial decay rate).}

\revboth{After each active learning iteration (i.e., once new data points are selected and added to the training data), the model is initialized with the same weights from the previous iteration.  The optimizer state and learning rates are reset before training resumes, as we found this led to better performance than that obtained by resuming training with the same optimizer.  }

\section{Active Learning Network Architectures}
\label{appendix:al_architectures}

\setcounter{figure}{0}
\renewcommand{\thefigure}{D\arabic{figure}}
\setcounter{table}{0}

\subsection{FNO Models for Advection-Diffusion}
\label{appendix:advection_architectures}

\revboth{The FNO models used for the advection-diffusion active learning problem were constructed based on the implementations provided in NVIDIA's open-source PhysicsNeMo package~\cite{contributors2023nvidia}.  We use the predefined `$\texttt{SpectralConv2d}$' layers to encode the input function into a latent representation.  More precisely, we used four spectral convolutional layers with $150$ latent channels, $15$ Fourier modes, and Gaussian error linear unit (GeLU) activations~\cite{hendrycks2016gaussian}.  After applying the spectral encoder, we implemented a decoder consisting of three convolutional layers with kernel size $3$ and output channel counts of $[32, 32, 76]$.  All decoder layers used SiLU activation functions.}

\vspace{0.085in}\revboth{{\bf{NLL Uncertainty Predictions.}}} %
\revboth{To extend standard FNO models to provide uncertainty predictions, we tested two broad classes of architectures.  In the first approach, we experimented with different network layers applied after the spectral encoder and in the second approach, we implemented a standalone network component for UQ predictions (i.e., a network that takes the raw function values as input and produces UQ estimates directly from the input, without any connection to the mean prediction).}

\revboth{For FNOs, we found that the standalone UQ predictions resulted in the best performance.  The final architecture we arrived at begins with a series of three convolutional layers with kernel size $3$ and channel counts of $[16, 16, 16]$.  Max pooling layers were applied after the first and last convolutional layers, and the output was then processed by a series of four fully connected layers with node counts $[256, 128, 128, 36]$.  The final $36$ output values are then interpreted as the log standard deviation predictions for the network output at each time step, and the loss is calculated using Equation~\ref{eq:prob_loss}.}

\vspace{0.085in}\revboth{{\bf{Monte Carlo Dropout.}}} %
\revboth{The MC dropout model incorporated dropout before each of the final three convolutional layers using PyTorch's `$\texttt{Dropout2d}$' layers.  After initial tests with higher dropout rates, we found a relatively low rate of $0.05$ produced the best performance.  During evaluation, the network was evaluated $25$ times with dropout still enabled and the mean and standard deviation statistics were computed from these empirical samples at each prediction location.}

\vspace{0.085in}\revboth{{\bf{FNO Ensembles.}}} %
\revboth{We conducted a series of preliminary tests to identify the appropriate ensemble size for the advection-diffusion problem.  Based on these tests, we chose to use an ensemble of $15$ FNO models, which provides a reasonable balance of tradeoffs between sample size and computational costs. Each model was trained independently during each active learning iteration.  When performing inference, the models were all evaluated independently and the mean and standard deviation statistics were computed from the collection of outputs of the ensemble at each prediction location. }

\subsection{DeepONet Models for Wave Equation}

\revboth{The DeepONet models used for the active learning experiments were implemented using PyTorch~\cite{paszke2019pytorch}.  To improve efficiency, we elected to use convolutional layers to process the structured inputs to the branch network.  Due to GPU memory constraints, we could not efficiently evaluate large datasets with a standard trunk architecture.  We employed a discrete-trunk variant of the standard DeepONet architecture to overcome this, which we describe in detail below.}

\vspace{0.085in}\revboth{{\bf{Discrete-Trunk Implementation.}}} %
\revboth{The standard DeepONet trunk network is designed to produce basis function $\{\varphi_n\}_{n=1}^N$ that can be evaluated at arbitrary locations on the problem domain.  This has several clear-cut advantages (e.g., bypassing the need for interpolation and providing stable gradient calculations), but it can also be cumbersome to evaluate when predicting at high resolutions.} 
\revboth{In contrast to DeepONets, standard FNOs implicitly assume a structured grid representation on the output locations where model predictions are made.  In the case of our active learning setup, the predictions assume a fixed shape $(128, 128, 76)$ corresponding to a grid across the two spatial variables and time variable.}

\revboth{We can incorporate this same assumption into DeepONet models by replacing the basis functions produced by a standard trunk with {\emph{basis matrices}} $\widehat{\varphi}_n \, \in \, \mathbb{R}^{128\times 128 \times 76}$.  In this setup, the trunk no longer requires coordinate information as input; it will instead attempt to learn a global set of basis matrices directly.  To implement this, we start with a fixed input tensor filled with ones (chosen to be shape $(4, 4)$ for our experiments, although this choice has minimal impact on model performance).  The fixed input tensor is then processed by a series of fully connected layers with node counts $[256, 256, 256, 512, 512]$ and sigmoidal linear unit (SiLU) activations~\cite{hendrycks2016gaussian}.}  %
\revboth{The $512$ hidden units produced by the fully connected layers are then reshaped to an $8 \times 8 \times 8$ tensor which is fed into a series of four 2D convolutional and bilinear upsampling layers to produce a final collection of basis matrices stored in a tensor of shape $(N, 128, 128, 76)$.  The first two convolutional layers use kernel sizes of $3$ with output channels set to $16$, and the final two convolutional layers use kernel sizes of $5$ with output channels set to $32$ and $N$, respectively. }


\vspace{0.085in}\revboth{{\bf{Convolutional Branch Architecture.}}} %
\revboth{Since the input functions for the active learning problem are naturally structured on a $128\times 128\times 76$ grid, it is natural to use convolutional layers for the initial processing stages in the branch network.  For our experiments, we used a branch network consisting of four 2D convolutional and max pooling layers with kernel sizes of $3$ and output channel counts of $16$.  These layers are followed by a sequence of fully connected layers with hidden node counts $[1028, 1028, N]$.  All layers used SiLU activation functions. }

\revboth{In the discrete-trunk setup, the output of the branch network is treated just as it is in a standard DeepONet setup: each component of the output is a weight for a corresponding basis produced by the trunk.  The final model output is still defined by $\widehat{u} = \sum_{n=1}^N b_n \cdot \widehat{\varphi}_n$ where $b^T = \operatorname{Branch}(f)$ and $\varphi = \operatorname{Trunk()}$, and the only practical difference is that the trunk outputs are now matrices. }


\vspace{0.085in}\revboth{{\bf{Uncertainty Predictions.}}} %
\revboth{Since the active learning procedure only requires function-level uncertainty predictions, we implemented a simplified form of predictive uncertainty which produces a single scalar UQ value for each time step and function input.   For the branch network, the initial processing layers are shared with the mean prediction.  The output of the first fully connected layer is then taken from the mean branch, and fed into a series of separate fully connected layers with node counts $[128, 128, 1]$ and SiLU activations.  This scalar output is used as a scaling factor for the time-dependent uncertainty profile output from the trunk. }


\revboth{The trunk UQ predictions are learned using a separate network from the mean prediction, but using the same initial tensor of ones as the trunk mean.  The network consists of a series of fully connected layers with node counts $[128, 128, 128, 128, 128, 76]$ and SiLU activations. The $76$ outputs from the trunk are scaled by the branch network's UQ output to produce the final uncertainty estimates.  The resulting $76$ output values are then interpreted as the log standard deviation predictions for the network output at each time step, and the loss is calculated using Equation~\ref{eq:prob_loss}.}

\subsubsection{FNO Models for Wave Equation}

\revboth{For the FNO models, we used the implementations provided in NVIDIA's open-source PhysicsNeMo package~\cite{contributors2023nvidia}.  We use the predefined `$\texttt{SpectralConv2d}$' layers to encode the input function into a latent representation.  More precisely, we used four spectral convolutional layers with $150$ latent channels, $15$ Fourier modes, and Gaussian error linear unit (GeLU) activations~\cite{hendrycks2016gaussian}.  After applying the spectral encoder, we implemented a decoder consisting of three convolutional layers with kernel size $3$ and output channel counts of $[32, 32, 76]$.  All decoder layers used SiLU activation functions.}

%

\vspace{0.085in}\revboth{{\bf{Uncertainty Predictions.}}} %
\revboth{To extend standard FNO models to provide uncertainty predictions, we tested two broad classes of architectures.  In the first approach, we experimented with different network layers applied after the spectral encoder and in the second approach, we implemented a standalone network component for UQ predictions (i.e., a network that takes the raw function values as input and produces UQ estimates directly from the input, without any connection to the mean prediction).}

\revboth{For FNOs, we found that the standalone UQ predictions resulted in the best performance.  The final architecture we arrived at begins with a series of three convolutional and max pooling layers, with kernel size $3$ and channel counts of $[16, 16, 16]$.  The output of the convolutional layers is then processed by a series of four fully connected layers with node counts $[1028, 128, 128, 76]$.  The final $76$ output values are then interpreted as the log standard deviation predictions for the network output at each time step, and the loss is calculated using Equation~\ref{eq:prob_loss}.}

\section{Active Learning Numerical Results}
\label{appendix:al_table_results}

\setcounter{figure}{0}
\renewcommand{\thefigure}{E\arabic{figure}}
\setcounter{table}{0}

\begin{figure}[hbt]
  \centering
  \begin{tabular}{cccc}
    \hspace{-0.05in}\includegraphics[width=0.21\textwidth]{./solution_1_5_crop.png} &
    \hspace{-0.02in}\includegraphics[width=0.21\textwidth]{./solution_1_15_crop.png} &
    \hspace{-0.02in}\includegraphics[width=0.21\textwidth]{./solution_1_25_crop.png} &
    \hspace{-0.02in}\includegraphics[width=0.21\textwidth]{./solution_1_35_crop.png}  \\
    \includegraphics[width=0.22\textwidth]{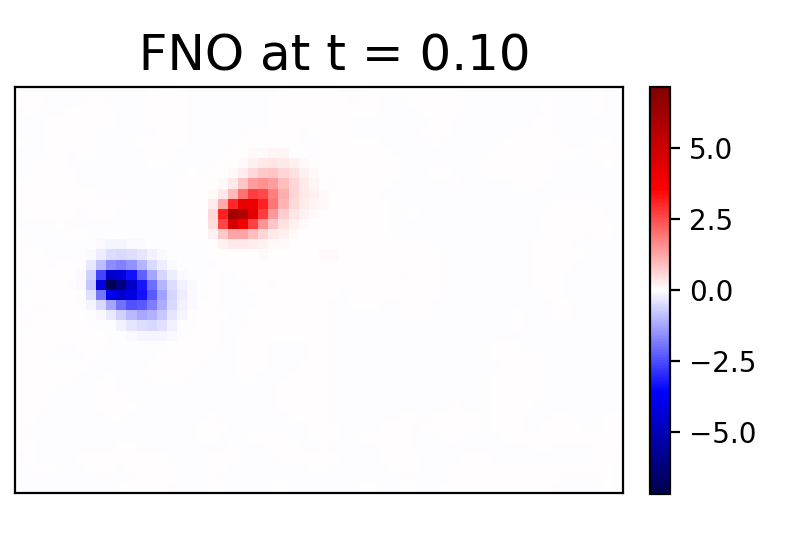} & 
    \includegraphics[width=0.22\textwidth]{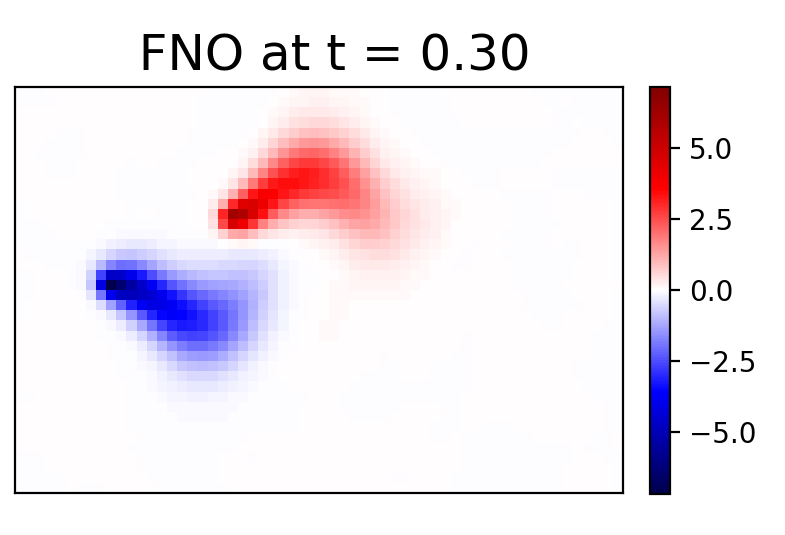} & 
    \includegraphics[width=0.22\textwidth]{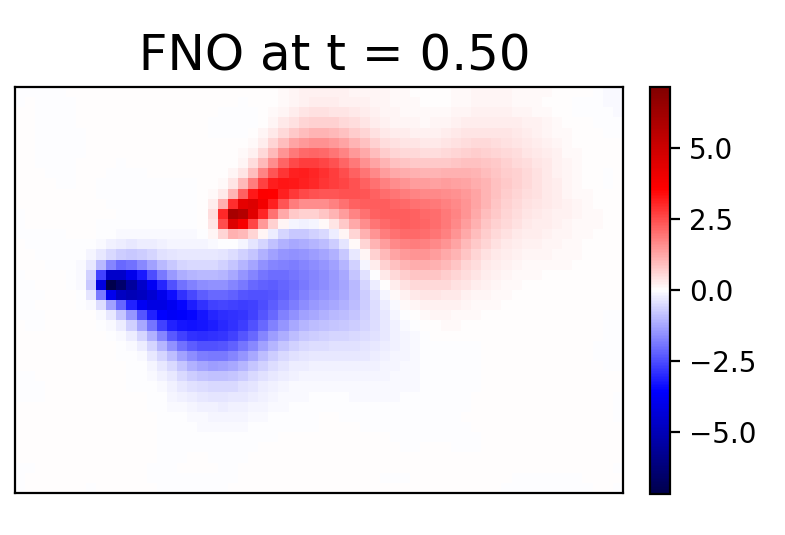} & 
    \includegraphics[width=0.22\textwidth]{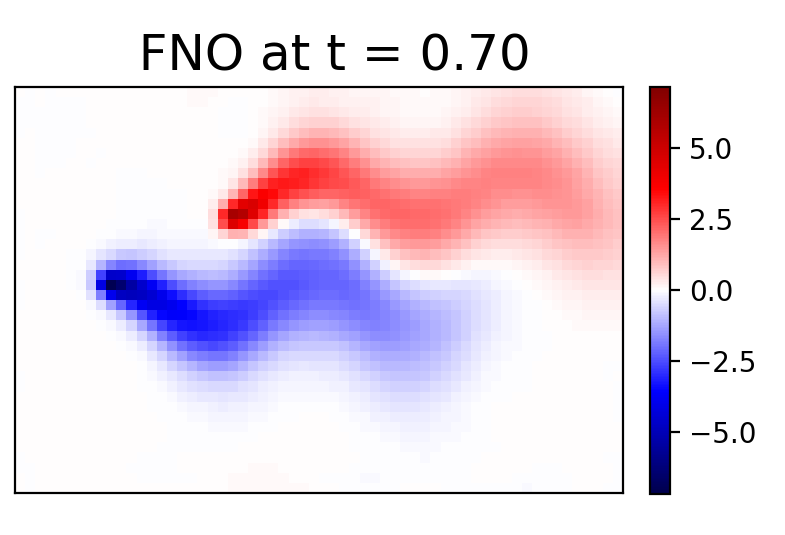}  \\
    \hspace{-0.05in}\includegraphics[width=0.21\textwidth]{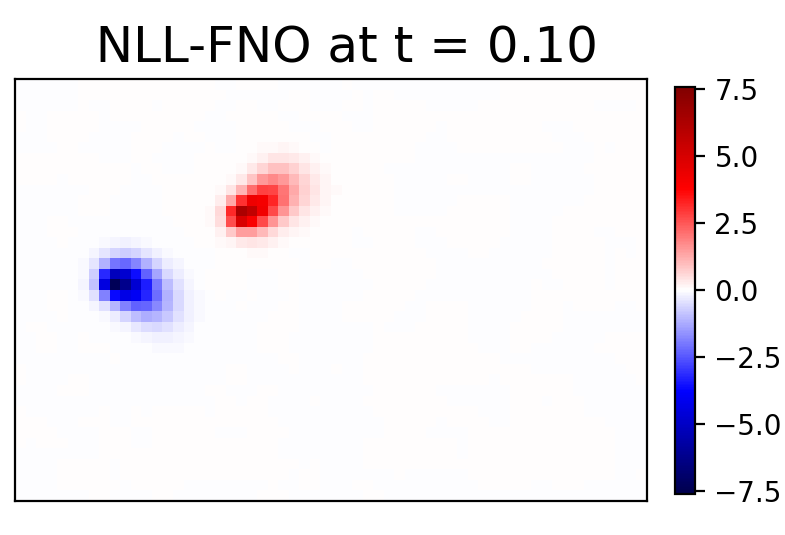} &
    \hspace{-0.02in}\includegraphics[width=0.21\textwidth]{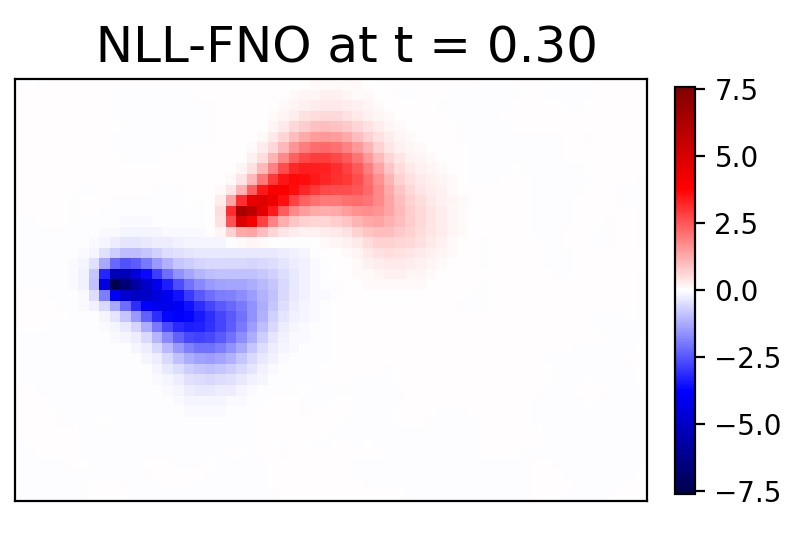} &
    \hspace{-0.02in}\includegraphics[width=0.21\textwidth]{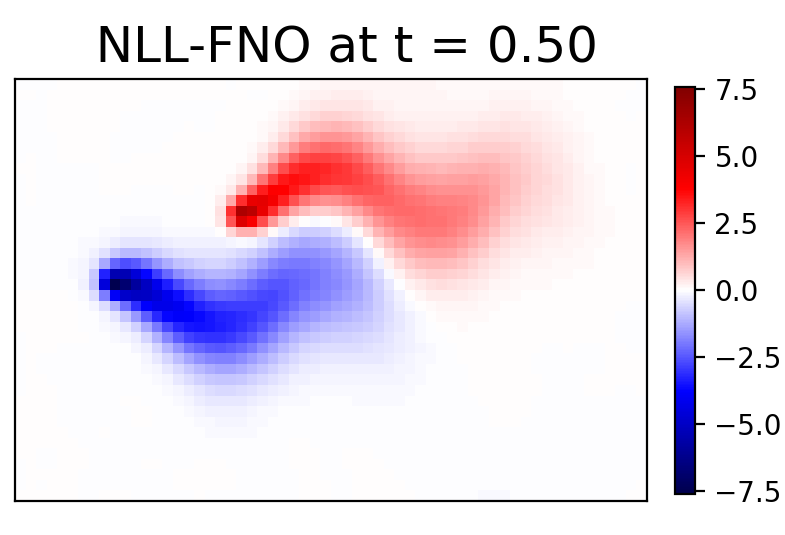} &
    \hspace{-0.02in}\includegraphics[width=0.21\textwidth]{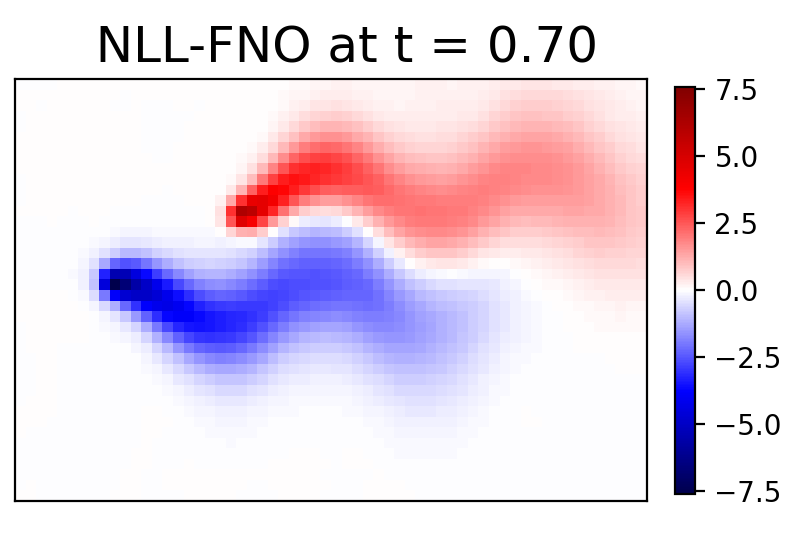}  \\    
    \includegraphics[width=0.22\textwidth]{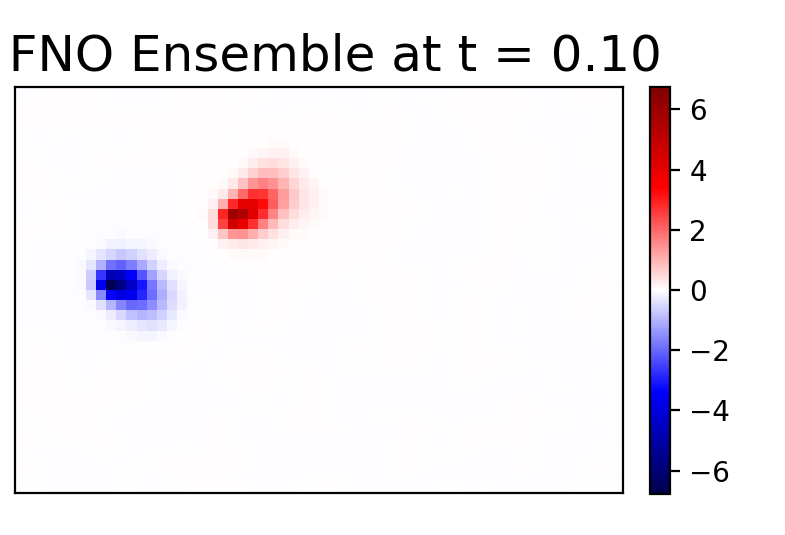} & 
    \includegraphics[width=0.22\textwidth]{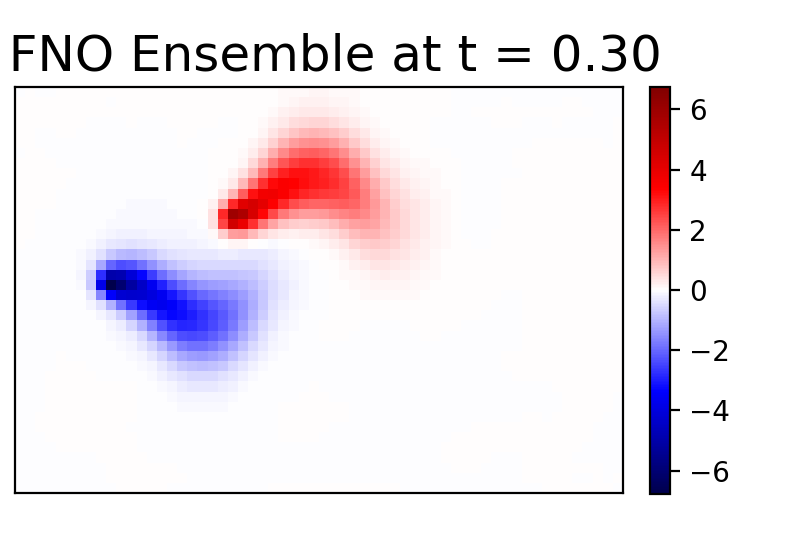} & 
    \includegraphics[width=0.22\textwidth]{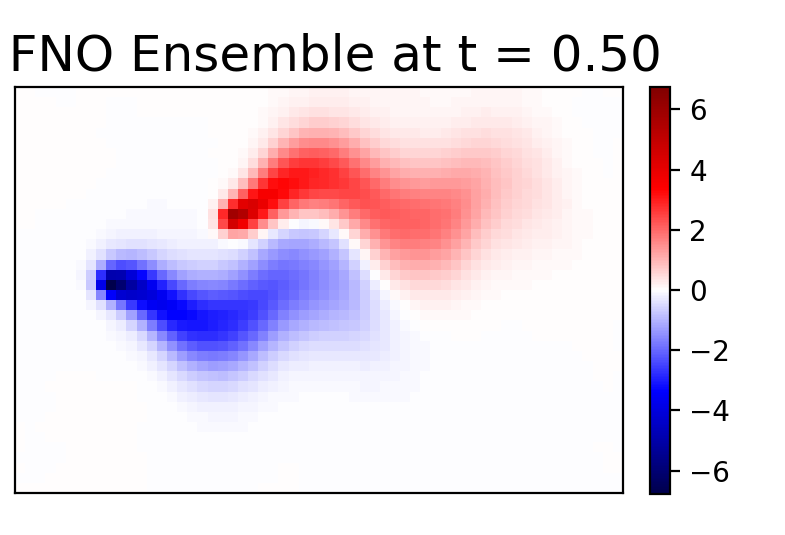} & 
    \includegraphics[width=0.22\textwidth]{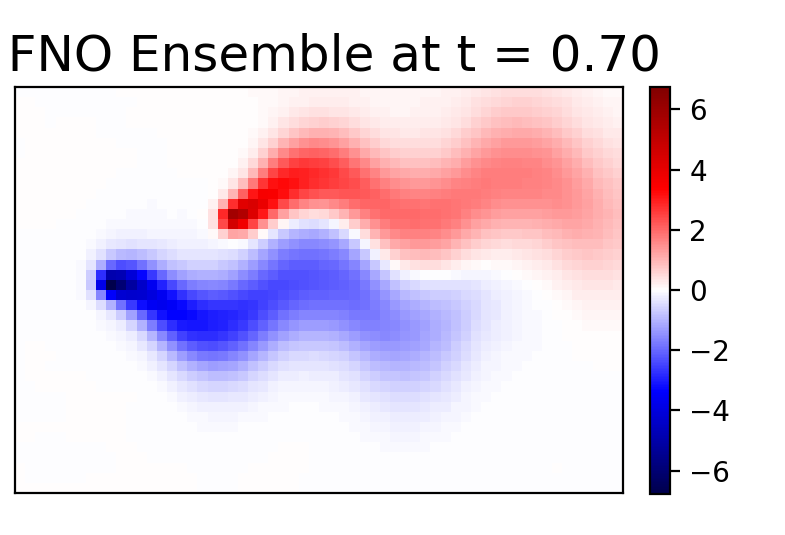}  \\
    \includegraphics[width=0.22\textwidth]{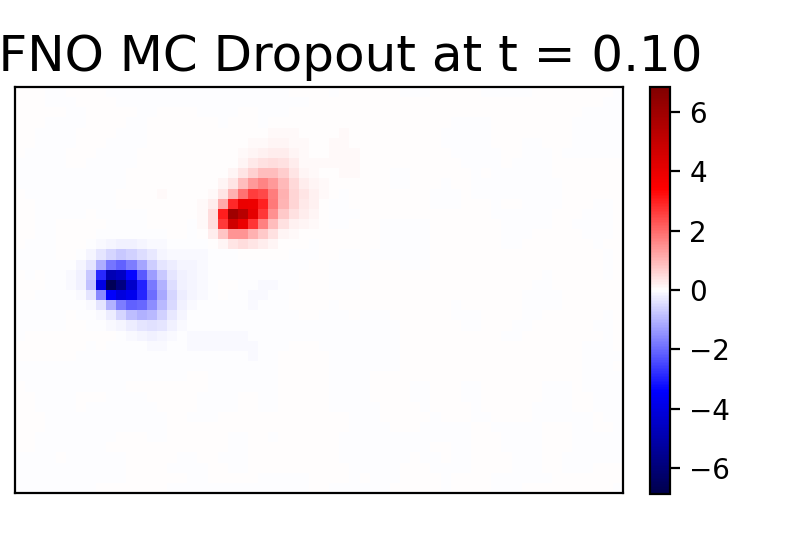} & 
    \includegraphics[width=0.22\textwidth]{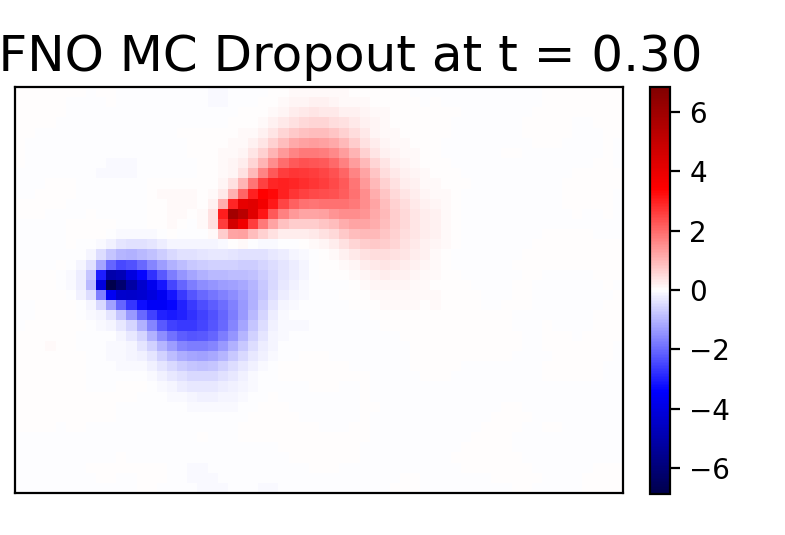} & 
    \includegraphics[width=0.22\textwidth]{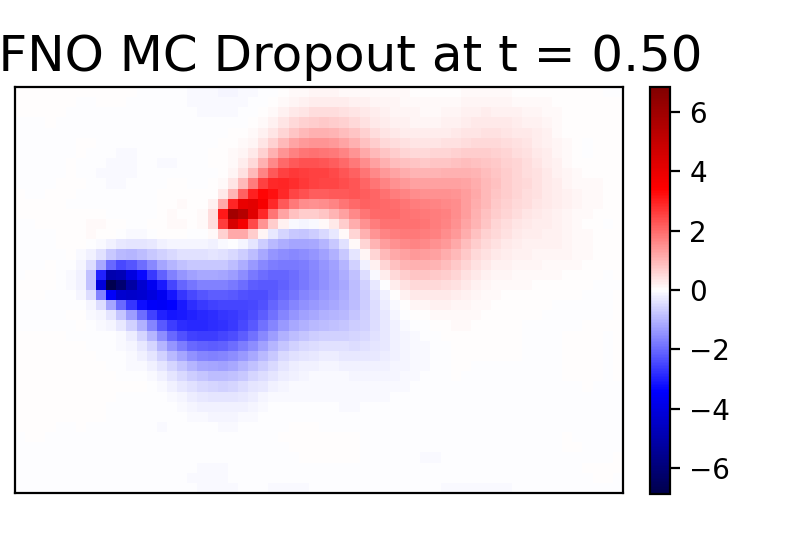} & 
    \includegraphics[width=0.22\textwidth]{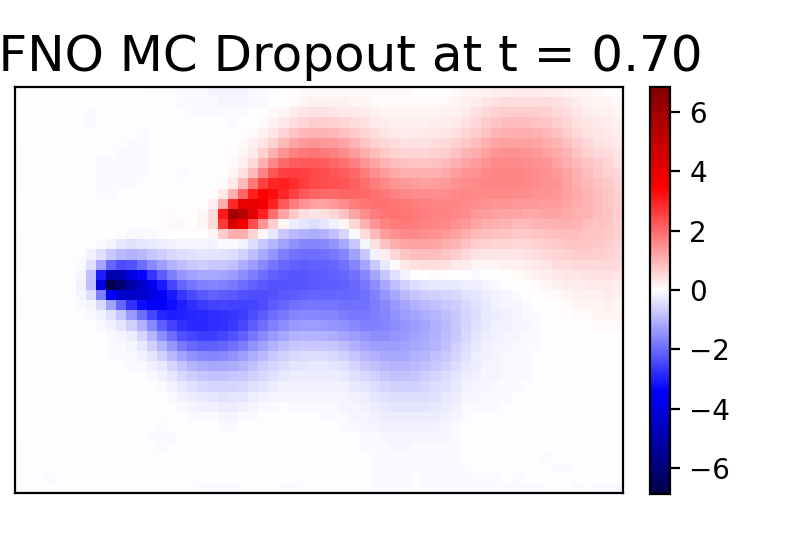}  \\
  \end{tabular}
  \caption{\revboth{Example solution and operator network predictions for the advection-diffusion problem setup.}}
  \label{fig:advection_qualitative}
\end{figure}

\begin{figure}
  \centering
  \begin{tabular}{cccc}
    \includegraphics[width=0.22\textwidth]{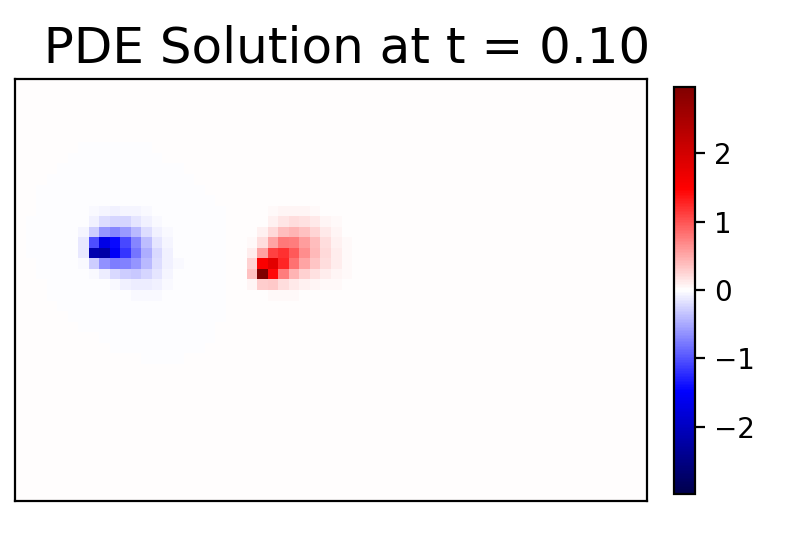} & 
    \includegraphics[width=0.22\textwidth]{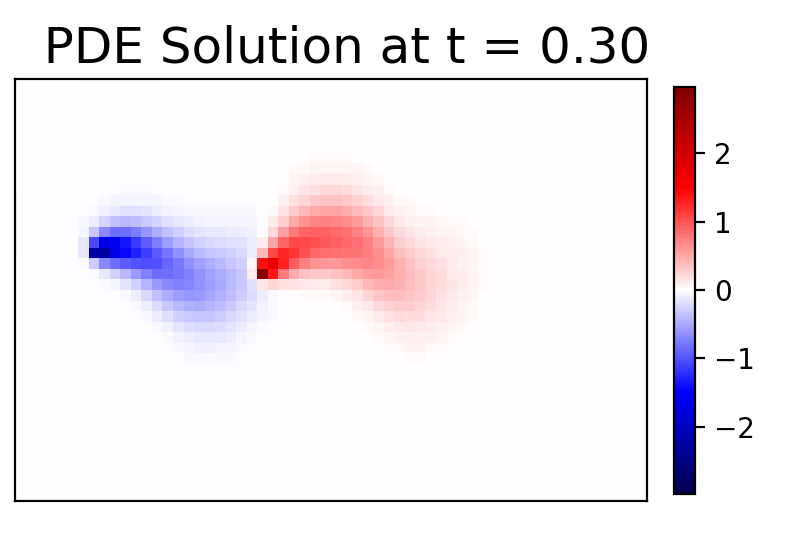} & 
    \includegraphics[width=0.22\textwidth]{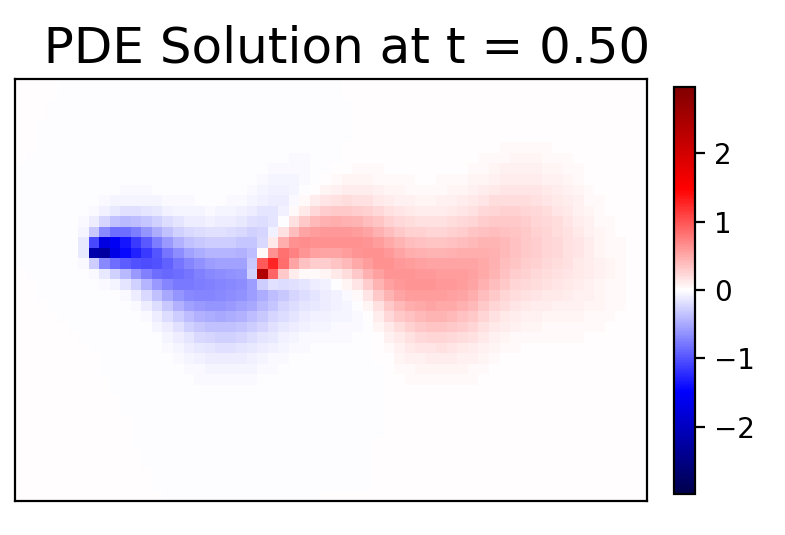} & 
    \includegraphics[width=0.22\textwidth]{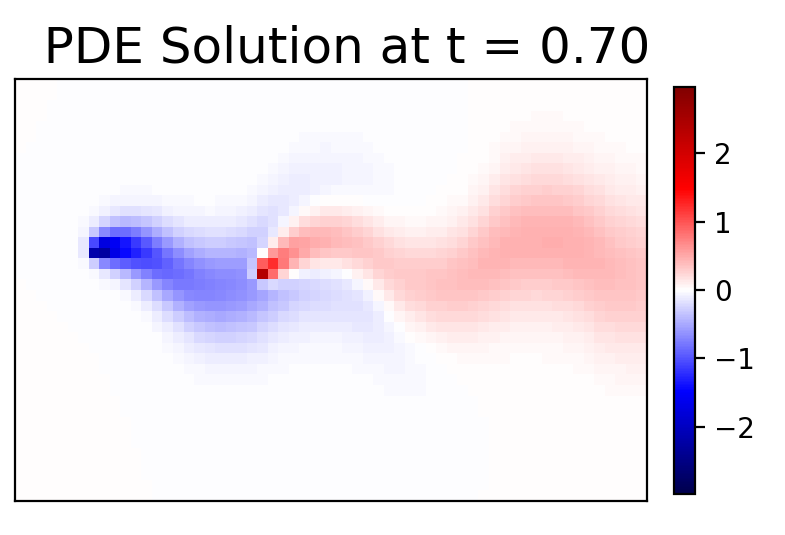}  \\
    \includegraphics[width=0.22\textwidth]{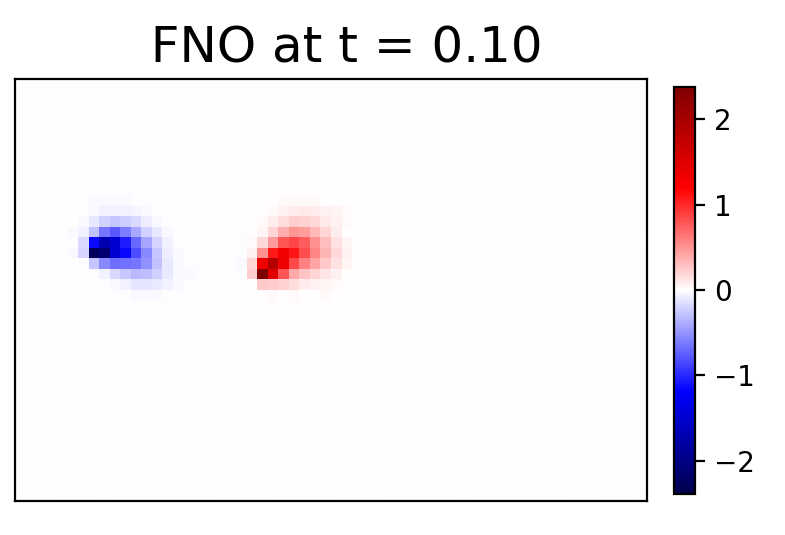} & 
    \includegraphics[width=0.22\textwidth]{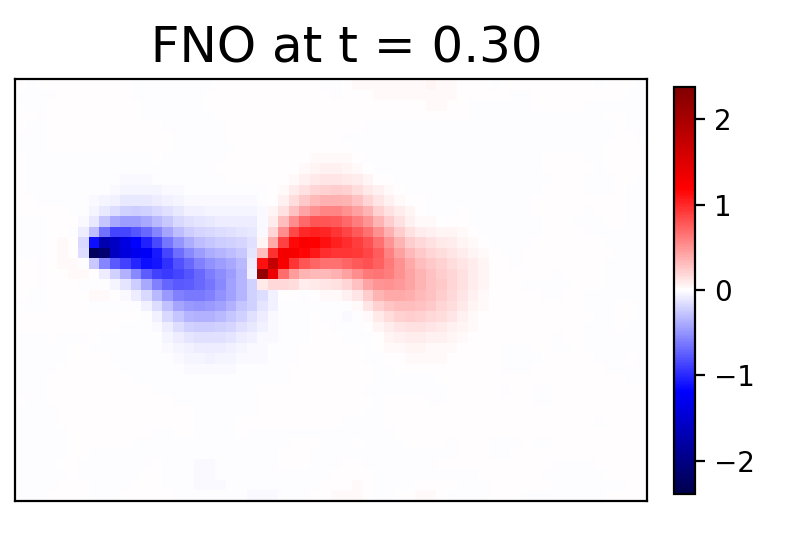} & 
    \includegraphics[width=0.22\textwidth]{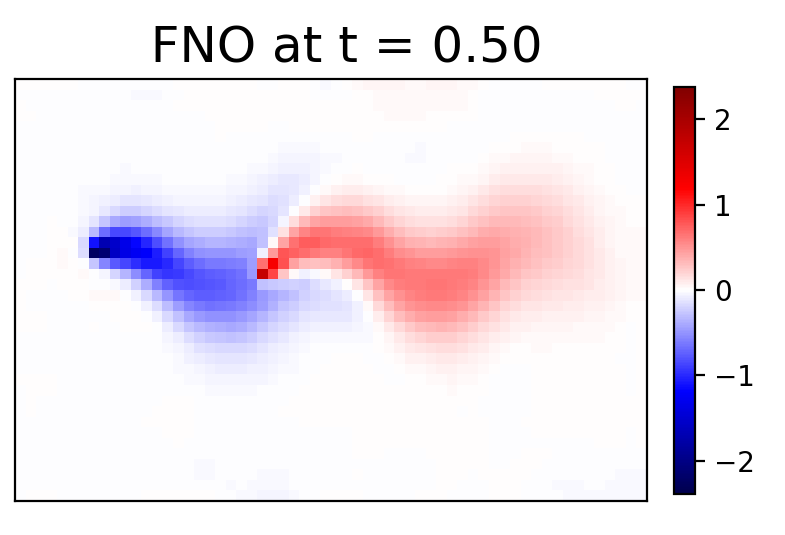} & 
    \includegraphics[width=0.22\textwidth]{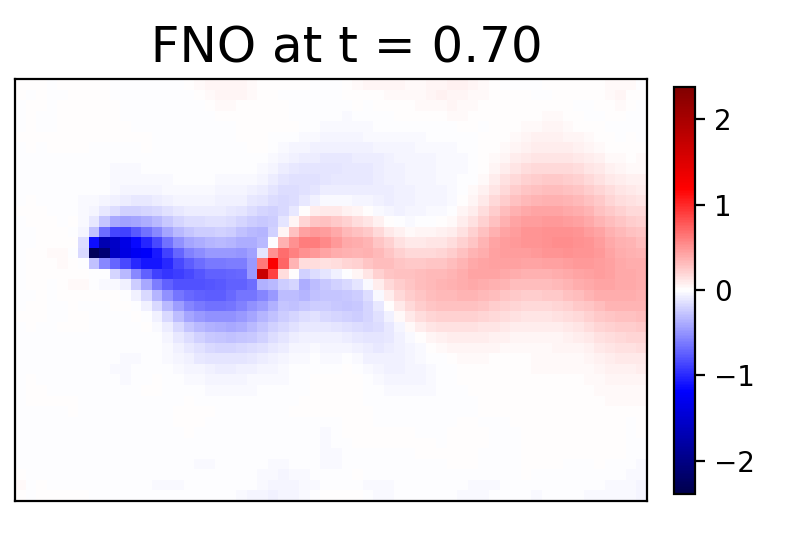}  \\
    \includegraphics[width=0.22\textwidth]{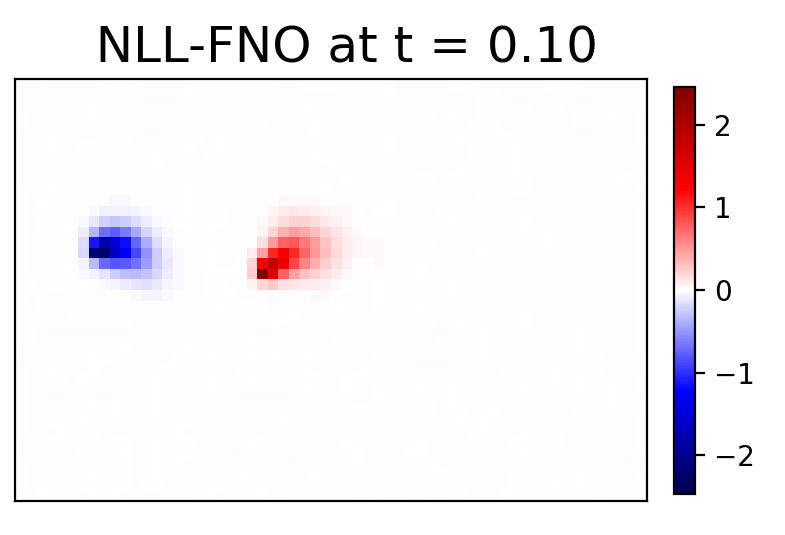} & 
    \includegraphics[width=0.22\textwidth]{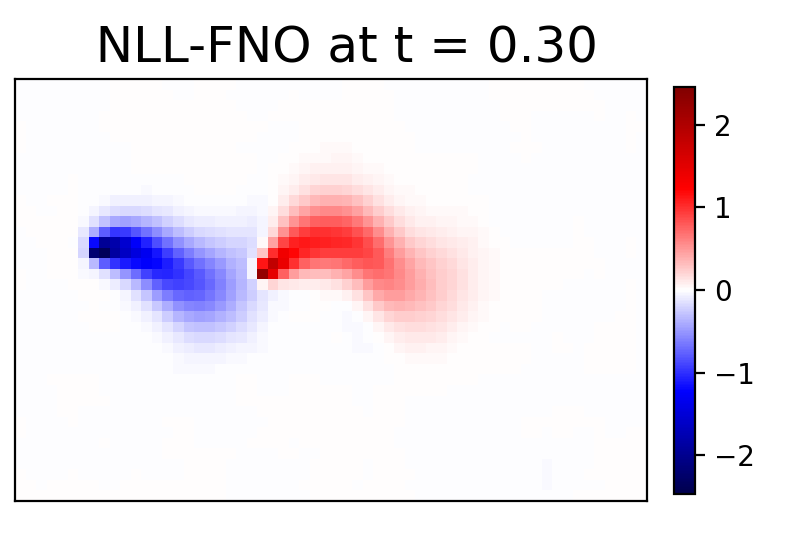} & 
    \includegraphics[width=0.22\textwidth]{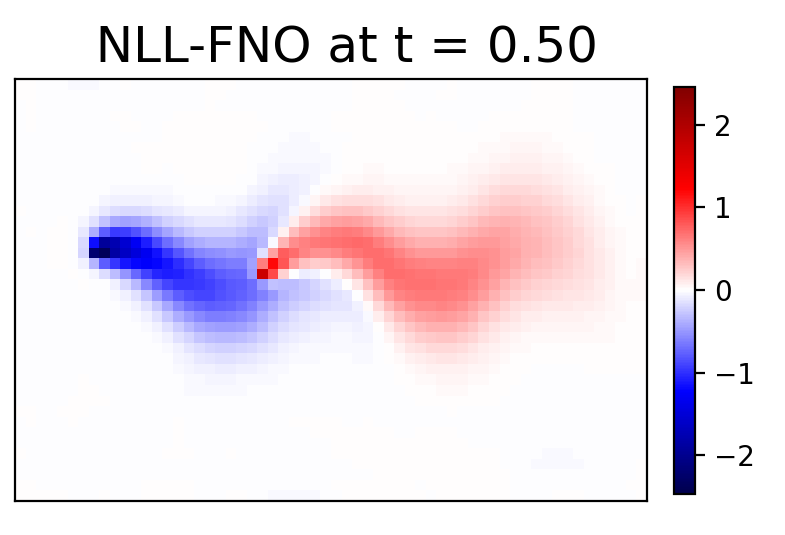} & 
    \includegraphics[width=0.22\textwidth]{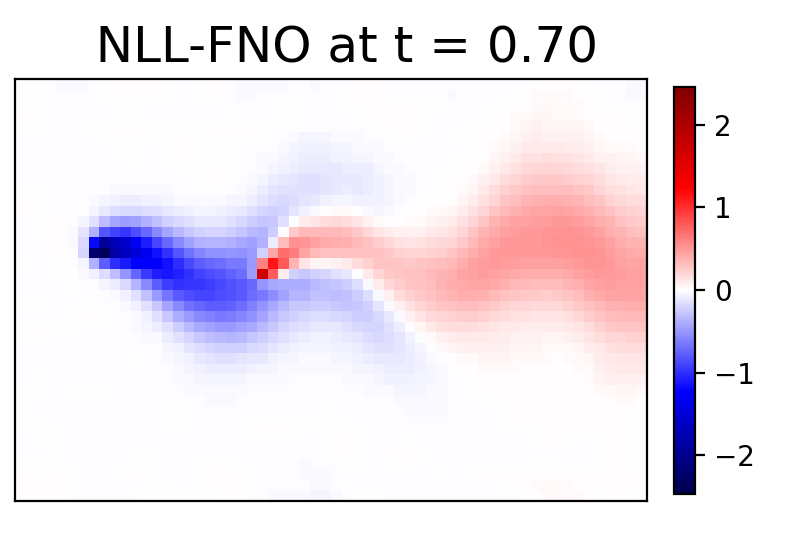}  \\
    \includegraphics[width=0.22\textwidth]{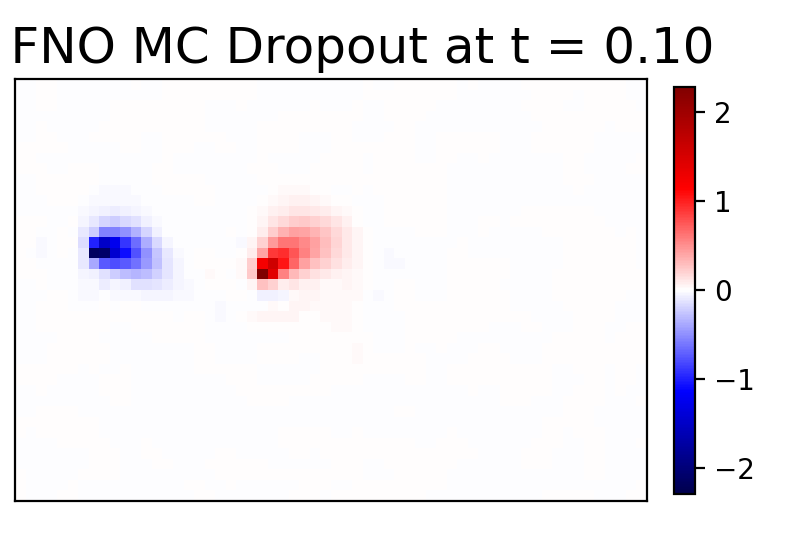} & 
    \includegraphics[width=0.22\textwidth]{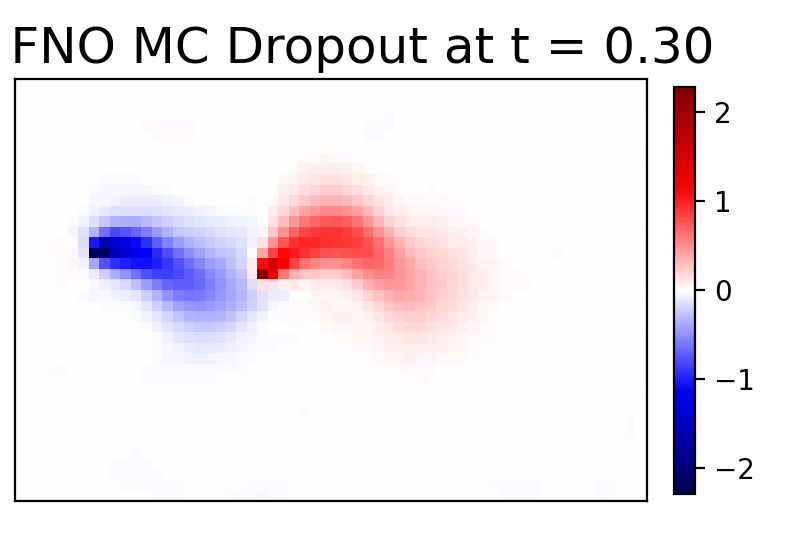} & 
    \includegraphics[width=0.22\textwidth]{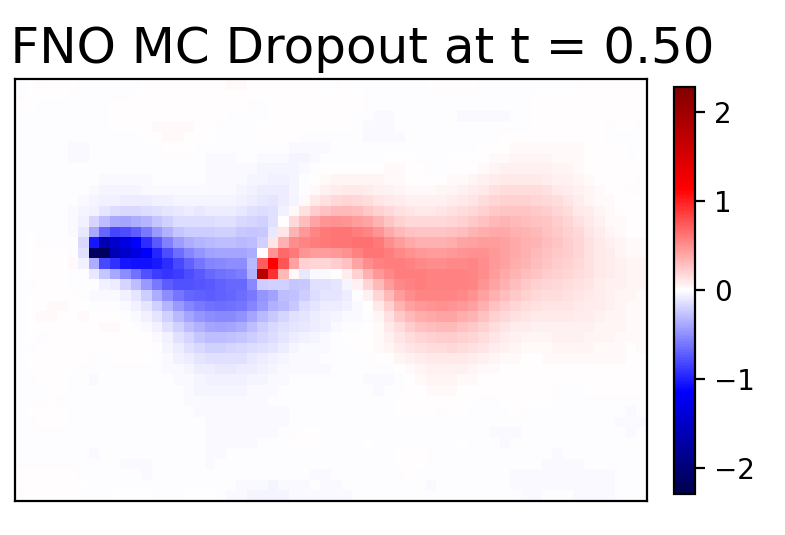} & 
    \includegraphics[width=0.22\textwidth]{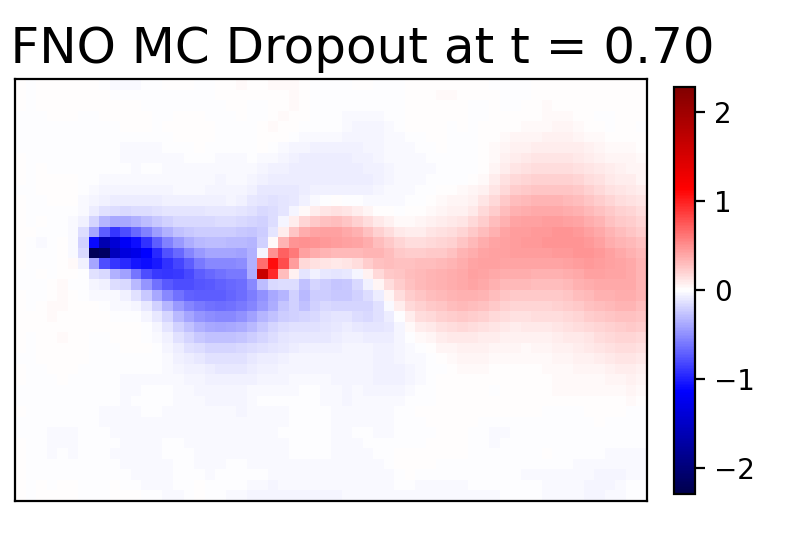}  \\
    \includegraphics[width=0.22\textwidth]{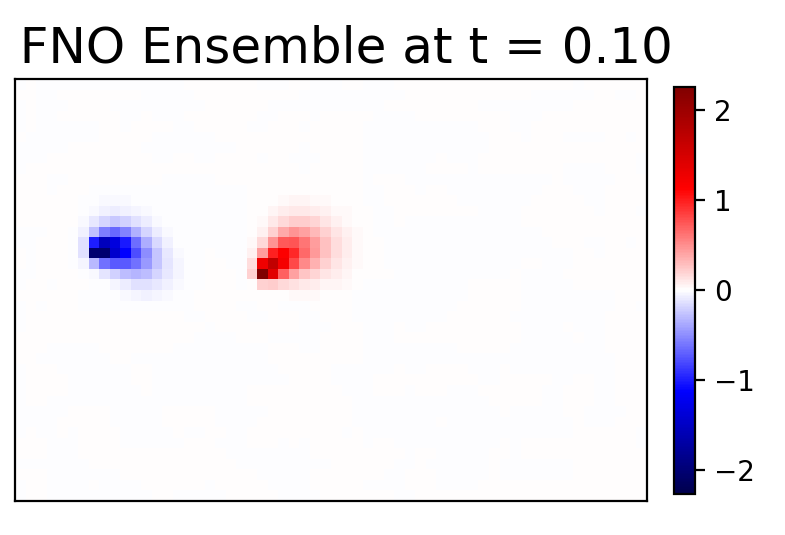} & 
    \includegraphics[width=0.22\textwidth]{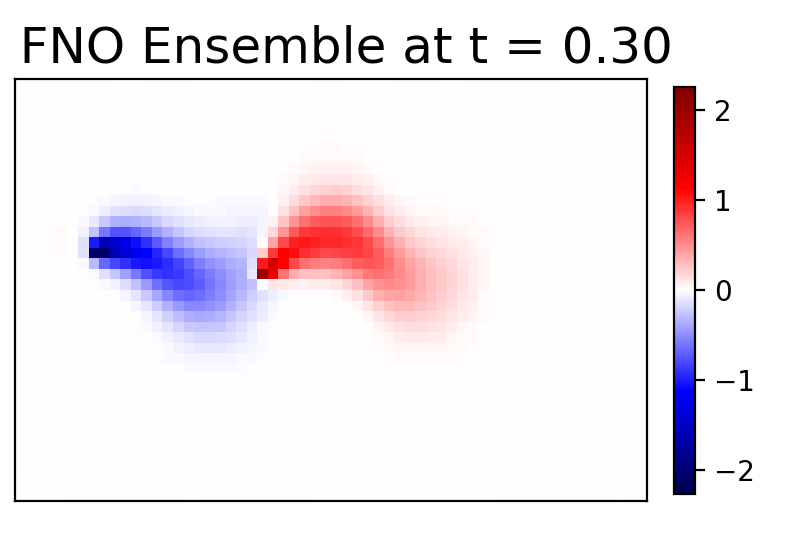} & 
    \includegraphics[width=0.22\textwidth]{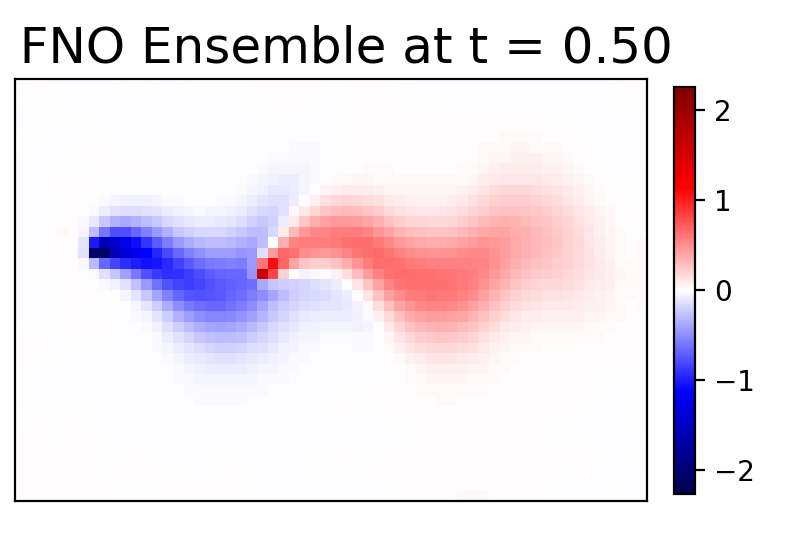} & 
    \includegraphics[width=0.22\textwidth]{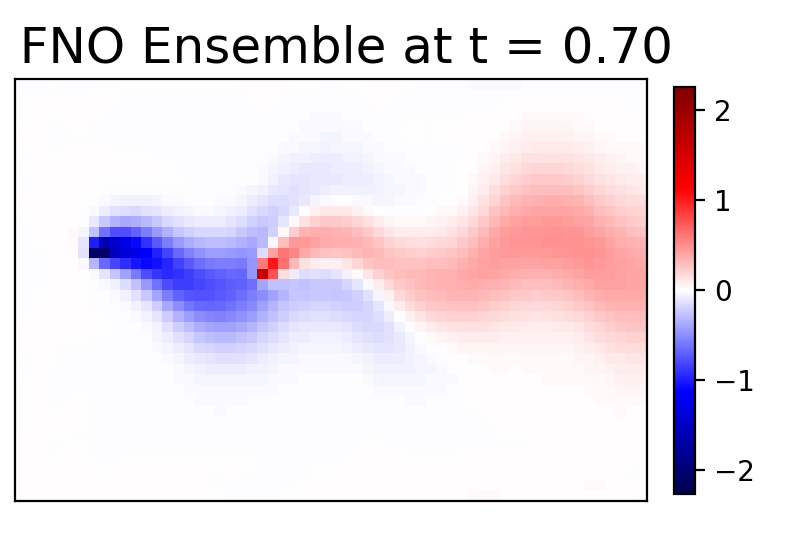}  \\
  \end{tabular}
  \caption{\revboth{Example solution and operator network predictions for the advection-diffusion problem setup.}}
  \label{fig:advection_qualitative_2}
\end{figure}

\revboth{Qualitative results for the advection-diffusion problem setup are shown in Figures~\ref{fig:advection_qualitative}-\ref{fig:advection_qualitative_2}.  A complete summary of the numerical results for the active learning problem across 10 random initializations are provided in Table~\ref{table:advection_combined_results}.}

\begin{figure}
  \centering
  \begin{tabular}{cccc}
    \includegraphics[width=0.22\textwidth]{./solution_3_0_crop.png} & 
    \includegraphics[width=0.22\textwidth]{./solution_3_20_crop.png} & 
    \includegraphics[width=0.22\textwidth]{./solution_3_40_crop.png} & 
    \includegraphics[width=0.22\textwidth]{./solution_3_75_crop.png} \\
    \includegraphics[width=0.22\textwidth]{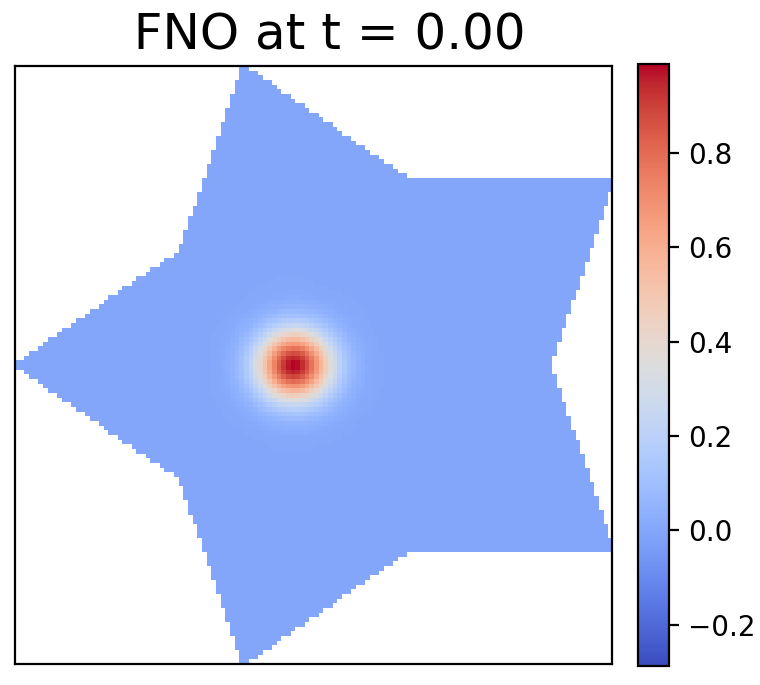} & 
    \includegraphics[width=0.22\textwidth]{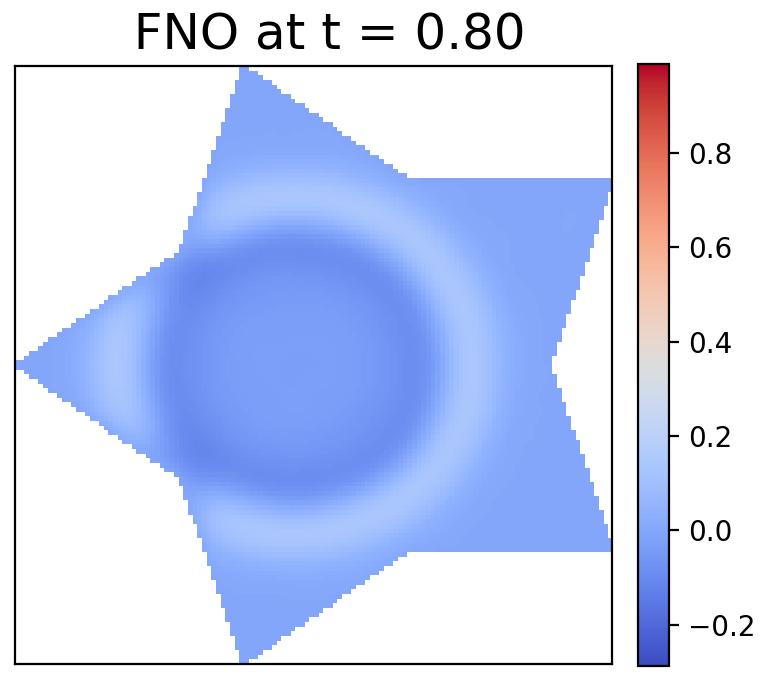} & 
    \includegraphics[width=0.22\textwidth]{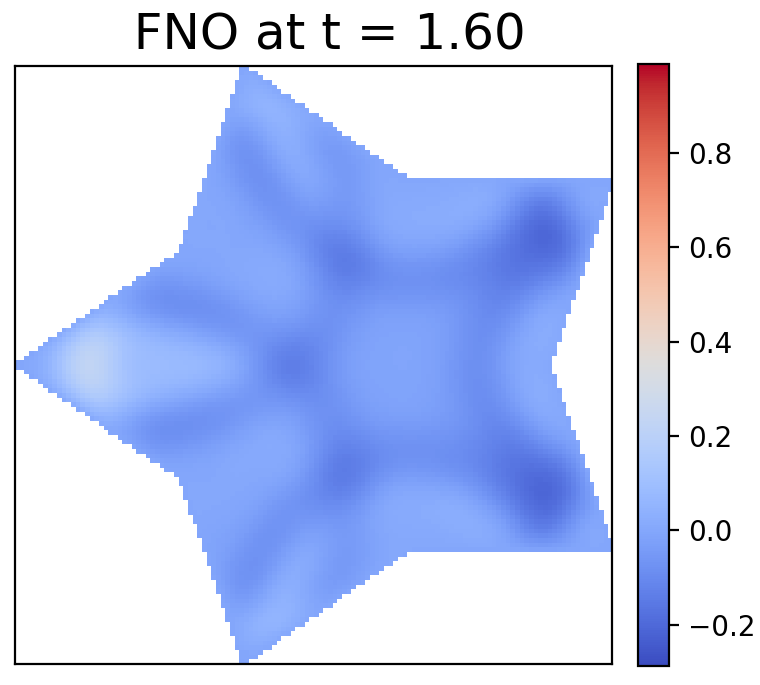} & 
    \includegraphics[width=0.22\textwidth]{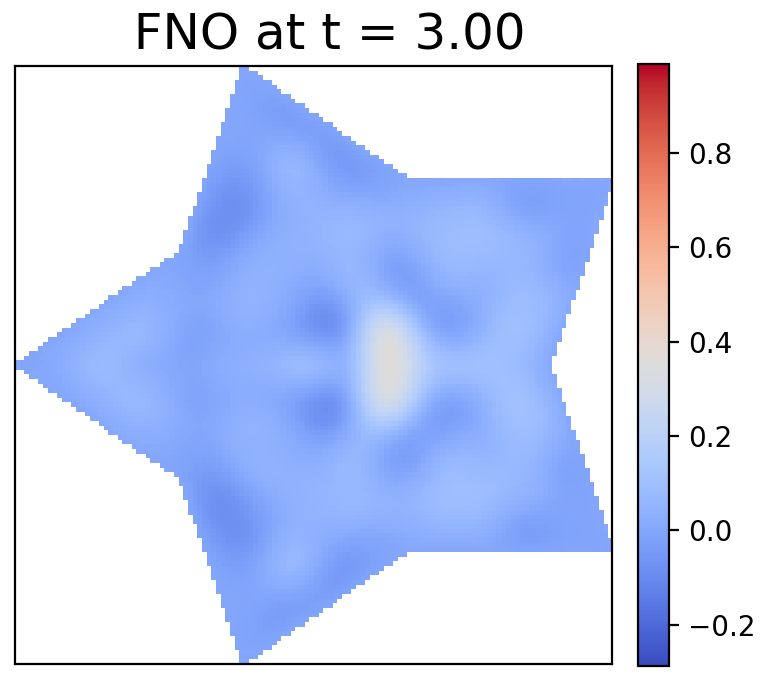} \\
    \includegraphics[width=0.22\textwidth]{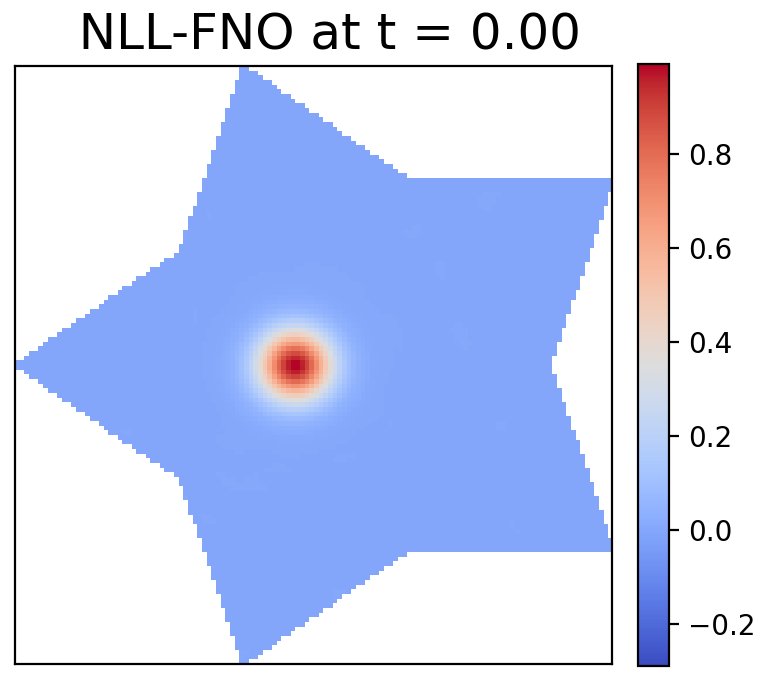} & 
    \includegraphics[width=0.22\textwidth]{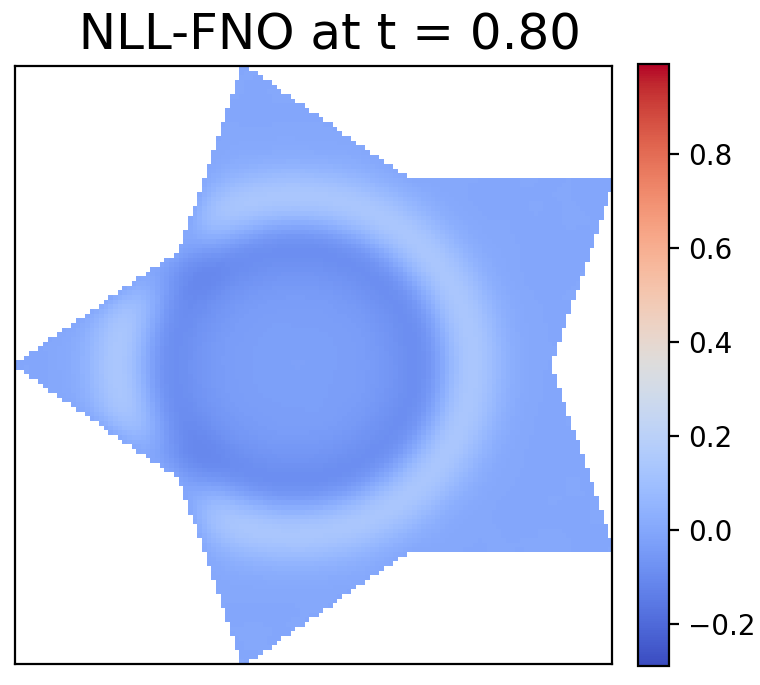} & 
    \includegraphics[width=0.22\textwidth]{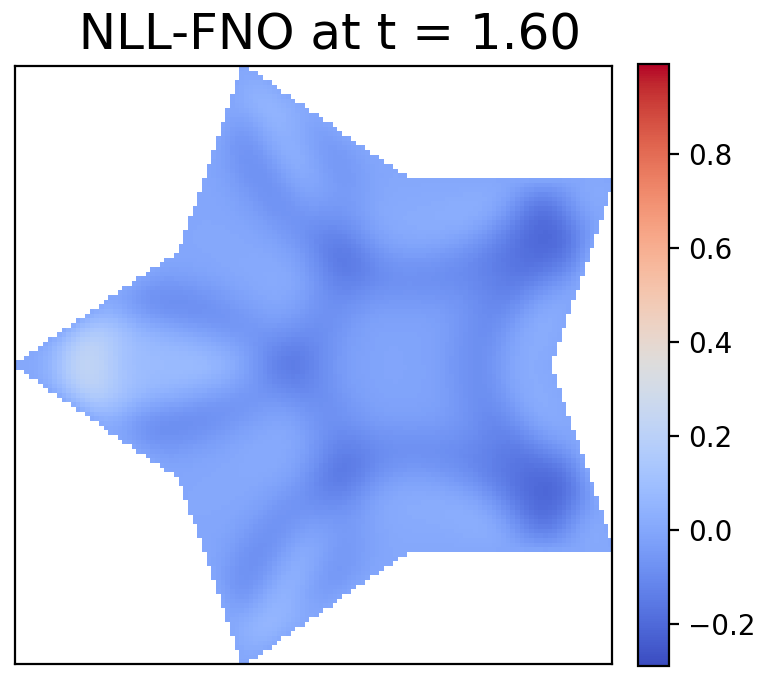} & 
    \includegraphics[width=0.22\textwidth]{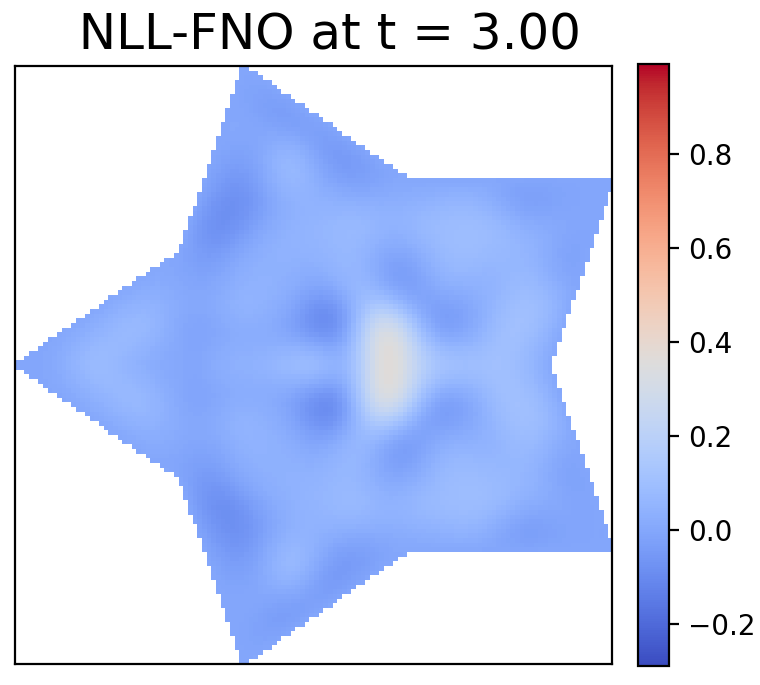} \\
    \includegraphics[width=0.22\textwidth]{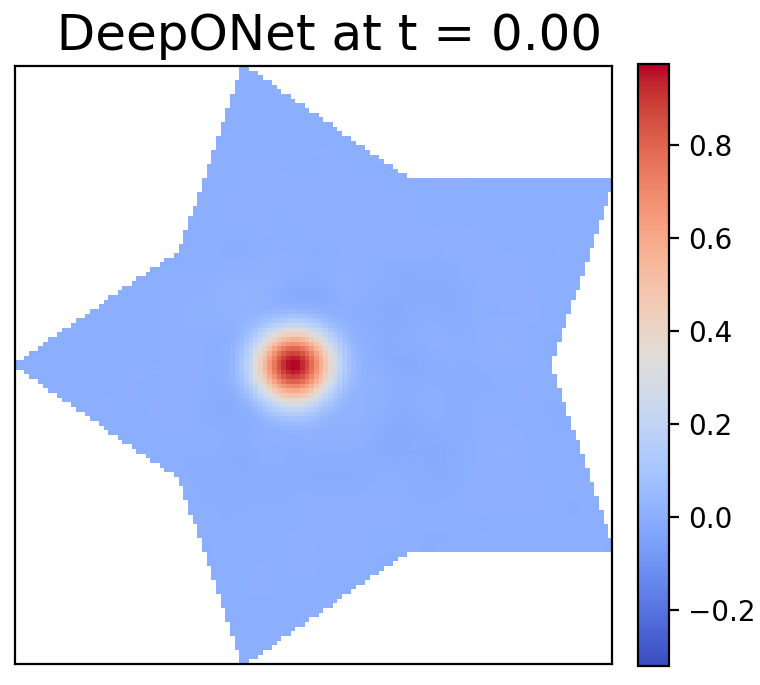} & 
    \includegraphics[width=0.22\textwidth]{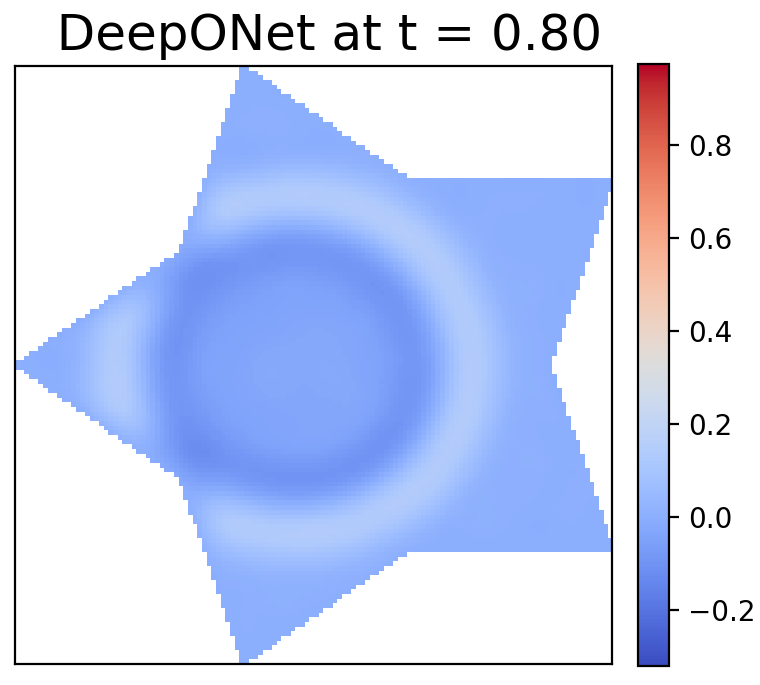} & 
    \includegraphics[width=0.22\textwidth]{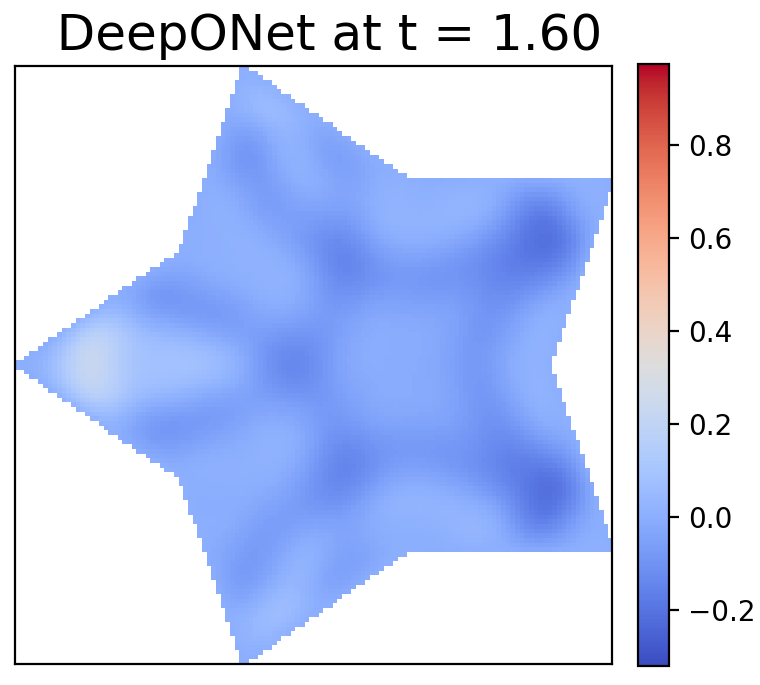} & 
    \includegraphics[width=0.22\textwidth]{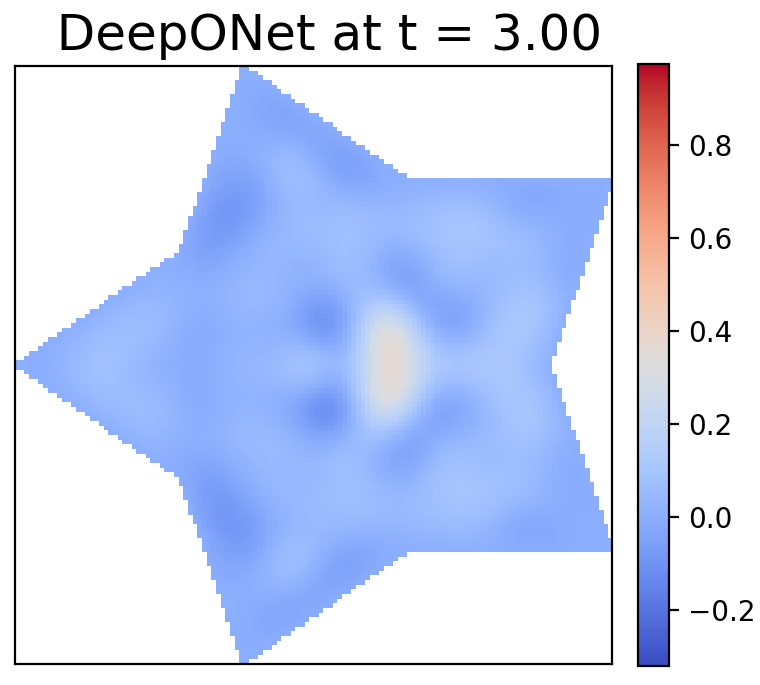} \\
    \includegraphics[width=0.22\textwidth]{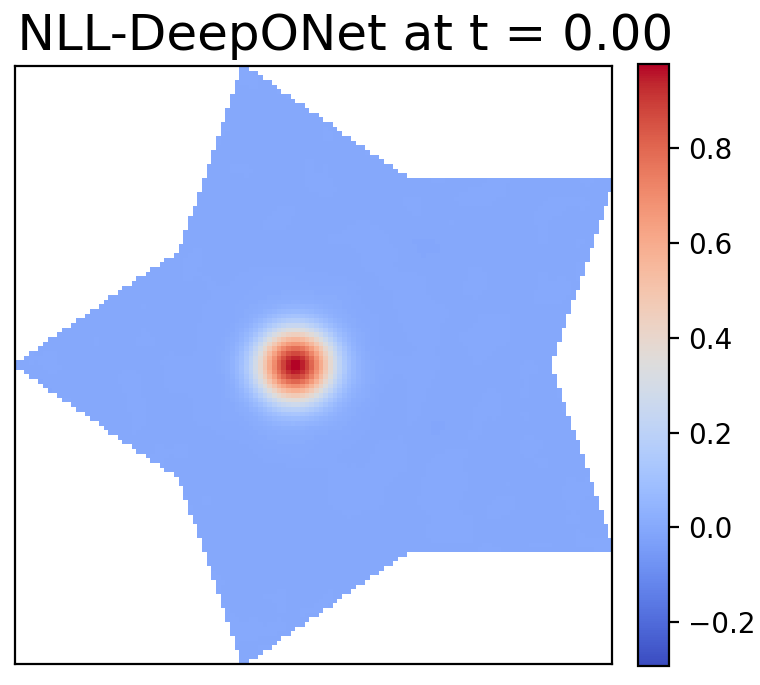} & 
    \includegraphics[width=0.22\textwidth]{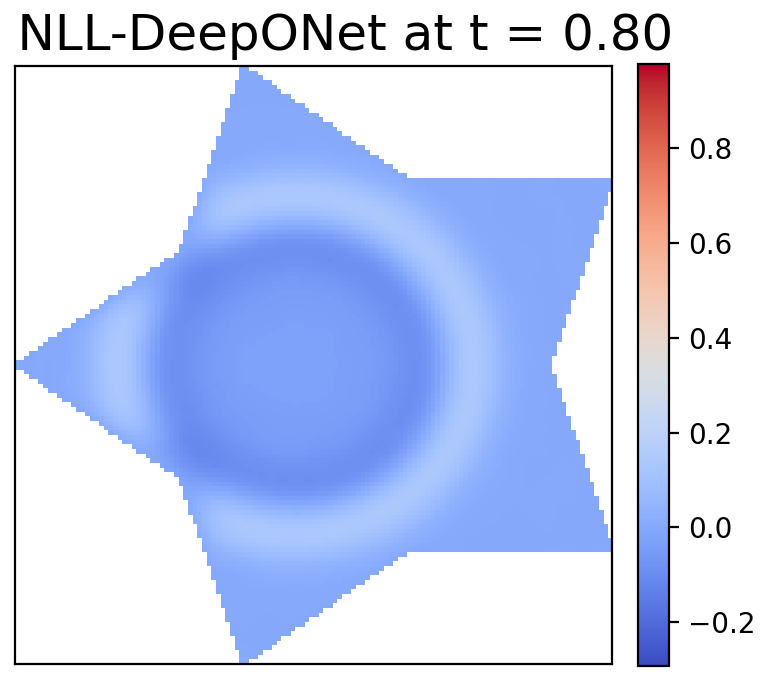} & 
    \includegraphics[width=0.22\textwidth]{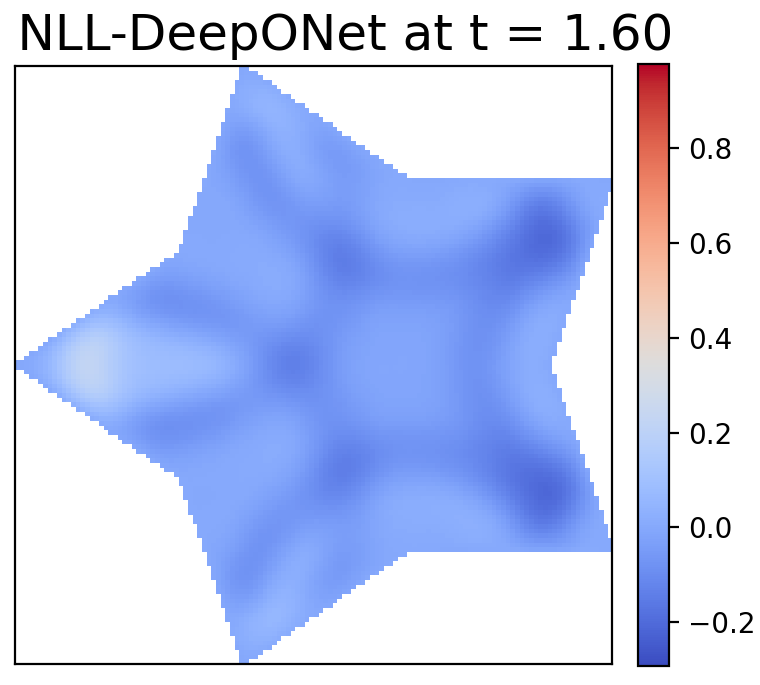} & 
    \includegraphics[width=0.22\textwidth]{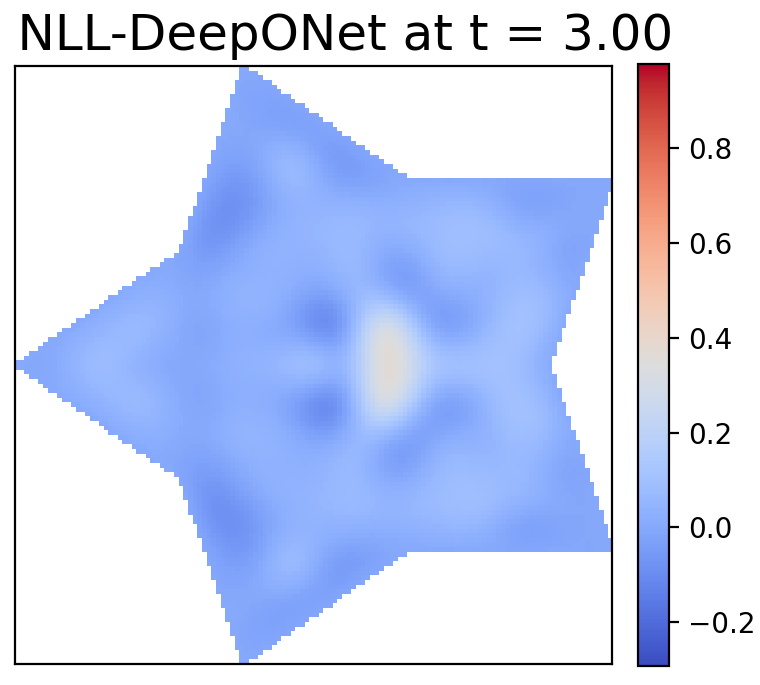} \\
  \end{tabular}
  \caption{\revboth{Example solution and operator network predictions for the wave equation problem setup.}}
  \label{fig:wave_qualitative}
\end{figure}

\revboth{Qualitative results for the wave equation problem setup are shown in Figure~\ref{fig:wave_qualitative}.  A complete summary of the numerical results for the active learning problem across 10 random initializations are provided in Table~\ref{table:al_combined_results}.}

\begin{table}[htbp]
\centering
{\fontsize{9}{9}\selectfont{
\begin{tabular}{ccccccccccc}
\hline
\multicolumn{11}{l}{\Tstrut\altspacer
{\bf{Active Learning Results for Advection Eq.: $L^2$ Relative Error Across 10 Random Initializations}}\Bstrut} \\
\hline
\hline
\Tstrut Data Count & 300 & 350 & 400 & 450 & 500 & 550 & 600 & 650 & 700 & 750 \Bstrut \\
\hline
\hline
\Tstrut FNO & 27.44\% & 18.77\% & 15.39\% & 13.43\% & 12.36\% & 11.32\% & 10.68\% & 10.28\% & 9.61\% & 9.14\% \\ 
NLL-FNO\, & 22.47\% & 16.34\% & 14.17\% & 12.65\% & 11.74\% & 10.94\% & 10.26\% & 9.67\% & 9.21\% & 8.78\% \\ 
Ensemble FNO & 27.08\% & 17.59\% & 14.28\% & 12.42\% & 11.10\% & 10.07\% & 9.33\% & 8.76\% & 8.24\% & 7.82\% \\ 
MC Dropout& 21.61\% & 17.11\% & 15.06\% & 13.44\% & 12.50\% & 11.47\% & 11.13\% & 10.17\% & 9.76\% & 9.38\%  \hspace{-0.05in}\Bstrut \\
\hline
\Tstrut FNO & 27.76\% & 18.75\% & 15.75\% & 13.83\% & 12.66\% & 11.53\% & 10.89\% & 10.24\% & 9.64\% & 9.19\% \\ 
NLL-FNO\, & 21.80\% & 15.82\% & 13.56\% & 12.12\% & 11.10\% & 10.10\% & 9.38\% & 8.80\% & 8.35\% & 7.99\% \\ 
Ensemble FNO & 27.92\% & 17.64\% & 14.05\% & 12.14\% & 10.90\% & 9.91\% & 9.19\% & 8.67\% & 8.17\% & 7.79\% \\ 
MC Dropout& 23.17\% & 17.72\% & 14.93\% & 13.42\% & 12.39\% & 11.20\% & 10.46\% & 9.96\% & 9.52\% & 9.18\%   \hspace{-0.05in}\Bstrut \\
\hline
\Tstrut FNO & 26.67\% & 18.13\% & 15.10\% & 13.45\% & 12.68\% & 11.51\% & 10.74\% & 10.17\% & 9.62\% & 9.07\% \\ 
NLL-FNO\, & 22.10\% & 16.00\% & 13.60\% & 12.05\% & 11.07\% & 10.19\% & 9.53\% & 9.02\% & 8.57\% & 8.16\% \\ 
Ensemble FNO & 28.51\% & 17.64\% & 13.96\% & 11.90\% & 10.71\% & 9.77\% & 9.07\% & 8.51\% & 8.01\% & 7.60\% \\ 
MC Dropout& 22.08\% & 17.03\% & 14.54\% & 12.93\% & 11.82\% & 10.95\% & 10.49\% & 10.74\% & 9.48\% & 8.97\%   \hspace{-0.05in}\Bstrut \\
\hline
\Tstrut FNO & 31.70\% & 19.36\% & 15.73\% & 13.39\% & 11.93\% & 10.77\% & 10.01\% & 9.45\% & 8.94\% & 8.74\% \\ 
NLL-FNO\, & 21.40\% & 15.24\% & 12.86\% & 11.47\% & 10.64\% & 9.85\% & 9.25\% & 8.77\% & 8.31\% & 7.94\% \\ 
Ensemble FNO & 28.31\% & 17.82\% & 14.13\% & 12.02\% & 10.77\% & 9.83\% & 9.10\% & 8.46\% & 7.95\% & 7.58\% \\ 
MC Dropout& 21.95\% & 16.97\% & 14.28\% & 12.95\% & 12.22\% & 11.48\% & 10.82\% & 10.03\% & 9.68\% & 9.35\%   \hspace{-0.05in}\Bstrut \\
\hline
\Tstrut FNO & 29.18\% & 18.78\% & 15.37\% & 13.21\% & 11.96\% & 10.98\% & 10.21\% & 9.60\% & 9.10\% & 8.70\% \\ 
NLL-FNO\, & 21.41\% & 15.28\% & 13.13\% & 11.78\% & 10.77\% & 9.98\% & 9.28\% & 8.75\% & 8.34\% & 7.93\% \\ 
Ensemble FNO & 27.32\% & 17.40\% & 13.84\% & 11.80\% & 10.63\% & 9.71\% & 8.97\% & 8.38\% & 7.93\% & 7.52\% \\ 
MC Dropout& 23.36\% & 17.49\% & 14.98\% & 13.24\% & 12.24\% & 11.17\% & 10.52\% & 9.99\% & 9.72\% & 9.13\%   \hspace{-0.05in}\Bstrut \\
\hline
\Tstrut FNO & 32.33\% & 20.14\% & 16.49\% & 14.16\% & 12.75\% & 11.46\% & 10.59\% & 9.91\% & 9.30\% & 8.91\% \\ 
NLL-FNO\, & 21.70\% & 16.08\% & 13.94\% & 12.44\% & 11.40\% & 10.38\% & 9.63\% & 9.07\% & 8.60\% & 8.21\% \\ 
Ensemble FNO & 28.82\% & 17.81\% & 14.02\% & 12.04\% & 10.79\% & 9.82\% & 9.09\% & 8.48\% & 7.93\% & 7.54\% \\ 
MC Dropout& 21.82\% & 17.35\% & 15.05\% & 13.56\% & 12.14\% & 11.64\% & 10.69\% & 9.95\% & 9.50\% & 9.12\%   \hspace{-0.05in}\Bstrut \\
\hline
\Tstrut FNO & 28.05\% & 18.97\% & 15.71\% & 13.88\% & 12.64\% & 11.63\% & 10.80\% & 10.13\% & 9.57\% & 9.16\% \\ 
NLL-FNO\, & 23.36\% & 16.67\% & 14.17\% & 12.63\% & 11.61\% & 10.67\% & 10.04\% & 9.51\% & 9.02\% & 8.61\% \\ 
Ensemble FNO & 28.16\% & 17.87\% & 14.39\% & 12.49\% & 11.06\% & 9.98\% & 9.26\% & 8.61\% & 8.14\% & 7.68\% \\ 
MC Dropout& 24.08\% & 18.41\% & 15.41\% & 13.74\% & 12.71\% & 11.58\% & 10.86\% & 10.25\% & 9.62\% & 9.30\%   \hspace{-0.05in}\Bstrut \\
\hline
\Tstrut FNO & 32.99\% & 20.61\% & 16.55\% & 14.14\% & 12.87\% & 11.78\% & 10.91\% & 10.23\% & 9.75\% & 9.41\% \\ 
NLL-FNO\, & 25.27\% & 17.55\% & 14.94\% & 13.33\% & 12.21\% & 11.18\% & 10.40\% & 9.76\% & 9.31\% & 8.90\% \\ 
Ensemble FNO & 28.27\% & 17.61\% & 14.17\% & 12.10\% & 10.81\% & 9.88\% & 9.17\% & 8.63\% & 8.17\% & 7.75\% \\ 
MC Dropout& 21.71\% & 16.44\% & 14.05\% & 12.48\% & 11.66\% & 10.94\% & 10.15\% & 9.83\% & 9.66\% & 8.92\%   \hspace{-0.05in}\Bstrut \\
\hline
\Tstrut FNO & 30.48\% & 19.36\% & 15.49\% & 13.45\% & 12.14\% & 11.09\% & 10.32\% & 9.66\% & 9.08\% & 8.63\% \\ 
NLL-FNO\, & 22.34\% & 15.94\% & 13.74\% & 12.40\% & 11.35\% & 10.41\% & 9.69\% & 9.15\% & 8.73\% & 8.34\% \\ 
Ensemble FNO & 28.13\% & 17.64\% & 14.11\% & 12.00\% & 10.71\% & 9.71\% & 8.94\% & 8.28\% & 7.83\% & 7.46\% \\ 
MC Dropout& 21.71\% & 16.63\% & 14.03\% & 12.51\% & 12.15\% & 10.62\% & 10.06\% & 9.54\% & 9.18\% & 8.96\%   \hspace{-0.05in}\Bstrut \\
\hline
\Tstrut FNO & 29.56\% & 19.71\% & 15.92\% & 14.03\% & 12.69\% & 11.61\% & 10.98\% & 10.26\% & 9.63\% & 9.20\% \\ 
NLL-FNO\, & 22.22\% & 15.73\% & 13.55\% & 12.12\% & 11.21\% & 10.29\% & 9.70\% & 9.25\% & 8.92\% & 8.49\% \\ 
Ensemble FNO & 27.35\% & 17.34\% & 13.76\% & 11.81\% & 10.56\% & 9.69\% & 8.94\% & 8.35\% & 7.87\% & 7.47\% \\ 
MC Dropout& 21.70\% & 16.46\% & 14.12\% & 12.75\% & 11.95\% & 11.44\% & 10.96\% & 9.94\% & 9.75\% & 9.60\%   \hspace{-0.05in}\Bstrut \\
\hline
\end{tabular}
}}
\caption{\revboth{Summary of $L^2$ relative error for the active learning advection problem across random initializations.}}
\label{table:advection_combined_results}
\end{table}

\clearpage

\clearpage

\begin{table}[htbp]
\centering
{\fontsize{9}{9}\selectfont{
\begin{tabular}{ccccccccccc}
\hline
\multicolumn{11}{l}{\Tstrut\altspacer
{\bf{Active Learning Results for Wave Eq.: $L^2$ Relative Error Across 10 Random Initializations}}\Bstrut} \\
\hline
\hline
\Tstrut Data Count & 300 & 350 & 400 & 450 & 500 & 550 & 600 & 650 & 700 & 750 \Bstrut \\
\hline
\hline
\Tstrut FNO & 18.55\% & 7.98\% & 6.38\% & 5.47\% & 4.57\% & 3.99\% & 3.54\% & 3.18\% & 2.96\% & 6.38\% \\ 
NLL-FNO\, & 16.99\% & 10.65\% & 7.91\% & 6.22\% & 4.98\% & 4.41\% & 3.47\% & 3.05\% & 2.92\% & 2.73\% \\ 
DeepONet & 21.89\% & 17.12\% & 13.71\% & 11.89\% & 11.00\% & 8.92\% & 8.30\% & 7.00\% & 6.94\% & 6.04\% \\ 
NLL-DeepONet\, & 21.69\% & 15.69\% & 12.96\% & 10.95\% & 9.54\% & 8.17\% & 7.47\% & 6.43\% & 5.80\% & 5.35\%  \hspace{-0.05in}\Bstrut \\
\hline
\Tstrut FNO & 12.71\% & 8.54\% & 6.42\% & 5.56\% & 4.87\% & 4.44\% & 3.51\% & 3.14\% & 2.88\% & 2.83\% \\ 
NLL-FNO\, & 16.01\% & 9.96\% & 7.70\% & 6.17\% & 5.16\% & 4.39\% & 3.79\% & 3.14\% & 2.76\% & 2.40\% \\ 
DeepONet & 25.00\% & 16.77\% & 13.74\% & 11.93\% & 10.29\% & 8.95\% & 7.78\% & 7.84\% & 7.21\% & 6.17\% \\ 
NLL-DeepONet\, & 23.00\% & 15.11\% & 12.28\% & 10.79\% & 9.57\% & 8.45\% & 7.53\% & 6.70\% & 6.10\% & 5.50\%  \hspace{-0.05in}\Bstrut \\
\hline
\Tstrut FNO & 11.40\% & 8.98\% & 6.94\% & 5.51\% & 5.20\% & 5.52\% & 3.94\% & 5.79\% & 3.34\% & 2.54\% \\ 
NLL-FNO\, & 17.37\% & 11.32\% & 7.85\% & 6.26\% & 5.36\% & 4.81\% & 4.13\% & 3.01\% & 2.59\% & 2.24\% \\ 
DeepONet & 28.79\% & 21.93\% & 16.94\% & 15.36\% & 12.56\% & 10.83\% & 9.59\% & 9.25\% & 7.79\% & 7.12\% \\ 
NLL-DeepONet\, & 22.97\% & 16.47\% & 13.30\% & 11.43\% & 9.90\% & 8.75\% & 7.76\% & 7.02\% & 6.26\% & 5.95\%  \hspace{-0.05in}\Bstrut \\
\hline
\Tstrut FNO & 11.82\% & 8.65\% & 9.47\% & 6.33\% & 7.24\% & 10.51\% & 4.15\% & 3.48\% & 3.15\% & 2.60\% \\ 
NLL-FNO\, & 13.84\% & 9.25\% & 7.33\% & 6.07\% & 5.01\% & 4.35\% & 3.68\% & 3.15\% & 2.69\% & 2.34\% \\ 
DeepONet & 38.09\% & 31.35\% & 25.93\% & 21.56\% & 17.11\% & 14.04\% & 11.66\% & 9.24\% & 7.93\% & 7.66\% \\ 
NLL-DeepONet\, & 19.24\% & 14.31\% & 11.94\% & 10.18\% & 9.02\% & 7.71\% & 6.60\% & 5.91\% & 5.26\% & 4.77\%  \hspace{-0.05in}\Bstrut \\
\hline
\Tstrut FNO & 11.57\% & 8.44\% & 5.93\% & 5.22\% & 4.85\% & 3.85\% & 4.20\% & 3.06\% & 3.03\% & 2.48\% \\ 
NLL-FNO\, & 16.69\% & 10.74\% & 8.48\% & 7.17\% & 5.65\% & 4.94\% & 3.60\% & 2.97\% & 2.60\% & 2.29\% \\ 
DeepONet & 29.46\% & 21.88\% & 17.93\% & 15.17\% & 13.40\% & 12.82\% & 10.00\% & 9.39\% & 8.82\% & 8.07\% \\ 
NLL-DeepONet\, & 22.03\% & 17.03\% & 14.54\% & 12.54\% & 11.06\% & 9.69\% & 8.69\% & 7.94\% & 6.96\% & 6.21\%  \hspace{-0.05in}\Bstrut \\
\hline
\Tstrut FNO & 11.15\% & 8.26\% & 7.79\% & 5.73\% & 96.28\% & 5.89\% & 4.63\% & 3.67\% & 3.46\% & 3.13\% \\ 
NLL-FNO\, & 15.82\% & 10.75\% & 7.69\% & 6.06\% & 4.97\% & 4.28\% & 3.57\% & 3.13\% & 3.12\% & 2.70\% \\ 
DeepONet & 25.49\% & 18.59\% & 18.03\% & 12.34\% & 11.13\% & 9.39\% & 8.42\% & 8.10\% & 9.24\% & 6.29\% \\ 
NLL-DeepONet\, & 22.50\% & 16.37\% & 13.91\% & 11.69\% & 9.91\% & 8.91\% & 7.70\% & 6.88\% & 6.43\% & 5.52\%  \hspace{-0.05in}\Bstrut \\
\hline
\Tstrut FNO & 10.75\% & 7.79\% & 6.26\% & 5.46\% & 4.63\% & 3.92\% & 3.19\% & 3.48\% & 2.38\% & 2.31\% \\ 
NLL-FNO\, & 16.22\% & 10.50\% & 8.29\% & 6.50\% & 5.78\% & 4.88\% & 4.10\% & 3.65\% & 3.28\% & 2.56\% \\ 
DeepONet & 33.47\% & 26.43\% & 20.09\% & 17.19\% & 18.41\% & 11.96\% & 10.58\% & 9.33\% & 8.49\% & 7.88\% \\ 
NLL-DeepONet\, & 24.55\% & 16.98\% & 13.81\% & 11.59\% & 9.86\% & 8.52\% & 7.43\% & 6.74\% & 6.07\% & 5.47\%  \hspace{-0.05in}\Bstrut \\
\hline
\Tstrut FNO & 11.62\% & 7.68\% & 6.05\% & 5.17\% & 4.81\% & 3.99\% & 3.33\% & 3.06\% & 2.90\% & 2.41\% \\ 
NLL-FNO\, & 15.09\% & 10.44\% & 7.32\% & 5.57\% & 4.73\% & 3.41\% & 3.01\% & 2.91\% & 2.38\% & 2.40\% \\ 
DeepONet & 19.73\% & 14.80\% & 11.58\% & 9.91\% & 9.09\% & 7.84\% & 7.03\% & 7.22\% & 5.94\% & 5.32\% \\ 
NLL-DeepONet\, & 20.08\% & 15.29\% & 12.46\% & 11.14\% & 10.07\% & 8.99\% & 7.84\% & 6.87\% & 6.05\% & 5.50\%  \hspace{-0.05in}\Bstrut \\
\hline
\Tstrut FNO & 13.48\% & 8.50\% & 6.61\% & 5.59\% & 5.00\% & 4.20\% & 3.92\% & 3.79\% & 3.20\% & 2.94\% \\ 
NLL-FNO\, & 14.75\% & 10.14\% & 7.88\% & 6.33\% & 4.98\% & 4.25\% & 3.59\% & 3.03\% & 2.64\% & 2.28\% \\ 
DeepONet & 40.40\% & 32.20\% & 30.55\% & 22.41\% & 18.74\% & 14.41\% & 12.11\% & 10.98\% & 10.12\% & 8.98\% \\ 
NLL-DeepONet\, & 30.60\% & 49.68\% & 14.62\% & 12.87\% & 10.87\% & 9.46\% & 8.39\% & 7.38\% & 6.79\% & 6.16\%  \hspace{-0.05in}\Bstrut \\
\hline
\Tstrut FNO & 11.39\% & 8.88\% & 6.69\% & 6.38\% & 4.53\% & 4.11\% & 3.45\% & 2.99\% & 2.87\% & 2.41\% \\ 
NLL-FNO\, & 16.87\% & 10.93\% & 8.80\% & 7.38\% & 6.30\% & 5.20\% & 4.36\% & 3.58\% & 3.07\% & 2.72\% \\ 
DeepONet & 25.77\% & 21.24\% & 15.86\% & 13.89\% & 11.74\% & 9.99\% & 8.82\% & 8.70\% & 7.48\% & 7.82\% \\ 
NLL-DeepONet\, & 22.86\% & 15.67\% & 12.89\% & 11.01\% & 9.59\% & 8.24\% & 7.17\% & 6.69\% & 5.91\% & 6.04\%  \hspace{-0.05in}\Bstrut \\
\hline
\end{tabular}
}}
\caption{\revboth{Summary of $L^2$ relative error for active learning wave equation setup across random initializations.}}
\label{table:al_combined_results}
\end{table}

\bibliographystyle{elsarticle-num} 
\bibliography{biblio.bib}


%
%

\end{document}